\documentclass[journal]{IEEEtran}

\usepackage{graphicx} 
\usepackage{textcomp}
\usepackage{booktabs}
\usepackage{makecell}
\usepackage{gensymb}
\usepackage{amssymb}
\usepackage{graphicx}

\usepackage{rotating}
\usepackage{tabularx}
\usepackage{array}
\usepackage{booktabs}
\usepackage{pifont}

\usepackage{multirow}

\usepackage[justification=centering]{caption}
\usepackage[table]{xcolor}
\usepackage{epstopdf}
\usepackage{bbm}

\usepackage{color}

\makeatletter
\def\ifundefined{\@ifundefined}
\makeatother

\usepackage{subfigure,epsfig,amsfonts,amsmath,amssymb,graphics, latexsym, times,algorithm, algorithmic,bbm,cite,bm}

\usepackage[version=4]{mhchem}
\usepackage{makecell}

\ifCLASSINFOpdf
  
\else
  
\fi

\usepackage{amsmath}

\begin{document}

\title{Radar and Camera Fusion for Object Detection and Tracking: A Comprehensive Survey}

\author{Kun Shi,
        Shibo He,~\IEEEmembership{Senior Member, IEEE},
        Zhenyu Shi, Anjun Chen, 
        Zehui Xiong, 
        Jiming Chen,~\IEEEmembership{Fellow,~IEEE}, \\
        and Jun Luo,~\IEEEmembership{Fellow,~IEEE} 
\thanks{Corresponding author: \textit{Jun Luo.}}
\thanks{Kun Shi and Jun Luo are with the School of Computer Science and Engineering, Nanyang Technological University, Singapore 639798, Email: \{kun.shi, junluo\}@ntu.edu.sg.}
\thanks{Shibo He, Zhenyu Shi, Anjun Chen, and Jiming Chen are with the State Key Laboratory of Industrial Control Technology, Zhejiang University, Hangzhou 310027, China. Email: \{s18he, shizhenyu123, anjunchen, cjm\}@zju.edu.cn. Jiming Chen is also with the Hangzhou Dianzi University, Hangzhou 310005, China.}
\thanks{Zehui Xiong is with the Pillar of Information Systems Technology and Design, Singapore University of Technology and Design, Singapore 487372, Email: \{zehui\_xiong\}@sutd.edu.sg.}
}

\markboth{IEEE Communications Surveys \& Tutorials}
{He \MakeLowercase{\textit{et al.}}: Radar and Camera Fusion for Object Detection and Tracking: A Comprehensive Survey}

\maketitle

\begin{abstract}
Multi-modal fusion is imperative to the implementation of reliable object detection and tracking in complex environments. Exploiting the synergy of heterogeneous modal information endows perception systems the ability to achieve more comprehensive, robust, and accurate performance. 
As a nucleus concern in wireless-vision collaboration, radar-camera fusion has prompted prospective research directions owing to its extensive applicability, complementarity, and compatibility.
Nonetheless, there still lacks a systematic survey specifically focusing on deep fusion of radar and camera for object detection and tracking.
To fill this void, we embark on an endeavor to comprehensively review radar-camera fusion in a holistic way.
First, we elaborate on the fundamental principles, methodologies, and applications of radar-camera fusion perception. Next, we delve into the key techniques concerning sensor calibration, modal representation, data alignment, and fusion operation.
Furthermore, we provide a detailed taxonomy covering the research topics related to object detection and tracking in the context of radar and camera technologies.
Finally, we discuss the emerging perspectives in the field of radar-camera fusion perception and highlight the potential areas for future research.
\end{abstract}

\begin{IEEEkeywords}
radar-camera fusion, object detection, object tracking, deep learning.
\end{IEEEkeywords}

\IEEEpeerreviewmaketitle

\section{Introduction}
\IEEEPARstart{M}{odern} information technology has spurred the unprecedented proliferation of versatile sensors and data sources in intelligent systems, resulting in a wealth of heterogeneous information. 
Multimodal fusion perception, emerging as a protrusive technique, catalyzes a revolutionary impact in both industrial and academic frontier
\cite{cheng2023intelligent}.
Meanwhile, a constellation of promising technologies combining ubiquitous wireless sensing and maturing vision perception have upsurged  \cite{cai2022ubiquitous}.
Among them, radar-camera fusion possesses the acknowledged prospect of seamlessly integrating the wireless industry with vision technique.
By leveraging complementary strengths across multi-dimensional data while mitigating redundant information, radar-vision fusion exhibits enhanced environmental adaptability, fortifying reliability and fault tolerance.

To attain a comprehensive mastery of perceived object, a perception system may encompass multiple tasks.
Among these perception tasks, object detection and tracking are two of the most essential components.
For intuitiveness, a performance comparison of common sensors in the realm of object detection and tracking is listed in Table \ref{table: sensors}.
In challenging conditions like inclement weather, adverse lighting, and severe occlusion, reliance on multi-modal fusion becomes imperative.

\begin{table}[htpb] 
\caption{Common sensors in object detection and tracking}
\label{table: sensors}
\centering
\resizebox{1\columnwidth}{!}{
\begin{tabular}{c|ccccc}
\toprule
        & {\makecell[c]{Ultrasonic\\sensor}} & {\makecell[c]{Infrared\\sensor}} & LiDAR & {\makecell[c]{RGB\\Camera}}  & Radar \\ 
\midrule
   {\makecell[c]{Semantic\\Information}}    & $\times$ & $\Box$ & $\triangle$ & $\bigcirc$ & $\times$  \\ \hline
   {\makecell[c]{Range\\Measurement}}    & $\times$ & $\triangle$ & $\Box$ & $\triangle$ & $\bigcirc$ \\ \hline
   {\makecell[c]{Angular\\Resolution}} & $\times$& $\bigcirc$ & $\Box$ & $\bigcirc$ & $\times$ \\ \hline
   {\makecell[c]{Velocity\\Measurement}} & $\triangle$ & $\times$ & $\times$ & $\times$ & $\bigcirc$ \\ \hline
   {\makecell[c]{Illumination\\Adaptability}} & $\bigcirc$ & $\triangle$ & $\bigcirc$ & $\times$ & $\bigcirc$ \\ \hline
   {\makecell[c]{Weather\\Robustness}}   & $\bigcirc$ & $\times$ & $\triangle$ & $\times$ & $\bigcirc$ \\ \hline
   {\makecell[c]{Smoke\\Susceptibility}} & $\triangle$ & $\times$ & $\times$ & $\times$ & $\bigcirc$ \\ \hline
   {\makecell[c]{Temperature\\Stability}}   & $\times$ & $\triangle$ & $\bigcirc$ & $\bigcirc$ & $\bigcirc$ \\
\bottomrule
\multicolumn{6}{l}{\scriptsize  $\times$: Poor, $\triangle$: Fair, $\Box$: Good, $\bigcirc$: Excellent}\\
\end{tabular}}
\end{table}

\begin{figure}[h]
\centering
\subfigure[Fusion functionality of radar and camera]{
\includegraphics[width=8.5cm]{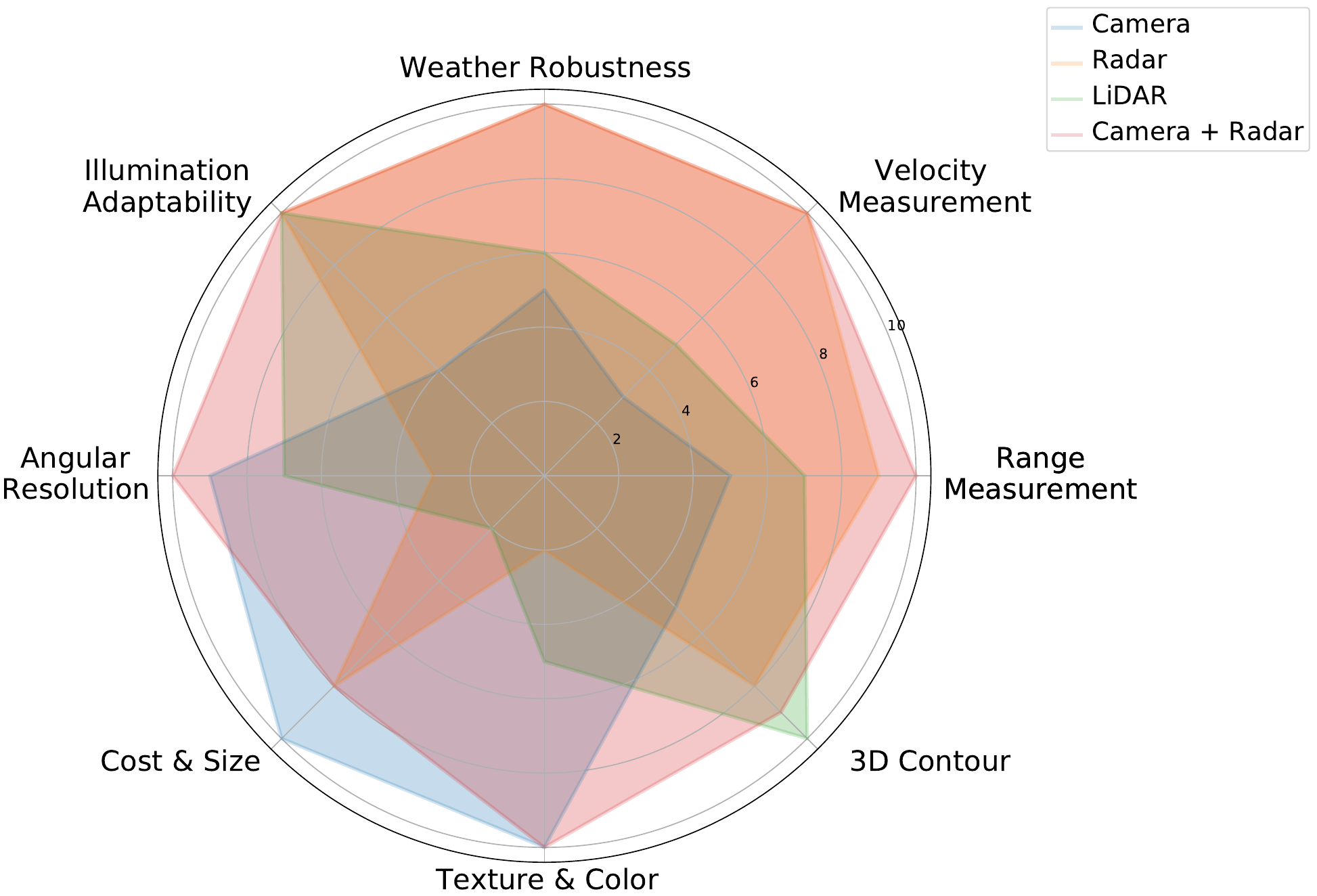}
}
\quad
\subfigure[Fusion functionality of LiDAR and camera]{
\includegraphics[width=8.8cm]{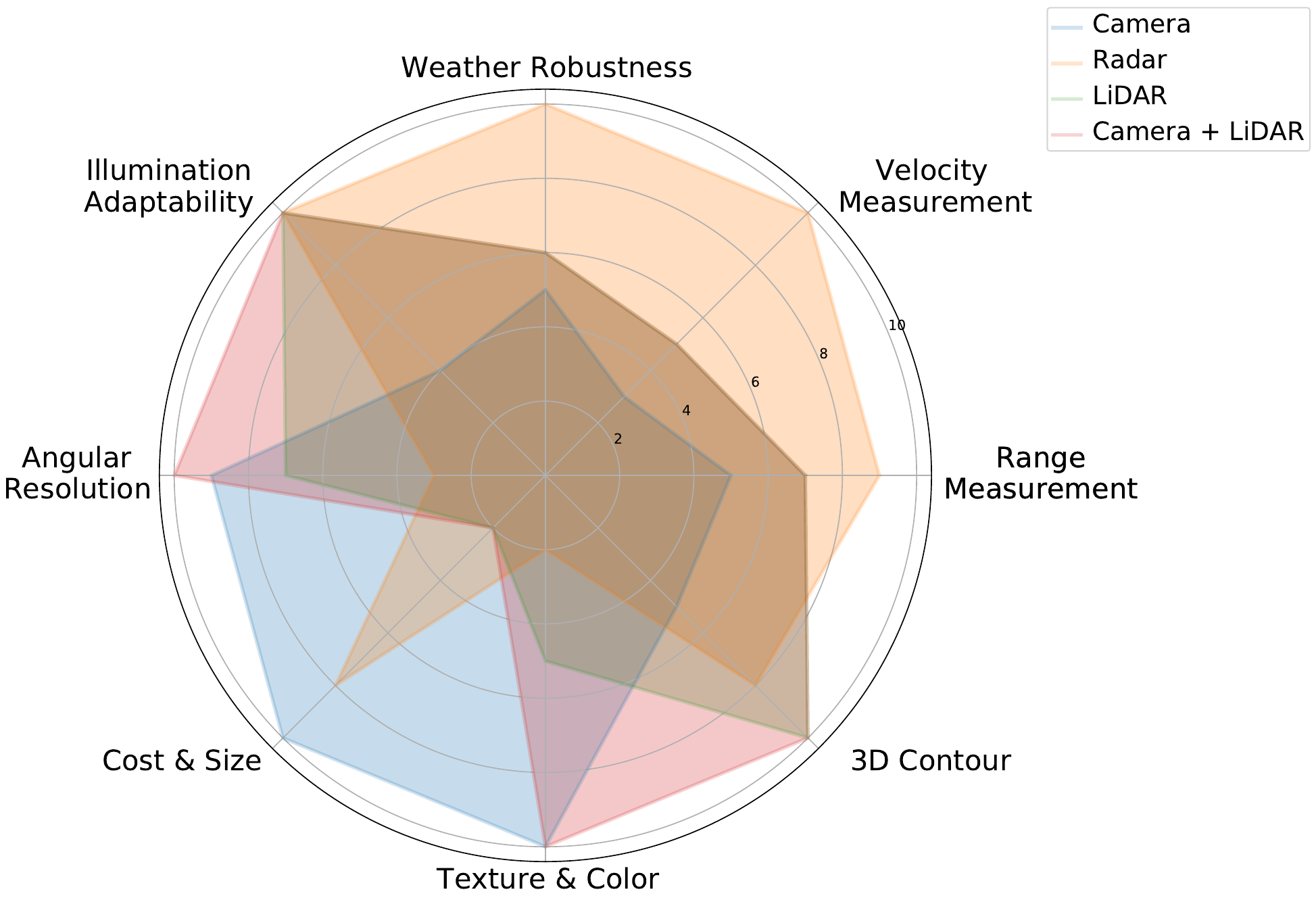}
}
\caption{Sensor characteristics of radar, camera and LiDAR.}
\label{fig:radar chart}
\end{figure}

Evidently, millimeter-wave (mmWave) radar is the unique sensing modality that operates effectively under all weather, illumination, and temperature conditions, which makes it compelling in a growing body of wireless sensing scenarios \cite{zhang2023survey}. 
Additionally, methods conducting radar-based environmental awareness also align synergistically with the Joint Communication and Sensing (JCAS) paradigm in 6G cellular communication \cite{han2022semantic}.
Albeit offering impressive and superior capabilities, radar severely suffers from data sparsity (i.e., limited angular resolution) and clutter interference (i.e., spurious measurement). 
One promising solution is to synergize and capitalize on the high-resolution camera images to complement the sparse and noisy radar data.
As illustrated in Fig.~\ref{fig:radar chart}(a), radar-camera fusion promotes a holistic perception across all the weather and lighting circumstances.
By contrast, as depicted in Fig.~\ref{fig:radar chart}(b), the perception capability of both LiDAR and camera sensors insurmountably deteriorates in inclement weather conditions such as snow and rain. Besides, LiDAR-camera fusion schemes are inherently confronted with deficient long-range and instantaneous-velocity measurement abilities. Moreover, LiDAR sensors are expensive in cost and heavy in energy/bandwidth consumption.
In terms of practical deployment costs, both radar and camera sensors have undergone mass production, facilitating their widespread adoption at the edge.

Despite the favorable complementarity between radar and camera data, relatively few work has been dedicated to radar-camera fusion on account of several challenges. 
First of all, radar data presents inherent deficiencies in terms of insurmountable sparsity, uncertainty, ambiguity, and incompleteness \cite{ahmed2023machine}. 
Secondly, the availability of public datasets tailoring for radar and camera data is inferior and inadequate (e.g., imprecise calibration and limited scale), posing challenges for researchers to conduct thorough analyses. 
Thirdly, the intrinsic discrepancy (e.g., density and geometric characteristics of point clouds) between radar and LiDAR renders straightforward imitations of LiDAR-camera fusion customized methods, unsatisfactory and even futile.
Last but not least, research on radar-camera fusion entails interdisciplinary expertise to effectively harness a diverse range of model-based and data-driven approaches, which are essential for overcoming modal disparateness, imperfection, and inconsistency in data.
Overall, these intricate challenges radically yield enormous impediments in the desirable implementation of radar-camera fusion. 
Fortunately, an upsurge in research has been aroused within the multimodal Generative Artificial Intelligence (GAI) \cite{xu2024unleashing} and multimodal Large Language Model (LLM) \cite{wang2024omnidrive}, which envision insightful and prospective directions for radar-camera fusion.

\renewcommand\arraystretch{1.2}
\begin{table*}
\centering
\caption{An overview of selected surveys in multi-modal object detection and tracking}
\label{tab:An overview of selected surveys}
\begin{tabular}{|m{1.5cm}<{\centering}|m{1.2cm}<{\centering}|m{0.9cm}<{\centering}|m{0.6cm}<{\centering}|m{12.3cm}|}
\hline
\rowcolor{gray!40} \textbf{Subject} & \textbf{Modality} & \textbf{Task} & \textbf{Ref}  & \multicolumn{1}{c|}{\textbf{Description}}   \\ \hline

\multirow{21}{*}{\makecell[c]{Uni-modal /\\Single-task}} 
 & R  & D \& T & \cite{zhou2020mmw}    & Overview of sensing techniques in Autonomous Driving (AD) based on mmWave radar. \\ \cline{3-5} 
 & R  & D \& T & \cite{srivastav2023radars}    & Survey of Deep Learning (DL)-based automotive radar sensing including detection-based and occupancy-based tracking, as well as uncertainty reduction methods.    \\ \cline{3-5} 
 & R  & Various & \cite{kong2024survey}   & Overview of mmWave radar-based sensing technologies in autonomous driving, smart home, and industry.  \\ \cline{3-5} 
 & C  & D \& T & \cite{pal2021deep}    & Review on DL-based visual models for object detection and object tracking, considering the two tasks both separately and holistically.     \\ \cline{3-5} 
 & C  & D \& T & \cite{zhu2021detection}    &   Summary of datasets and benchmarks for drone-oriented object detection and tracking. \\ \cline{3-5}
 & L  & D \& T & \cite{guo2020deep}    & Conducts an exhaustive survey on DL-based 3D point cloud understanding, covering shape classification, object detection, tracking, and segmentation. \\ \cline{3-5}
 & L  & D \& T & \cite{hasan2022lidar}    & Provides analysis on LiDAR-based perception in terms of object detection, person tracking, and person property estimation.    \\ \cline{2-5}
 & L / C  & D & \cite{feng2021review}    & Summarizes the uncertainty estimation methods in probabilistic object detection, while mainly focusing on camera-only and LiDAR-only domains.\\ \cline{3-5}
 &  C & D & \cite{zou2023object}    &  Conducts a thorough review on the landmark object detectors, essential technologies, and acceleration schemes throughout 20-year history.  \\ \cline{3-5}
 & L / C & D & \cite{qian20223d}  & Develops a systemic taxonomy of representative object detectors, encompassing monocular/stereo image-based, point cloud-based and camera-LiDAR fusion-based methods. \\ \cline{3-5}
 & L / C & D & \cite{wang2023multi}    &  Provides an overview of multi-sensor fusion for 3D object detection, concentrating on LiDARs and cameras.  \\ \cline{3-5} 
 & L / C & D & \cite{mao20233d}    &  Presents an extensive survey on 3D object detection from diverse aspects, including input modalities, temporal sequences, and label-efficient methods. \\ \cline{3-5}    
 & Various & T & \cite{wang2019multi}    & Elaborates on the characteristics of various dominant sensors (i.e., radar, camera, LiDAR, ultrasonic, GPS/IMU), as well as the issues related to motion models and data association in object tracking.  \\ \cline{3-5}
 & C & T & \cite{marvasti2021deep}   & Makes a review concerning DL-based visual tracking and discusses the mainstream object tracking methods.  \\ \cline{3-5}
 & C  & T & \cite{luo2021multiple}    & Gives a representative tutorial on the essential aspects of a multiple object tracking system, covering problem formulation, component categorization, core principles, and evaluation metrics.  \\ \cline{3-5}
 &  R / L & T & \cite{granstrom2023tutorial}    &  Provide a systemic review on the topic of multiple extended object tracking, involving the prevalent strategies for shape estimation, data association, and track management.\\ \cline{3-5} 
 & R  & T & \cite{pearce2023multi}    &  Presents an overview of traditional, up-to-date, and future approaches for mmWave radar object tracking.     \\ \hline
\multirow{8}{*}{\makecell[c]{Multi-modal\\Fusion}} 
 & L \& C & D \& T & \cite{cui2021deep}         &   Explores LiDAR-camera fusion problems in terms of depth completion, object detection, semantic segmentation, object tracking, and sensor calibration.     
 \\ \cline{2-5}
 & L \& C  & D \& T & \cite{wu2022deep}   & Reviews a variety of LiDAR-camera fusion strategies in autonomous driving tasks like classification, detection, tracking, and segmentation.         \\ \cline{2-5}
 & Various & D \& T & \cite{ravindran2020multi}      & Summarizes the multi-object detection and tracking studies pertaining to there predominant sensors (i.e., camera, LiDAR and radar), yet with merely a handful of discussions on literature fusing camera and radar.     \\ \cline{2-5} 
 & Various & D & \cite{feng2020deep}    &  Systematically surveys the multi-modal perception problems in terms of object detection and semantic segmentation, while  totally disregarding the analysis on radar-camera fusion.      \\ \cline{2-5} 
&  Various & Various & \cite{wang2024multi}   & Extensively reviews multi-modal fusion sensing with mmWave radar and other modalities in human sensing applications, yet still falls short in an in-depth and systematic elaboration on radar-vision fusion.  \\ \hline

 \multirow{8}{*}{\makecell[c]{Radar-camera\\Fusion}}  
 & R \& C & D & \cite{wei2022mmwave}     &  Presents an overview of mmWave radar and RGB camera fusion for object detection under the Front View (FV), but neglects the object tracking task.   \\ \cline{2-5} 
 & R \& C & D & \cite{singh2023vision}   & Briefly summarizes  radar-vision fusion for BEV detection, yet still overlook the in-depth review on tracking.   \\ \cline{2-5} 
 & R \& C & D & \cite{zhou2022review}    &  Provides a tutorial on mmWave radar and camera fusion technology, but caters to conventional fusion strategies which tend to be obsolete.   \\ \cline{2-5} 
 & R \& C & D \& T & \cite{tang2021road}      &  Outlines on-road object detection and tracking approaches with image-oriented fusion strategies. Nonetheless, this work lacks comprehensiveness and applicability in the context of modern radar-camera fusion.  \\ \cline{2-5} 
 & R \& C & D & \cite{yao2023radar}      & Takes a representative review on radar-camera fusion schemes concerning object detection and semantic segmentation, neglecting to explore the issues involved with object tracking.     
 \\ \hline
 
 \multicolumn{4}{|c|}{This Paper}      & Structurally and comprehensively surveys the up-to-date literature 
 on radar-camera fusion-based object detection and tracking in a holistic way, as well as elaborately reviewing key techniques for radar-camera fusion, including sensor calibration, fusion modal representation, data alignment, and fusion operation.
\\ \hline
\multicolumn{5}{l}{\scriptsize  R: Radar, C: Camera, L: LiDAR, D: Object Detection, T: Object Tracking}\\
\end{tabular}
\end{table*}

\renewcommand\arraystretch{1.1}
\begin{table*}[t]
\centering
\caption{An intuitive comparison list of representative surveys in object detection and tracking}
\label{tab:Related_Surveys_on_Radar-camera_fusion}
\begin{tabular}{|m{1.3cm}<{\centering}|m{2.2cm}<{\centering}|m{1.0cm}<{\centering}|m{1.0cm}<{\centering}|m{1.2cm}<{\centering}|m{1.3cm}<{\centering}|m{1.3cm}<{\centering}|m{1.3cm}<{\centering}|m{1.4cm}<{\centering}|m{1.3cm}<{\centering}|}
\hline
\rowcolor{gray!40}  Ref & Sensing Modality & Domain & RCF Dataset & RCF Taxonomy & Object Detection & Object Tracking & Data Alignment  & Adaptive Fusion & Application \\ \hline
  \cite{zhou2020mmw}     & Radar &     AD     &    &                 &   \checkmark             &    \checkmark              &               &           &             \\ \hline
  \cite{srivastav2023radars}      & Radar  &   AD &                 &                    &   \checkmark             &   \checkmark               &          &      &                \\ \hline
  \cite{kong2024survey}       & Radar   &      Various &             &                     &       \checkmark             &     \checkmark                &               &                &   \checkmark  \\ \hline
 \cite{pal2021deep}      &  Camera         &     AD &              &                    &    \checkmark            &    \checkmark              &               &  &                \\ \hline
  \cite{zhu2021detection}   &     Camera           &   UAV   &                 &                    &  \checkmark               &    \checkmark               &               &             &      \\ \hline
    \cite{guo2020deep}      & LiDAR  &   AD  &       &                    & \checkmark               &  \checkmark                 &               &           &             \\ \hline
  \cite{hasan2022lidar}     & LiDAR  &   AD   &       &                    & \checkmark               &  \checkmark                 &               &           &            \\ \hline
  \cite{feng2021review}       &   LiDAR, Camera   &      AD    &             &                     &   \checkmark               &                  &               &                &  \\ \hline
 \cite{zou2023object}       &  Camera     &    AD &       &                    &   \checkmark        &                  &           &  &                 \\ \hline
  \cite{qian20223d}    &   LiDAR, Camera             &      AD   &             &                    &      \checkmark           &                  &               &             &   \\ \hline
    \cite{wang2023multi}     &   LiDAR, Camera  &    AD  &      &                    &     \checkmark            &                  &               &           &            \\ \hline
  \cite{mao20233d}      & LiDAR, Camera  &           AD &          &                    &     \checkmark            &                  &          &      &               \\ \hline
  \cite{wang2019multi}     & Various        &  AD &                 &                     &                &       \checkmark            &               &                &   \\ \hline
 \cite{marvasti2021deep}       &  Camera         & AD  &                  &                    &                &  \checkmark                 &               &  &                 \\ \hline
  \cite{luo2021multiple}   &  Camera         &   AD   &                &                    &                &  \checkmark                 &               &  &               \\ \hline
    \cite{granstrom2023tutorial}     & Radar, LiDAR   &  AD  &                 &                    &                &  \checkmark                 &               &  &                 \\ \hline
  \cite{pearce2023multi}     & Radar  &       AD  &  &                    &                &     \checkmark              &          &      &               \\ \hline
  \cite{cui2021deep}       &    LiDAR \& Camera   &     AD   &              &                     &      \checkmark            &   \checkmark                 &               &                &   \\ \hline
 \cite{wu2022deep}      &    LiDAR \& Camera   &  AD &                 &                     &      \checkmark            &   \checkmark                 &               &                &   \\ \hline
  \cite{ravindran2020multi}   &   Various             &   AD  &                &                    &   \checkmark              &       \checkmark            &               &             &   \\ \hline
  \cite{feng2020deep}      &  Various  & AD  &          &                    &     \checkmark            &                  &               &     \checkmark       &            \\ \hline
  \cite{wang2024multi}      & Various  &       HS   &            &                    &     \checkmark            &   \checkmark                &          &      &   \checkmark            \\ \hline
  \cite{wei2022mmwave}       & Radar \& Camera  &     AD  &       &                     &  \checkmark       &                  &               &                &  \\ \hline
 \cite{singh2023vision}     &  Radar \& Camera         & AD &                   &                    &  \checkmark               &                  &               &  &                 \\ \hline
  \cite{zhou2022review}    & Radar \& Camera               &  AD &                 &                    &  \checkmark               &                 &               &             &   \\ \hline
   \cite{tang2021road}     & Radar \& Camera               &  AD &                  &                    &  \checkmark               &         \checkmark        &               &             &   \\ \hline
  \cite{yao2023radar}   & Radar \& Camera               & AD  &   \checkmark                &    \checkmark                &  \checkmark              &                  &               &             &    \\ \hline

This Paper &    Radar \& Camera          &   Various   &   \checkmark                 &    \checkmark               &    \checkmark              &       \checkmark             &       \checkmark            &         \checkmark       &   \checkmark  \\ \hline
\multicolumn{8}{l}{\scriptsize  RCF: Radar-Camera Fusion, AD: Autonomous Driving, UAV: Unmanned Aerial Vehicle, HS: Human Sensing}\\
\end{tabular}
\end{table*}

\subsection{Related Surveys}
So far, there has not been any tutorial survey specifically addressing radar-camera fusion issues for object detection and tracking. On the flip side, there have been reviews focusing on particular aspects, which are listed in Table \ref{tab:An overview of selected surveys} and Table \ref{tab:Related_Surveys_on_Radar-camera_fusion}. We categorize these existing reviews into three groups, and delineate their disparities from our work.

In object detection and tracking areas, most reviews concentrate on unimodal perception (e.g., radar-only \cite{zhou2020mmw,srivastav2023radars,kong2024survey}, camera-only \cite{pal2021deep,zhu2021detection}, LiDAR-only \cite{guo2020deep,hasan2022lidar}) or single task (e.g., object detection \cite{feng2021review,zou2023object,qian20223d,wang2023multi,mao20233d}, object tracking \cite{wang2019multi,marvasti2021deep,luo2021multiple,granstrom2023tutorial,pearce2023multi}). 
Additionally, several works highlight emerging techniques that have been introduced to multimodal object detection and tracking, mainly from the perspectives of LiDAR-camera fusion \cite{cui2021deep,wu2022deep}, or a wider range of modalities including cameras, LiDARs, radars and other sensor types \cite{feng2020deep,ravindran2020multi, wang2024multi}. 
These surveys provide valuable insights about multimodal fusion perception in specific applications, yet primarily emphasize on LiDAR and camera fusion. For example, the authors in \cite{feng2020deep} systematically survey the multi-modal perception problems in terms of object detection and semantic segmentation, while omitting the analysis on radar-camera fusion. 
Besides, the review \cite{ravindran2020multi} summarizes the multi-object detection and tracking studies pertaining to there predominant sensors (i.e., camera, LiDAR and radar), yet with merely a handful of discussions on literature fusing camera and radar.
Moreover, the authors in \cite{wang2024multi} present an extensive survey on multi-modal fusion sensing with mmWave radar and other modalities, covering a broad range of modalities and perception tasks. Nonetheless, the review \cite{wang2024multi} still lacks an in-depth and systematic elaboration on radar-vision fusion, especially in data alignment and fusion operation.

More recently, a couple of surveys have delved into radar-camera fusion \cite{wei2022mmwave,singh2023vision,zhou2022review,tang2021road,yao2023radar}. Specifically, the authors in \cite{wei2022mmwave} present an overview of mmWave radar and RGB camera fusion for object detection under Front View (FV) coordinates, but do not cover the object tracking task.
In contrast, the authors in \cite{singh2023vision} briefly summarize the radar-vision fusion for Bird’s-Eye View (BEV) detection, yet still overlook
the in-depth review on target tracking.
In addition, the authors in \cite{zhou2022review} provide a tutorial on mmWave radar and camera fusion technology, but cater to conventional fusion strategies which tend to be obsolete.
Moreover, the survey \cite{tang2021road} outlines on-road object detection and tracking approaches separately, as well as analyzes image-oriented fusion strategies. Nonetheless, this work lacks comprehensiveness and applicability in the context of modern radar-camera fusion.
Additionally, the authors in \cite{yao2023radar} extend \cite{feng2020deep} by conducting a representative review on the radar-camera fusion schemes in autonomous driving. But equally, they concern the tasks of object detection and semantic segmentation, neglecting to explore the issues involved with object tracking. 

In summary, nearly the full majority of surveys on radar-camera fusion focus on one specific application. 
Considering the unique features of sensing applications such as robotics and drones, the perception requirements are application-driven and diverse. As such, it is conducive and imperative to build and tackle radar-camera fusion problems according to the actual sensing requirements. On the other hand, understanding object detection and tracking in a holistic way instead of from separate perspectives enables researchers to get more insights into how fundamental perception technology evolves and promotes the fast-increasing applications. 
More importantly, most existing reviews fall short on shedding light on a thorough and in-depth elaboration of the extremely core technologies (e.g., modal representation, data alignment, and fusion operation) involved in radar-camera fusion.

\begin{figure*}[h]
	\centering
	\includegraphics[width=0.99\textwidth]{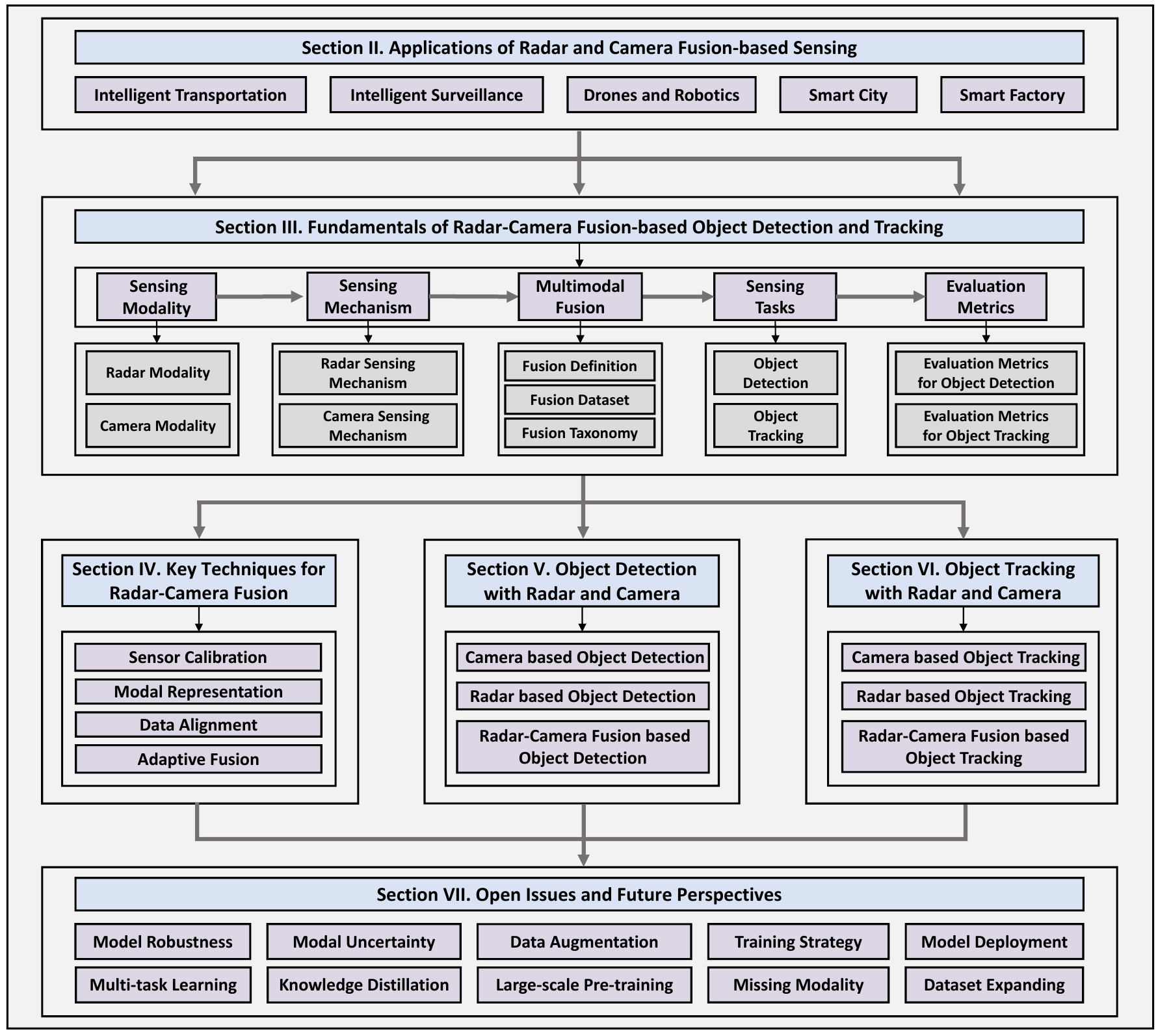}\\
	\caption{The organization and taxonomy of the article.}\label{fig:detailed diagram}
\end{figure*}

\subsection{Contributions}
To the best of our knowledge, there is no in-depth survey providing an extensive discussion on radar-camera fusion for object detection and tracking with 
emerging strategies such as fusion modal representation, data alignment, and adaptive fusion operation.
Through a systemic analysis, this study aims to elucidate the significance and potential of radar-camera fusion perception in advancing the capabilities of intelligent systems for real-world applications. 
Instead of enumerating individual works, we elaborately summarize existing literature based on various aspects, both structurally and hierarchically. In each aspect, key issues are categorized into different groups, whilst each group is thoroughly discussed regarding its principles, advancements, and limitations.
The contributions of this paper are summarized as follows:

\begin{itemize}
  \item  We identify the research gap among the existing reviews on radar-camera fusion perception by rigorously analyzing the representative surveys listed in Table \ref{tab:An overview of selected surveys} and Table \ref{tab:Related_Surveys_on_Radar-camera_fusion}, which highlight the demand for conducting a comprehensive survey on radar-camera fusion technologies in terms of object detection and tracking. 
  \item	We expatiate on the principles, methodologies, and applications for radar-camera fusion perception, as well as delve into the emerging strategies for modal representation, data alignment and fusion operation.
  \item We present a thorough classification for the research topics of object detection and tracking pertaining to radar and camera, and provide an up-to-date (2019-2024) summary of radar-camera fusion datasets and algorithms.
  \item We elaborately discuss the emerging perspectives in the direction of  radar-camera fusion sensing and highlight the future development trends.
\end{itemize}

For the reader’s convenience, Fig. \ref{fig:detailed diagram} presents the organizational structure of the article. In Section II, we showcase a selection of promising radar-camera fusion applications,
while Section III provides an introduction to the background
and essentials of radar and camera perception technologies.
Section IV outlines the key techniques pertinent to radar-camera fusion challenges. Moreover, Section V conducts a literature review on radar and camera sensing concerning object detection issues, while Section VI presents an overview of object tracking problems utilizing radar and camera data. The open issues and future perspectives are discussed in Section VII and conclusions are drawn in Section VIII.

\section{Applications of Radar-Camera Fusion} 
This section presents a concise overview of radar-camera fusion perception applications, ranging from intelligent transportation to smart factory.

\subsection{Intelligent Transportation} 
\begin{figure}[h]
    \centering
    \includegraphics[width=1\linewidth]{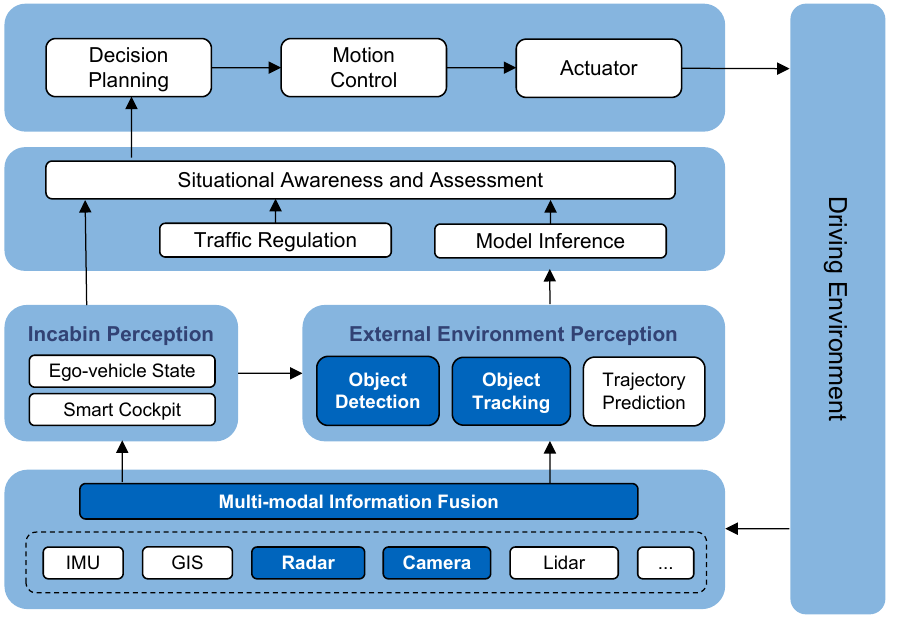}
    \caption{An exemplary application of multimodal fusion perception in autonomous driving.}
    \label{fig:exemplary}
\end{figure}

As the pioneering realm of radar-camera fusion, intelligent transportation has experienced profound advancements with the potential to enhance roadway safety and yield economic benefits \cite{shi2022efficient,shi2021road,liu2024towards}.
Taking the promising autonomous driving as an example, Fig.~\ref{fig:exemplary} illustrates the modular structure of multimodal fusion-based perception.
Notably, the perception module serves as the cornerstone of the system architecture. It realizes the sensing and understanding of tangible physical reality through sensor hardware and embedded software, and provides a solid foundation for subsequent decision planning and motion control modules.

Specifically, the authors in \cite{wang2016road} introduce a representative model-based fusion pipeline aimed at attaining the ideal equilibrium between detection precision and computational efficacy. Firstly, radar identifies vehicle targets and sends the location and 2D size of the  Region of Interest (RoI) to the corresponding camera image. Subsequently, an image processing component creates a boundary in the image plane based on the RoI information and utilizes the active contour algorithm to detect vehicles within the defined boundary.
Besides, the author in \cite{chadwick2019distant} present a seminal DL-based radar-camera fusion framework for distant vehicle detection. They also create a first-of-its-kind proprietary dataset which covers both radar and camera data.
Moreover, the authors in \cite{liu2021robust} develop a fusion scheme with radar as the predominant sensor and camera as the auxiliary to achieve reliable object tracking in inclement weather. The proposed scheme undergoes extensive on-road testing and the experimental results validate its effectiveness under severe conditions.

The aforementioned approaches are mainly based on a single ego-vehicle. Aligning data from multiple sensors within a single vehicle is straightforward due to the congruent perception spaces of the data.
However, single-vehicle perception inevitably encounters two main challenges: limited perception of obscured targets and sparse visibility of distant objects. 
Consequently, an increasing number of studies turn to multi-agent collaborative frameworks \cite{zheng2023autofed,alkhateeb2023deepsense,cheng2023intelligent,xiang2023multi}, where an ego-vehicle communicates with other nearby agents, e.g. vehicles or infrastructures, exploiting Vehicle-to-Vehicle (V2V) or Vehicle-to-Infrastructure (V2I) information to enhance perception accuracy and range. 
To be specific, the authors in \cite{cheng2023intelligent} pioneer a structured framework for intelligent integration of multi-modal perception and communication.
AutoFed \cite{zheng2023autofed} takes the first step towards enabling autonomous vehicles to train object detection models with a heterogeneity-aware federated learning framework.
Additionally, the authors in \cite{alkhateeb2023deepsense} create an innovative dataset dubbed DeepSense 6G, the first extensive real-world multi-modal sensing and communication dataset. This dataset holds potential for a broad spectrum of applications in the intersection of sensing, communication, and positioning.
Furthermore, the authors in \cite{xiang2023multi} scrutinize and outline the pioneering investigation and utilization of collaborative perception in intelligent transportation systems using multi-sensor fusion.

\subsection{Intelligent Surveillance}
A second significant application area of radar-camera fusion emerges in security surveillance, where detection and tracking are two foremost considerations \cite{ibrahim2016comprehensive}.

The authors in \cite{zyczkowski2011integrated} implement an integrated radar-camera security system for object detection, tracking, identification, localization and alarm notifications. The experimental results demonstrate the potential to configure the system as a stationary, mobile or portable equipment.
Besides, the authors in \cite{iepure2021novel} set up an automated surveillance system by fusing thermal and optical cameras with mmWave radar. The devised system demonstrates favorable tracking performance with commercially accessible surveillance systems.
Moreover, the authors in \cite{vandana2022intruder} advance an intruder alarm equipment combining radar and camera sensors to detect and track intruders within a confined area.
The authors in \cite{fu2023key} present a radar and video fusion-based alarm system for perimeter intrusion. They provide a detailed overview of video surveillance and perimeter alarm systems, including mathematical modeling, essential technologies, as well as the future development trajectory of intelligent security systems.
Additionally, the authors in \cite{lee2023efficient} develop an edge computing system utilizing radar, camera, and LiDAR sensors for detecting intruders across a broader coverage area. The system is deployed on an outdoor single-row barbed wire fence, with an intruder entering the detection area to assess its detection capabilities.

Apart from the overland surveillance, radar-camera fusion is
indispensable for intelligent maritime to enhance safety and efficiency in autonomous vessel situational awareness \cite{thombre2020sensors}. 
For instance, the authors in \cite{cormack2019joint} demonstrate an efficient strategy for performing joint sensor registration and fusion of infrared camera and scanning radar in a maritime environment.
The authors in \cite{kim2021robust} advance a reliable data association scheme to fuse camera and marine radar measurements for autonomous detection of surface vessels. The data alignment between camera and radar images is optimized through a pair of geometric parameters, enabling determination of the positions of all matched object features in the absence of precise calibration. 
Besides, the authors in \cite{clunie2021development} develop an obstacle detection and tracking  platform for Autonomous Surface Vehicles (ASVs) utilizing camera, LiDAR, and marine radar. The platform is validated through live experiments, demonstrating its capability to detect and track obstacles up to 450 meters away.
Moreover, the authors in \cite{liu2021hybrid} design a coarse-to-fine vessel recognition approach by fusing closed-circuit television and marine radar.

\subsection{Drones and Robotics}
A third domain of integrated radar-camera perception applications lies in  drones and robotics. 

The authors in \cite{yu2020autonomous} devise a Unmanned Aerial Vehicle (UAV) obstacle avoidance strategy based on the fusion of mmWave radar and monocular camera. First, a monocular camera is employed to identify obstacles and extract key information, including obstacle outlines and center point positions. Subsequently, the preliminary information is integrated with data from an mmWave radar to determine the spatial positions of obstacles.
Similarly, the authors in \cite{huang2021improved} develop a radar and image fusion based obstacle avoidance algorithm tailored to the flight requirements of plant protection for UAVs, emphasizing high efficiency, smoothness, and continuity.
In addition, the authors in \cite{chang2021radar} leverage radar-camera fusion to detect power lines as part of implementing a UAV autonomous navigation system, which applies to scenarios involving operations near power lines.
Besides, the authors in \cite{malle2022onboard} also design an mmWave-RGB fusion architecture for multi-cable power line detection and pose estimation, in that the power line perception system aids in aerial manipulation and infrastructure inspection.
 
Moreover, the authors in \cite{wu2023precise} establish a novel self-tuning fusion framework for UAV positioning by leveraging the Doppler motion feature from mmWave radar data and RGB feature from visual images. They also build a functional UAV positioning platform and create two mmWave-RGB datasets with six multi-scenario radar-camera sequences. Extensive experiments validate the fusion framework surpasses benchmark approaches in terms of accuracy while sustaining real-time performance.
To detect and track small UAVs, the authors in \cite{huang2023radar} build a deep semantic association network to establish connections between camera detection and radar point cloud. Several flight trials are conducted to collect a multi-modal dataset, and the testing results demonstrate the superior performance of the proposed fusion network compared to its single-modality counterpart.
Besides, the authors in \cite{de2023drone} perform an elementary  experimental investigation into tracking small UAVs, highlighting the potential benefits of fusing high resolution camera and staring radar in practical operational scenes. The results are obtained from challenging flight trajectories using a commercial UAV platform, covering distances of up to 7 km.
In contrast to the aforementioned work utilizing RGB cameras, the authors in \cite{safa2023fusing} present the first continual learning SLAM architecture for UAV navigation, fusing an event camera and a radar with Spiking Neural Networks (SNNs). Unlike conventional offline-training method, the devised method circumvents the pre-training phase. Instead, it depends on the continual and unsupervised adaptation of the SNN weights through spike timing dependent plasticity learning.

\subsection{Smart City}
In terms of smart city, multifarious resources are combined and streamlined to facilitate the management of public services. Typically, it encompasses several well-known domains, such as smart utilities, smart home, smart health, etc.

The authors in \cite{cheng2021robust} open the possibility of research on radar-camera fusion schemes to clean floating wastes with Unmanned Surface Vehicles (USVs). Besides, they provide a pioneering dataset dubbed FloW \cite{cheng2021flow} for detecting floating debris, acquired from the perspective of USVs in actual inland water environments under diverse circumstances.
Subsequently, the authors in \cite{yao2024waterscenes} extend FloW to WaterScenes, a comprehensive dataset covering more sensing modalities (e.g., 4D radar, Camera, IMU), tasks (e.g., object detection, tracking, segmentation), adverse
weather conditions (e.g., overcast, rainy, snowy), and target caterories (pier, boat, buoy), as well as a radar-vision fusion framework for the detection benchmark on WaterScenes.
Moreover, the authors in \cite{zhu2023millimeter} develop an approach for detecting tiny objects on water surfaces by associating mmWave radar with RGB information utilizing a metric learning model, which is able to accommodate variations in extrinsic parameters to some extent.

In addition, the authors in \cite{cao2022cross} build a inventive RGB-mmWave fusion architecture for cross-modal human re-identification. Similarly, they employ a robust cross-modal deep metric learning model to address the interference arising from the asynchronicity between the radar and camera.
Besides, the authors in \cite{liao2020left} introduce a novel system for detecting and tracking left-behind humans inside a vehicle by combing visual and microwave radar data.
The authors in \cite{sun2023heterogeneous} design an intelligent unmanned parking meter system by fusing mmWave heatmaps and camera images. Experimental testings exhibit the heterogeneous fusion scheme achieves an average accuracy of $99.33\%$.
Additionally, the authors in \cite{li2022pedestrian,wang2023end} construct a first-of-its-kind radar-vision fusion framework by utilizing the radar cross-section (RCS) information as a discriminative feature to differentiate between living pedestrians and portrait billboards. 
The authors in \cite{akbari2023new} design an intelligent door system which integrates facial detection and recognition techniques based on camera and mmWave radar sensors.
Furthermore, the authors in \cite{wang2023vital} present a radar-camera fusion scheme for monitoring vital signs (i.e., heartbeat and breathing frequency) in the presence of body movement. The core components involve the fusion of video camera and mmWave radar to identify targets, employing beamforming to mitigate interference and isolate the vital sign from multiple range bins, as well as extracting weak vital-sign signals. The scheme operates effectively in various dynamic environments without requiring scene-specific training.

\subsection{Smart Factory}
In the context of industrial manufacturing, collaboration between operators and robotic entities is paramount and mandatory in factory environments, significantly enhancing productivity and efficiency \cite{ tang2022computational}.

Considering the anonymous monitoring and workspace safety during human-robot cooperation, the authors in \cite{kianoush2020multisensory} present an implementation of a multi-modal fusion-based edge-cloud platform for continuous and accurate operator perception. The platform fuses heterogeneous radio signals collected from various interconnected facilities, including a WiFi, an imaging camera, a network of radars and infrared sensors.
Besides, the authors quantify the effects of detection accuracy and latency on safe human-robot cooperation, and determine robustness as the protective separation distance between operators and robots. They also conduct three case studies, covering operator counting, motion detection, and operator-robot co-presence monitoring.
Similarly, the authors in \cite{minora2023radar} formulate an industrial multi-source fusion framework that analyzes diverse sensor information to enable anonymous detection of workers within a cobot environment. The system incorporates sensing and industrial IoT devices (involving infrared array sensors, networked sub-THz radars, and a 100 GHz imaging camera)  to monitor the boundary-free collaborative area shared by humans and robots.
In addition, the authors in \cite{zoghlami2021tof} devise a worker detection and positioning system for automated production based on the fusion of a Time of Flight (ToF) camera and a 60 GHz radar sensor. The platform is also capable of estimating the direction of movement along with the instantaneous speed of each person.
Moreover, the authors in \cite{liu2023multimodal} introduce a radar-camera fusion based dynamic hand gesture recognition for intelligent robot control.

\section{Background and Fundamentals of Radar-Camera Fusion-based Perception} 
\label{sec:Background and Fundamentals of RCF-based Perception}
In this section, we first provide a succinct description of sensing modalities and their sensing mechanisms broadly employed in multimodal fusion-based perception, focusing on radar and camera sensors. Then, we elaborately delve into the critical issues for multimodal fusion, including fusion taxonomies and fusion methodologies. Subsequently, we expatiate on elementary definitions and illustrative methods for two essential and pivotal sensing tasks, i.e., object detection and object tracking. Finally, we methodically discuss the prevalent evaluation metrics for these sensing tasks.

\subsection{Sensing Modality}
This subsection aims to furnish the reader with a basic comprehension of the technical concepts as well as pinpoint the key considerations of radar and camera systems.

\subsubsection{Radar}
The terminology RADAR originates as an acronym representing RAdio Detection And Ranging. Radar sensors have evolved tremendously since their initial stages when their capabilities were confined to object recognition and distance measurement. In contrast, advanced radars are also applicable for object tracking, imaging, and classification.

The intuition behind radar sensing is that the radar system emits radio-frequency (RF) electromagnetic (EM) waves toward a specified Region of Interest (RoI) and captures, as well as identifies these EM waves upon reflected from objects within the RoI. 
The EM waves emitted and received by a radar engage with various elements, namely, the antenna, the atmosphere, and the target. The pertinent physical principles dictating these interactions include diffraction (antenna), attenuation, refraction, depolarization (atmosphere), and reflection (target) \cite{richards2010principles}. 
Specially, scattering, as another common term, refers to the reflection of the incident EM wave from the surface of a target. 
The study of scattering phenomenology involves the quantification of a crucial parameter known as \textit{Radar Cross Section} (RCS). 
The RCS quantifies not just the portion of the EM wave reflected by the target but also accounts for the intercepted portion and the fraction directed back towards the radar's receiver.
It is noteworthy that RCS is prone to being confused with the term \textit{intensity}, which is synonymous with power density, indicating the power per unit area of the propagating wave. 

In radar technology, the received signals from the targets are inevitably accompanied by interference, manifesting in four distinct forms: 1) intrinsic and extrinsic electronic noise; 2) unwanted EM waves from non-target objects, a.k.a. \textit{clutter}; 3) unintentional EM waves generated by other human-made sources, viz., EM interference; and 4) deliberate jamming emanating from an electronic countermeasures system, presented as noise or spurious targets. 

\subsubsection{Camera}
A camera is a  device employed for recording and storing visual information, achieved either through digital means using an electronic image sensor or through chemical processes involving light-sensitive materials like photographic film. 
Generally, cameras can facilely acquire the appearance features of objects, encompassing shape, texture, and color. 
Nevertheless, most cameras operate as passive sensors, meaning that image formation relies on incident light exposure. Adverse conditions, such as inadequate illumination, inclement weather, and lens contamination by water droplets or dust, tend to deteriorate imaging results.

To date, the vast majority of visual perception systems leverage images obtained from \textbf{RGB cameras} as sensing primitives.
The acronym RGB corresponds to the color model where red, green, and blue primary colors of light are combined to produce various perceptible colors. An RGB camera represents an image as a pixel grid, where each pixel holds values for the red, green, and blue color channels. This camera operates within the visible light spectrum, typically ranging from 400 to 700 nanometer (nm).
Besides, an \textbf{infrared (IR) camera}, a.k.a. thermographic camera or thermal imaging camera, utilizes IR radiation to generate images. In contrast to the 400–700 nm range of the RGB camera, infrared cameras are responsive to wavelengths ranging from approximately 1,000 nm to 14,000 nm.
In addition, as another highly preferred camera in specific embedded vision applications, an \textbf{RGBD camera} delivers real-time outputs comprising both color (RGB) and depth (D) information. Herein, depth data is obtained through a depth map generated by a 3D depth sensor, such as a stereo sensor or time-of-flight sensor. These cameras seamlessly merge pixel-to-pixel RGB data with depth information, presenting both aspects in a unified single frame.
Moreover, as an emerging bionic camera, an \textbf{event camera} produces an event image representation based on pixel-level changes in brightness. Possessing sub-millisecond latency, high dynamic range, and resilience to motion blur, event cameras offer remarkably potential for object detection and tracking in real-time situations.

\subsection{Sensing Mechanism}
In this subsection, we delve into the fundamental principles underlying radar and camera technology. Comprehending these physical mechanisms is essential for understanding the capabilities of sensor perception.

\subsubsection{Camera Sensing Mechanism}
In this part, we provide a brief description on the camera sensing mechanisms. We take monocular cameras as an illustrative example in view of their representativeness.

A typical monocular camera comprises a lens, an image sensor, an image signal processor, and an Input/Output (I/O) interface \cite{shi2022toward}. The lens converges light waves reflected from objects onto the image sensor. Subsequently, the image sensor (e.g., CMOS sensor) transforms the light into electrical signals and further converts these signals into digital values using an Analog-to-Digital Converter (ADC). Then, the image signal processor conducts post-processing and transforms the digital values into a pixel image data format. Lastly, the images are transmitted and displayed through the I/O interface.
As a contrast, a stereo camera comprises two or more lenses, each with a separate image sensor or film frame. This configuration enables the camera to mimic human binocular vision, facilitating the capture of three-dimensional images.

\subsubsection{Radar Sensing Mechanism}
\begin{figure}[H]
    \centering
    \includegraphics[width=1\linewidth]{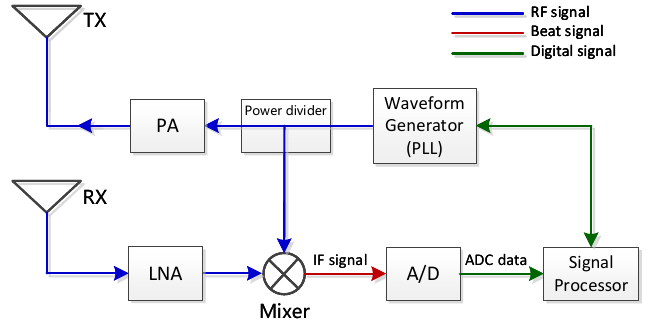}
    \caption{Major elements in the radar working pipeline.}
    \label{fig:radar working}
\end{figure}

Fig.~\ref{fig:radar working} depicts the key components in the radar working pipeline. 
Albeit the specifics of different radar systems vary, the essential subsystems comprises a waveform generator (a.k.a. synthesizer), TX/RX antennas, and signal processors.
Initially, the synthesizer, often based on a programmable phase-locked loop (PLL), produces a radiated waveform, which undergoes amplification via a power amplifier (PA) before being transmitted through transmitter antennas.
Next, receiver antennas capture the reflected waveform from the target, which is then processed through a low-noise amplifier (LNA). 
After that, an RF mixer combines the RX and TX signals to generate an Intermediate Frequency (IF) signal (viz., beat signal), which is subsequently amplified, filtered, and digitized by the ADC.
Typically, a radar system incorporates multiple TX and RX antennas, bringing about several IF signals. These IF signals implicitly contain crucial information about the target, e.g., range, Doppler velocity, and azimuth angle, which a signal processor calculates using predefined signal processing algorithms. 

In terms of the method to generate the radiated waveform, radar systems can be categorized into two main types, namely continuous-wave (CW) radars and pulse radars \cite{chen2021octopus,zheng2020v2ifi,zheng2021siwa,zheng2022catch}. 
For the CW radars, the radiated waveform is continuous, while for the pulse radars, the signal is emitted intermittently over a short duration \cite{zhang2022can}. Both types of radar can employ modulated or unmodulated signals, leading to additional classifications based on the modulation approach. In terms of the technical literature on object detection and tracking, the mainstream types of radar are frequency-modulated continuous-wave (FMCW) radar and phase-modulated continuous-wave (PMCW) radar. 
An FMCW waveform, a.k.a. a \textit{chirp}, is a continuous wave signal characterized by a linear increase in frequency over time. 
When a transmitted chirp reflects off targets in the environment and is received, there is a time delay between the transmitted and received chirps. This delay can be estimated by mixing the two chirps and measuring the IF of the mixed signal. Subsequently, the range between the radar and the target can be determined.
In general, chirps are sent back-to-back within a short period known as a \textit{frame}. The duration of one chirp is typically called the \textit{fast time}, while the duration across multiple chirps is termed the \textit{slow time}. The target motion induces phase variations between adjacent chirps in a frame, facilitating the radial velocity estimation of the target.
Besides, the target's angle is determined using an antenna array, where different angles of the target cause distinct phase differences across the antennas. 
In PMCW radars, a single-frequency signal is employed as the carrier wave at the transmission end. This carrier wave undergoes phase modulation using encoding methods such as binary coding before being transmitted, received, and processed to extract target distance, velocity, and angle information. 
Unlike FMCW radars that leverage the frequency difference between transmitted and received signals as an indirect way for range detection, PMCW radars directly employ digital code correlation. This obviates the requirement for strict linearity in frequency ramping over time, which is a significant challenge in analog circuit. Therefore, PMCW radars demonstrate considerably lower error in range detection.

\begin{table*}
\scriptsize
\centering
\caption{An overview of radar-related multi-modal fusion datasets}
\label{tab:radar related datasets}
\setlength\tabcolsep{3pt}
\begin{tabular}{c|c|cccc|cc}
\hline \textbf{Radar} & \textbf{Dataset} & \textbf{LiDAR} & \textbf{Camera} & \begin{tabular}{l} 
\textbf{Radar Type} \end{tabular}  & \begin{tabular}{l} 
\textbf{Semantic Labels} \\
\textbf{Object Categories}
\end{tabular} & \textbf{Environment} & \begin{tabular}{l} 
\textbf{Inclement} \\
\textbf{Weather}
\end{tabular} \\
\hline
\specialrule{0.1em}{1pt}{1pt}
\multirow{4}{*}{\begin{tabular}{l} 
Scanning \\
Radar
\end{tabular}} 
& \begin{tabular}{l} 
Oxford Radar  \cite{barnes2020oxford}
\end{tabular} & \ding{51} & Stereo/Mono & Navtech CTS350-X, 79 GHz & N/A & Dense Urban & Rain, Fog \\
& \begin{tabular}{l} 
Boreas \cite{burnett2023boreas}
\end{tabular} & \ding{51} & Mono & Navtech CIR304-H, 79 GHz & 2D Boxes (4 Categories) & Sparse Urban & \begin{tabular}{l} 
Rain, Snow, Fog
\end{tabular} \\
& MulRan \cite{kim2020mulran} & \ding{51} & \ding{55} & Navtech CIR204-H, 79 GHz  & N/A & Mixed Urban & N/A \\
& RADIATE \cite{sheeny2021radiate} & \ding{51} & Stereo & Navtech CTS350-X, 79 GHz &  \begin{tabular}{l} 
2D Boxes (8 Categories)
\end{tabular} & Mixed Urban & \begin{tabular}{l} 
Rain, Snow
\end{tabular} \\
\hline

\multirow{8}{*}{\begin{tabular}{l} 
SoC \\
Radar
\end{tabular}} 
& nuScenes \cite{caesar2020nuscenes} & \ding{51} & 6$\times$Mono & Continental ARS408-21, 76$\sim$77GHz & \begin{tabular}{l} 
3D Boxes (23 Categories)
\end{tabular} & \begin{tabular}{l} 
Mixed Urban, Rural 
\end{tabular} & Rain \\

& \begin{tabular}{l} 
RadarScenes \cite{schumann2021radarscenes}
\end{tabular} & \ding{55} & Mono & Continental ARS404, 76$\sim$77 GHz  & \begin{tabular}{l} 
Point-wise (11 Categories)
\end{tabular} & \begin{tabular}{l} 
Mixed Urban
\end{tabular} & Rain, Fog \\
& Astyx \cite{meyer2019automotive} & \ding{51} & Mono & Astyx 6455 HiRes, 77GHz  & 3D Boxes (7 Categories) & Sparse Urban & N/A \\
& RaDICaL \cite{lim2021radical}  & \ding{51} & Stereo & TI IWR1443, 76$\sim$81 GHz  & 2D Boxes (2 Categories) & Indoors, Sparse Urban  & N/A \\
& CRUW  \cite{wang2021rodnet}  & \ding{55} & Stereo & TI FMCW radar, 77 GHz  & 2D Positions (3 Categories) &  Mixed Urban, Campus  & N/A \\
& \begin{tabular}{l} 
K-Radar \cite{paek2022k}
\end{tabular} & \ding{51} & Stereo & RETINA-4ST, 77$\sim$81 GHz  & \begin{tabular}{l} 
3D Boxes (5 Categories)
\end{tabular} & \begin{tabular}{l} 
Mixed Urban, Rural 
\end{tabular} & \begin{tabular}{l} 
Fog, Rain, Snow  
\end{tabular} \\
& \begin{tabular}{l} 
VoD \cite{palffy2022multi}
\end{tabular} & \ding{51} & Stereo &  ZF FRGen21, 77GHz & \begin{tabular}{l} 
3D Boxes (13 Categories)
\end{tabular} & Mixed Urban & N/A \\
& Dual Radar \cite{zhang2023dual} & \ding{51} & Mono &  \begin{tabular}{l} 
ARS548 RDI,  76$\sim$77 GHz \\
Arbe Phoenix,  76$\sim$77 GHz
\end{tabular} & 3D Boxes (5 Categories) & Mixed Urban & Rain, overcast \\
\hline
\end{tabular}
\end{table*}

In the field of detection and tracking, another prevailing taxonomy roughly divides radars into scanning radar and System-on-Chip (SoC) radar. 
Specifically, scanning radar consists of a physically rotating radar sensor and provides a highly accurate polar image of the surroundings by measuring target distances at various angles. Target distances are obtained by performing an FFT on ADC samples for each angle and identifying the relative peaks. Unlike LiDAR sensors, scanning radar cannot offer elevation angle of returned targets. Additionally, scanning radar generally is incapable of obtaining velocity information, as it sends and receives only one pulse per antenna angle, rather than multiple pulses needed for velocity calculation.
In contrast, SoC radar integrates processing units into a limited number of chips that are either directly mounted on patch antennas or embedded in a printed circuit board. Due to this integration, SoC radars typically demonstrate lower weight and power consumption compared to the scanning radar. The accuracy and resolution of SoC radar depend on the specific antenna array used and the processing methods employed by the manufacturer.

Unlike cameras that rely on light waves, radar sensors radiate radio waves with longer wavelengths. The inherent characteristic enables radar waves to penetrate most environmental impurities, such as rain, fog, and snow. As a result, radars are capable of effectively working even in harsh circumstances, rendering them highly reliable across diverse real-world applications.
Furthermore, radar waves can penetrate materials like walls and vegetation, reflecting off hidden targets. This capability makes radars able to identify targets positioned around corners or blocked by obstacles.

Although radar sensors possess irreplaceable advantages, it is crucial to acknowledge their inherent deficiencies, predominantly the limited angular resolution and the susceptibility to interference. First, the inferior angular resolution poses challenges in distinguishing adjacent objects, potentially leading to a high rate of omissions. Besides, the sparse radar data struggles to outline object contours as well as extract detailed geometric information. 
Second, radar signals are vulnerable to interference from various sources, especially the multipath effect and clutter, which adversely contaminate the radar measurements, bringing about false alarms. 
Despite these limitations, radar sensors remain valuable tools in various applications, and ongoing efforts are focused on improving their performance and overcoming these challenges by multi-modal fusion.
\begin{figure}[htpb]
	\centering
	\includegraphics[width=0.48\textwidth]{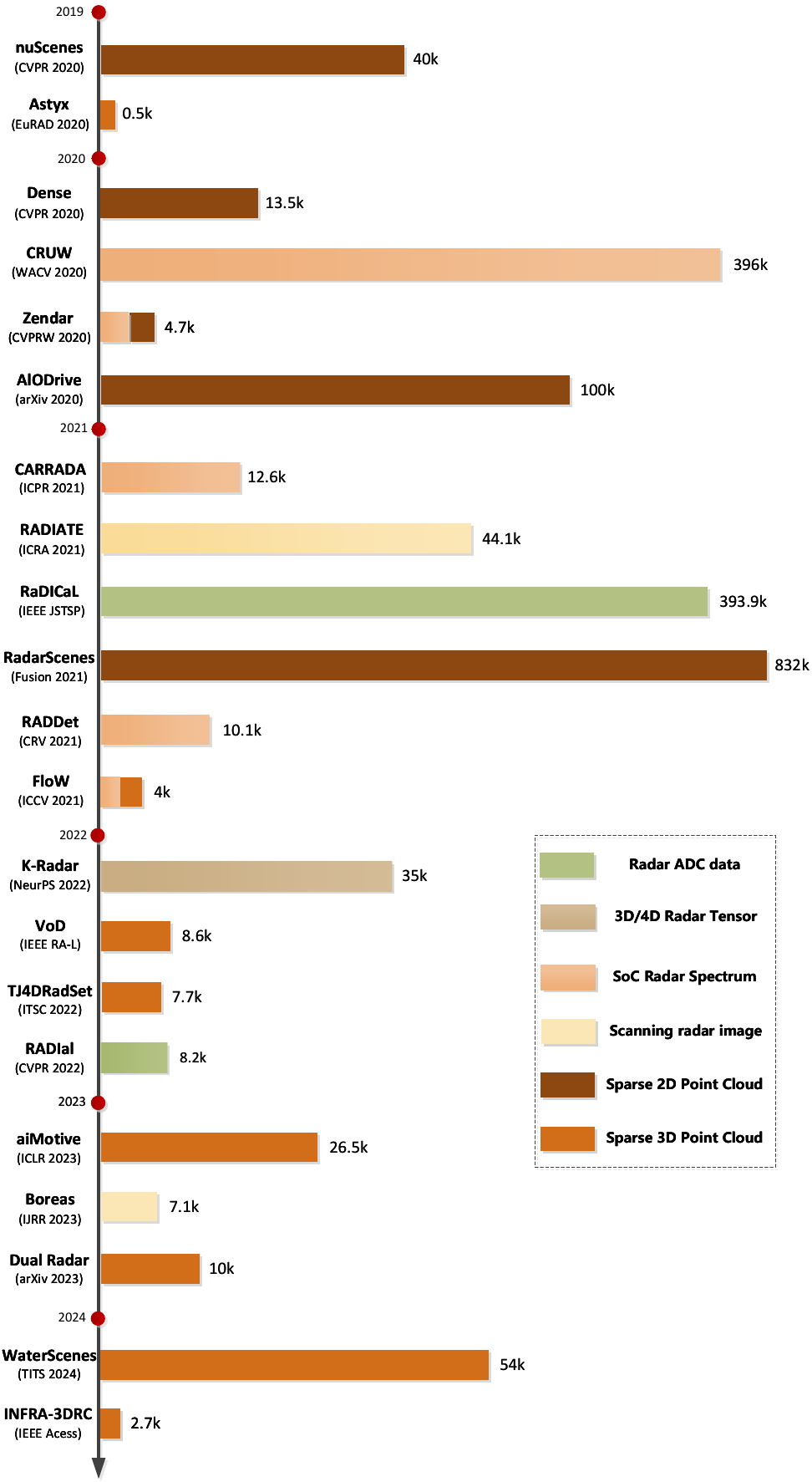}
	\caption{A thorough overview of public datasets with radar and camera data for object detection and tracking.}
	\label{fusion dataset}
\end{figure}

\subsection{Multi-modal Fusion}
In this subsection, we first provide a brief introduction on the definition and development of multi-modal fusion. Then, we summarize the multi-modal datasets and key elements concentrating on the radar-camera fusion.

\subsubsection{Definition and Development of Multi-modal Fusion}
Multi-modal information fusion, a.k.a. multi-source heterogeneous information fusion, can be formally defined as a multilevel, multifaceted process handling the automatic detection, association, correlation, estimation, and combination of data and information from heterogeneous sources \cite{khaleghi2013multisensor}.
To be concise, multi-modal fusion refers to integrating information from diverse modalities with the objective of predicting an outcome, either a class via classification or a continuous value via regression.

Multimodal fusion is an interdisciplinary research domain drawing upon concepts from a wide array of fields, including signal processing, information theory, statistical estimation, and artificial intelligence.
Information fusion technology first emerged in the military domain, aiming to provide accurate, comprehensive, and real-time battlefield assessment capabilities. Initially, research on information fusion mainly focused on data processing, leading to the early adoption of the term \textit{data fusion} to denote information fusion.
Theoretically, information fusion offers several advantages, primarily enhancing the authenticity and availability of the data \cite{he2022collaborative}. The former includes improvements in detection accuracy, confidence, and reliability, along with reducing data ambiguity, while the latter encompasses expanded spatial and temporal coverage. 

Traditional information fusion technologies include estimation theory methods (e.g., Kalman filter), statistical inference methods (e.g., Bayesian inference), information theory methods (e.g., entropy methods), decision theory methods, etc. 
In recent years, with the fast proliferation of artificial intelligence technology, the field of multimodal information fusion has witnessed the emergence of various novel approaches, including expert systems, fuzzy logic, machine learning, and deep neural networks. 
Among them, the maturing deep learning methods have attracted considerable attention from both industry and academia. Researchers widely regard DL-based multimodal fusion technology as the next frontier in overcoming the limitations of existing systems.

\subsubsection{Datasets for DL-based Radar-Camera Fusion}
The scale and versatility of multimodal datasets play a crucial role in DL-driven perception algorithms. With the mass production and popularization of radars, researchers have dedicated to collecting large-scale datasets containing both radar and camera data. Fig.~\ref{fusion dataset} summarizes the radar-vision fusion datasets for object detection and tracking, highlighting radar data representations and the number of labeled frames. In addition, we also provide an in-depth summary of the representative radar-related multi-modal fusion datasets in Table ~\ref{tab:radar related datasets}, highlighting the radar type and operating frequency.

\begin{figure}[h]
	\centering	\includegraphics[width=0.48\textwidth]{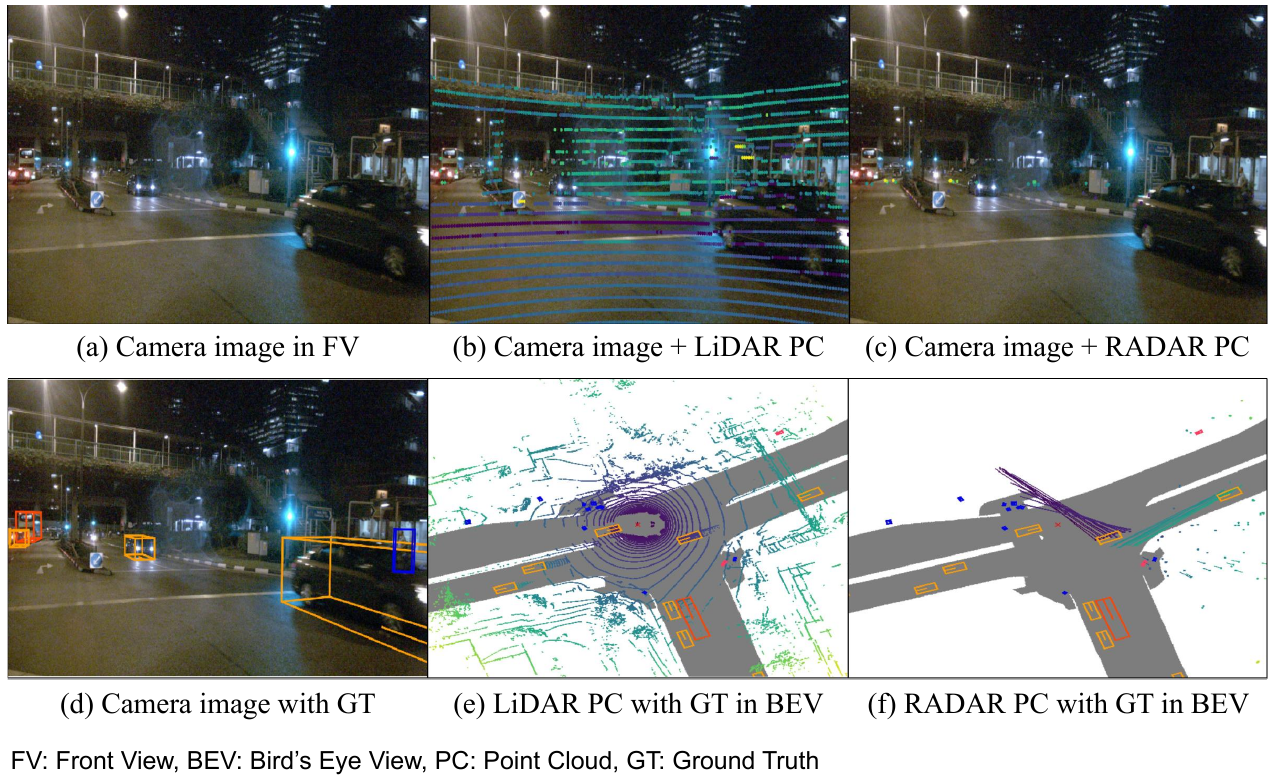}
	\caption{A visualization example of nuScenes dataset.}
	\label{nuscenes data}
\end{figure}

Notably, nuScenes \cite{caesar2020nuscenes} is the first authoritative large-scale dataset containing mmWave radar information that contributes to object detection and tracking tasks. 
The dataset comprises 1000 road scenes (each 20 seconds), showcasing diverse weather and lighting conditions, with nighttime scenes constituting 11.6$\%$ and rainy scenes 19.4$\%$.
For intuitiveness, we provide a visualization example of nuScenes dataset in Fig.~\ref{nuscenes data}. 

\begin{figure*}[htpb]
    \centering
    \includegraphics[width=1\linewidth]{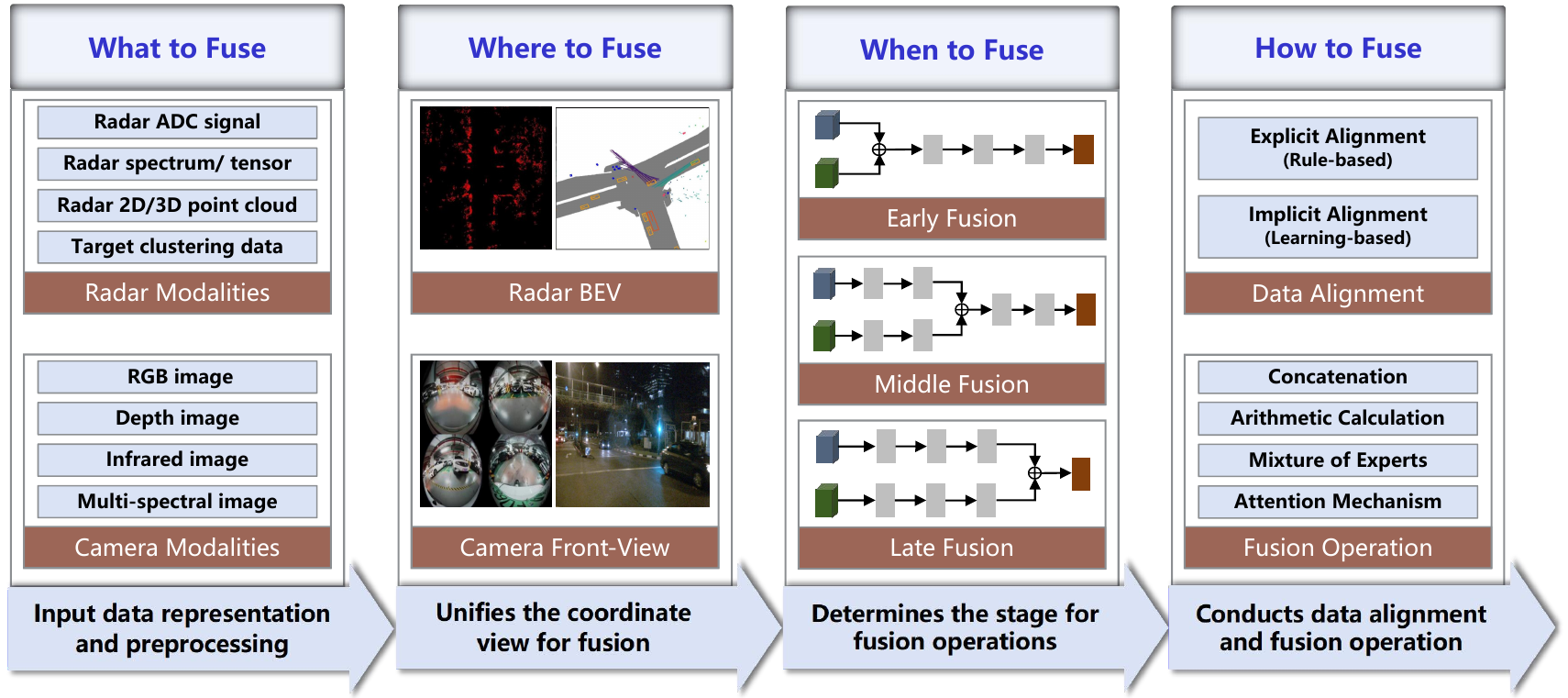}
    \caption{A holistic overview of key elements in radar-camera fusion.}
    \label{fig:fusion elements}
\end{figure*}
Besides, Astyx \cite{meyer2019automotive} presents a pioneering dataset involving high-precision 3D mmWave radar point clouds, albeit with limited scale.
Additionally, Dense \cite{bijelic2020seeing} is the first-of-its-kind multimodal dataset containing all-sided adverse weather conditions such as rain, snow, and fog.
Besides, CRUW \cite{wang2021rodnet} is a groundbreaking public dataset containing radar spectrum (range-azimuth matrix) data, specifically oriented toward detection task for pedestrian, bicycles, and vehicles. 
CARRADA \cite{ouaknine2021carrada} is the pioneering radar-camera dataset that provides both range-azimuth matrix and range-Doppler matrix annotations at the same time.
Moreover, RaDICaL \cite{lim2021radical} is the first radar dataset containing raw ADC signals, supporting object detection  for pedestrian and vehicle. 
K-Radar \cite{paek2022k} is a seminal dataset providing 4D radar tensors along range-Doppler-azimuth-elevation dimensions.
For the reader's convenience, we further category the radar-camera fusion datasets for object detection and tracking into four groups in terms of annotated bounding box dimension and perception task:

\begin{itemize}
\item \textit{2D object detection}: CRUW\cite{wang2021rodnet}, FloW\cite{cheng2021flow}, 
    CARRADA\cite{ouaknine2021carrada}, Zendar\cite{mostajabi2020high}, 
    RADIATE\cite{sheeny2021radiate}, RaDICaL\cite{lim2021radical}, 
    RADDet\cite{zhang2021raddet}, RADIal\cite{rebut2022raw}, Boreas\cite{burnett2023boreas}, INFRA-3DRC \cite{agrawal2024semi}.
\item \textit{3D object detection}: nuScenes\cite{caesar2020nuscenes}, Astyx \cite{meyer2019automotive}, Dense \cite{scheiner2020seeing}, RadarScenes\cite{schumann2021radarscenes}, AIODrive\cite{weng2020all}, TJ4DRadSet\cite{zheng2022tj4dradset}, K-Radar\cite{paek2022k}, aiMotive\cite{matuszka2022aimotive}, VoD\cite{palffy2022multi},   Dual Radar\cite{zhang2023dual}.
\item \textit{2D object tracking}: CARRADA\cite{ouaknine2021carrada}, RADIATE\cite{sheeny2021radiate}.
\item \textit{3D object tracking}: nuScenes\cite{caesar2020nuscenes}, AIODrive\cite{weng2020all}, RadarScenes\cite{schumann2021radarscenes}, TJ4DRadSet\cite{zheng2022tj4dradset}, K-Radar\cite{paek2022k}, Dual Radar\cite{zhang2023dual}.
\end{itemize} 

\subsubsection{Key Elements about Radar-Camera Fusion}
This part outlines a systematic overview of five key issues in the radar and camera fusion domain, namely \textit{why to fuse}, \textit{what to fuse}, \textit{where to fuse}, \textit{when to fuse}, and \textit{how to fuse}. For intuitiveness, Fig.~\ref{fig:fusion elements} provide an illustration of these nucleus concerns.
\begin{itemize}
  \item Why to fuse: reveals insight on the positive benefits of radar-camera fusion, including improved accuracy, robustness and versatility.
  \item What to fuse: pertains to the diversified modal representation and preprocessing of radar and camera data before fusion.    
  \item Where to fuse: depicts the coordinate relationships between modalities and determines the coordinate perspective for fusion.
  \item When to fuse: aims to select the optimal stage/level for multi-modal fusion operations.
  \item How to fuse: encompasses the spatial-temporal alignment between heterogeneous data and the specific fusion operations.
\end{itemize}

\textbf{\emph{a) Why to fuse}}. The combination of radar and camera sensors is essential to the implementation of reliable object detection and tracking in complex environment. This fusion approach effectively merges the radar’s strength in offering robust measurements of distance and velocity, regardless of challenging illumination and weather scenarios, with the camera's proficiency in capturing intricate visual details such as color, contour, and texture. By synergizing these technologies, the system not only mitigates the limitations inherent to each individual sensor but also ensures seamless operation regardless of lighting and weather conditions. In the realm of autonomous systems, this integrated approach significantly elevates the precision of object detection, tracking, and environmental awareness, which is pivotal for high-stakes operations like autonomous driving, robotic navigation, and surveillance. Moreover, it fortifies the system's reliability by providing redundancy, thereby ensuring uninterrupted functionality even in the event of a sensor malfunction. This heightened level of performance is critical for supporting complex functionalities such as real-time decision making, trajectory planning, and dynamic interaction with unpredictable environments. Beyond mobile sensing, industries like manufacturing, logistics, and even healthcare are capitalizing on radar-camera fusion technology, which ensures precision in automated processes, optimizes inventory management, and facilitates innovative monitoring and assistance systems for patient care. As technological progress continues to drive down costs and enhance capabilities, the amalgamation of radar and camera sensors stands as a cornerstone for innovation, offering a robust and versatile solution that meets the stringent demands of modern applications and regulatory standards, paving the way for safer and more intelligent systems.

\begin{figure}[h]
    \centering
    \includegraphics[width=0.99\linewidth]{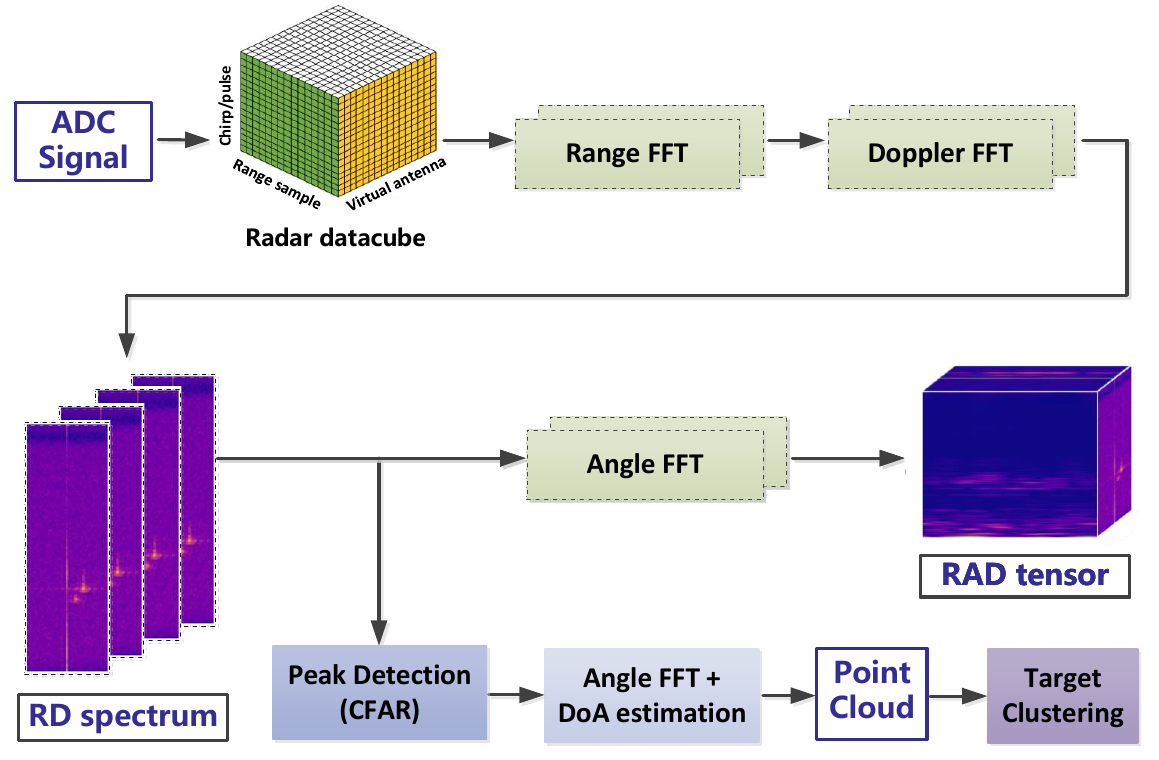}
    \caption{The typical radar signal processing.}
    \label{fig:radar singal processing}
\end{figure}

\textbf{\emph{b) What to fuse}}. 
Before fusion, it is essential to specify the representation of each sensory input and perform certain preprocessing. For camera sensors, the data format commonly includes RGB images, depth images, and infrared images, presented as low-dimensional 2D images that can directly utilize deep neural networks for feature extraction. In addition, some literature harnesses extra optic information, such as optical flow \cite{long2021radar} and multi-spectral images \cite{valada2017deep}.
In contrast, the data dimension of radar sensors tends to be higher and the modal representations are diverse. In light of the typical radar working principle and the signal processing pipeline (as shown in Fig.~\ref{fig:radar singal processing}), prevailing radar data representations implicate: \textbf{ADC signal} (a.k.a. raw radar data), \textbf{radar spectrum/tensor} (e.g., RD matrix, RAD tensor), \textbf{radar point cloud}, target clustering information, etc.

For modern commercial radars, the most primitive data structure is the binary data stream obtained after sampling and quantization of the IF signal by the ADC. Generally, the time-domain ADC signals are complex-valued, comprising both real and imaginary parts. Although maintaining the richest information, these original ADC signals are extremely unintuitive and require significant memory overhead. 
In practice, researchers tend to rearrange the ADC data into a radar datacube with three dimensions: range sample (fast time), chirp/pulse number (slow time), and antenna. Subsequently, different Fast Fourier Transforms (FFTs) are applied to obtain three types of 2D radar spectrum data, namely Range-Doppler (RD) matrix, Range-Azimuth (RA) matrix, and Azimuth-Dopper (AD) matrix, as well as 3D/4D radar tensor data (e.g., RAD tensor) \cite{major2019vehicle}.
While the aforementioned structured form enhances the intuitiveness of the data, storage demands for spectrum/tensor-form radar data still remain substantial.
As such, conventional radar systems typically employ peak detection algorithms, e.g., the Constant False Alarm Rate (CFAR) processor, to filter out background and noise, resulting in sparse point cloud data. 
Furthermore, due to the dispersed distribution characteristic of sparse point clouds and the interleaving of points reflected by adjacent targets, clustering algorithms, like Density-Based Spatial Clustering of Applications with Noise (DBSCAN) \cite{ester1996density}, are frequently applied to separate target point cloud clusters. 
The structured target data after clustering contains information such as the position and velocity of targets, commonly used for multi-target tracking.

\begin{figure}[htbp]
\centering
\includegraphics[width=0.90\linewidth]{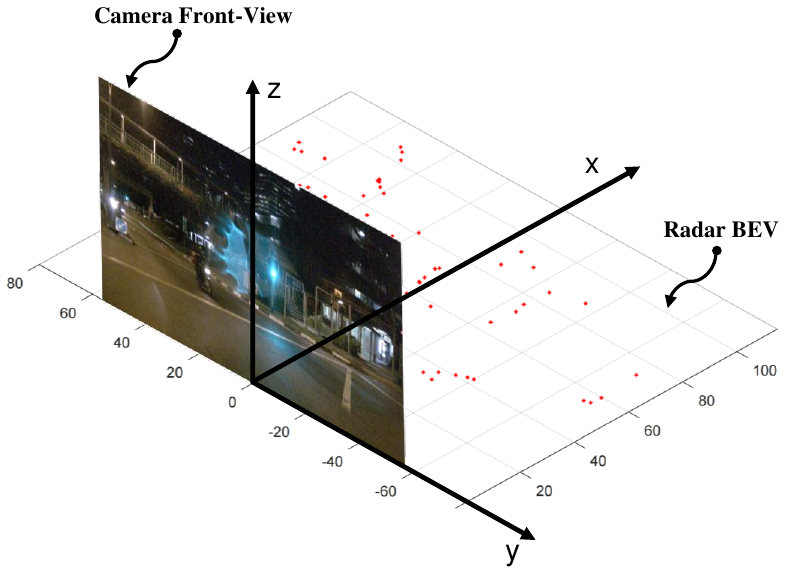}
\caption{An illustration of two coordinates (from \cite{shi2023radar}).}\label{FV_BEV}
\end{figure}

\textbf{\emph{c) Where to fuse}}. 
Given the intrinsic modality disparateness between radar and camera data, unifying the modal view is conducive to fusion operations. With regard to radar-camera fusion, the most common coordinates are the \textbf{Front View} (FV) and \textbf{Bird's-Eye View} (BEV). Concretely, the spatial relationship between these two coordinate views is depicted in Fig.~\ref{FV_BEV}.

The front view refers to a visual representation of a scene from the frontal direction. In the context of radar-camera fusion, a front view typically involves the perspective captured by a camera sensor facing forward, providing a direct view of the environment ahead. 
Fusion at the FV coordinate entails the projection of the radar information to the 2D perspective plane of the camera image through coordinate transformation. Depending on how radar data is utilized, this fusion paradigm roughly falls into two categories.
One branch of research first generates RoIs within the FV image plane around the projected radar data that potentially contain valid targets. Subsequently, feature extraction and post-processing are performed merely within these RoIs \cite{nabati2019rrpn,yadav2020radar+}. Although the RoI-based fusion effectively enhances computational efficiency, it may lead to insurmountable missed detection.
Another line of work first projects the radar data onto the camera's perspective plane, whereas followed by creating radar pseudo-images that reflect radar information (e.g., distance, velocity, RCS) through certain rendering techniques, aiming to complement the image-based features \cite{nobis2019deep,chang2020spatial}.
The major defects of this method lie in that it struggles to differentiate adjacent or occluded objects within the front view, as well as exacerbates the sparsity of radar data.
Moreover, considering the imperfection (e.g., the incompleteness and ambiguity) of radar data, along with the potential mismatch between heterogeneous modalities in spatial-temporal registration, the radar data projected onto the perspective view may deviate from the actual targets in the corresponding image.

Bird's-eye view is a term used to describe an overhead perspective of a scene as observed from above, allowing for a comprehensive understanding of spatial relationships and layouts. Fusion at the BEV coordinates has recently  increasingly garnered research attention \cite{nabati2021centerfusion,kim2023crn,zhou2023bridging}, as it mitigates the daunting occlusion issues in FV, as well as maintains both the spatial information provided by radar and the semantic features contained in camera images.
For radar sensors, sparse point cloud data inherently applies to BEV coordinates, which provide an intuitive and effective way to represent the object's bounding box and orientation, thus circumventing the geometric distortion.
Nonetheless, radar BEV maps are typically generated by converting radar point clouds in a voxel or grid format, which is inefficient and impractical when dealing with sparse radar point clouds. 
Besides, the discretization is prone to causing the loss of essential information required for refining bounding boxes.
For camera sensors, the projection of image data onto the BEV can generate dense feature maps, thus preserving semantic information.
Classically, Inverse Projection Mapping (IPM) methods
\cite{bertozz1998stereo,oliveira2015multimodal} are applied to map an image to a BEV coordinate. However, challenges arise as the depth associated with each pixel is inherently ambiguous.
More recently, Lift-Splat-Shoot (LSS) \cite{philion2020lift} formulates a view transform scheme that explicitly predicts depth distribution and converts images to a BEV map, which has inspired a series of follow-up work \cite{liu2023bevfusion,huang2021bevdet,li2023bevdepth}.

\renewcommand\arraystretch{0.1}
\begin{table*}[h]
 \centering
 \caption{The advantages and limitations of different fusion levels} \label{tab:fusion-level}
  \begin{tabular}{|m{1.7cm}<{\centering}|m{7.5cm}|m{7.5cm}|}
   \hline \rowcolor{gray!40} 
     \textbf{Fusion Level} & \multicolumn{1}{c|}{\textbf{Advantage}}  & \multicolumn{1}{c|}{\textbf{Limitation}}  \\ 	\hline
     Early fusion & ~ \begin{itemize}
  \item  Incorporates the most comprehensive modal information.
\end{itemize}    & ~ \begin{itemize}
  \item  Struggles to large data volume, substantial computational cost, susceptibility to sensor misregistration, poor robustness, and limited flexibility.
\end{itemize}     \\ \hline

    Middle fusion & ~ \begin{itemize}
  \item Allows data compression to enhance real-time performance. 
  \item Applies for adaptively exploiting synergies between heterogeneous modalities based on DL-driven framework.
  \item Possesses moderate model robustness and flexibility.
\end{itemize}     & \begin{itemize}
  \item The technology remains in its nascent stage, especially for radar-camera fusion.
\end{itemize} \\ \hline

   Late fusion & ~ \begin{itemize}
  \item Achieves flexible modular construction and low communication overhead.
\end{itemize}  &  ~ \begin{itemize}
  \item Encounters insurmountable difficulties in losing a substantial amount of valuable information when exploring complementary interactions between heterogeneous modalities.
\end{itemize}  \\ \hline
   \end{tabular}
\end{table*}

\textbf{\emph{d) When to fuse}}. 
Based on the level of fusion and the abstraction level of the fused data, multimodal fusion levels can be classified into: data-level, feature-level, and decision-level. In the context of radar-camera fusion, data-level fusion is often referred to as \textbf{early fusion}, decision-level fusion is also called object-level fusion or \textbf{late fusion}, and feature-level fusion is also known as \textbf{middle fusion}. 
The characteristics of the three mainstream fusion levels are summarized as follows:
\begin{itemize}
  \item Early fusion: Fusion is performed at the low level, where the raw data or pre-processing data from different modalities are merged at the early stage. Subsequently, the integrated data undergoes further processing according to task requirements to produce the final output. 
  \item Middle fusion: Fusion occurs at the intermediate level, where features are first extracted individually from each modality. These features are then fused together to generate new feature tensors for subsequent data processing.
  \item Late fusion: Fusion takes place at a high level, where each modality is separately processed to extract features and make independent predictions. Finally, these predictions are integrated to produce a unified result.
\end{itemize}

The selection of fusion level implicates not only the accuracy requirements but also multiple factors such as model flexibility and computational cost. Table.~\ref{tab:fusion-level} succinctly outlines the most prominent benefits and drawbacks of the three aforementioned fusion levels.
With regard to radar-camera fusion, early work mostly focuses on late fusion. Recently, with the development of DL technology and the release of public radar datasets, most mainstream radar-camera fusion methods have endeavored to explore middle fusion strategies.

In early fusion, input data from radars and cameras are associated at the initial stage of the model. Therein, achieving precise spatio-temporal calibration of heterogeneous modal information is a crucial prerequisite. 
On the one hand, early fusion involves the most detailed information and thus possesses the ability to maximize the extraction of comprehensive features from sensor data. On the other hand, processing raw data is accompanied by large data volume, high computational costs, limited flexibility, and susceptibility to sensor misregistration. More importantly, 
due to the significant modal disparity between the radar data and camera data, realizing accurate association between radar data and image pixels is extremely challenging.

Late fusion merges the results independently inferred from the radar branch and camera branch at the final stage of the model. Therein, how to optimally match the decision results of the two heterogeneous modalities is a foremost consideration. One strategy is based on similarity assessment and employs techniques such as Kalman filter, Bayesian theory, and Hungarian algorithm for matching. Another strategy relies on coordinating the output results' positional relationship through a coordinate transformation matrix between radar and camera sensors. 
Literally, late fusion offers high flexibility and modular characteristics, while its performance depends on the accuracy of the inference results of each modality. For example, in cases of severe obstruction or adverse weather for cameras, decision-level fusion might rely solely on the prediction results from radar. Additionally, the processing mode of decision-level fusion is prone to losing valuable intermediate features from different modalities. 

Middle fusion integrates the strengths of early fusion and late fusion, making it well-suited for DL networks specialized in feature extraction. Typically, the dissimilar modal data in radar-camera fusion contributes variably to the perception outcomes. One modality might play a dominant role, while the other modality provides auxiliary complementary information. Feature-level fusion is ideal for exploring these complementary interactions between heterogeneous modalities.
Within middle fusion methods, there exits a special paradigm known as \textit{hybrid-level fusion}. This approach leverages information from various feature levels, effectively preserving features from each stage. Nonetheless, hybrid-level fusion often lacks interpretability and necessitates experimentation to determine the optimal network architecture based on empirical knowledge. 
Moreover, models employing hybrid-level fusion generally incorporate more branches in neural networks, leading to increased computational overhead to some extent.

\textbf{\emph{e) How to fuse}}.   
In radar-camera fusion, the foremost concern lies in overcoming the heterogeneity and exploiting the synergy of multi-source information.
Consequently, in contrast with the aforementioned three key elements, the paramount problems of how to fuse (e.g., data alignment and fusion operation) tend to be of top priority. Therefore, we endeavor to elaborately highlight the key techniques in the following Section IV.

\subsection{Sensing Tasks}
\subsubsection{Object Detection}
The object detection task aims to identify and categorize specific targets within camera images or radar scans, as well as determine their positions by encapsulating them within rectangular or cuboid bounding boxes.
In the case of 2D object detection, a valid target is normally represented as $(x, y, w, h, c)$, where $(x, y)$ denotes the center of the rectangular bounding box, $w$ and $h$ indicate the width and height of the bounding box, and $c$ refers to the class of the object. 
As for the 3D object detection, the target of interest is commonly described as $(x, y, z, w, h, l, \theta, c)$, where $(x, y, z)$ represents the center of the 3D bounding box, $w$, $h$ and $l$ are the width, height and length of the cuboid bounding box, $\theta$ denotes the orientation of the target, and $c$ indicates the category information. 
In particular, BEV detection is a prevalent approach of 3D object detection  where height information is usually disregarded, and objects are portrayed as 2D bounding boxes on the ground plane.

\subsubsection{Object Tracking}
Once an object is detected, its state is measured, including the position in range, azimuth/elevation angle, and even its velocity as well as shape. Essentially, the object tracker is a state estimator, a.k.a., a filter, that continuously measures the state of targets over time, and the consecutive position measurements are merged and smoothed to estimate the trajectory of the target.
In general, an ideal object tracker entails the following functions: 1) assigning and maintaining a unique trajectory identity (ID) for each detected target throughout the entire tracking process; 2) eliminating false-alarm detection; 3) maintaining tracking of targets that have not been detected for a period of time (missed detection); 4) refining the state information provided by upstream modules and supporting downstream tasks.

\subsection{Evaluation Metrics}
A series of mathematical metrics are formulated to evaluate the sensing performance of radar-camera fusion.

\subsubsection{Evaluation Metrics for Object Detection}
During the initial stage of radar and camera fusion research, object detection metrics commonly include accuracy, precision, recall, average precision (AP), average recall (AR), mean average precision (mAP), and mean Intersection over Union (mIoU), akin to those used in image-based  object detection task. 
Unfortunately, crucial information about an object, including velocity, range, and orientation, are neglected in these metrics. 
Moreover, in 3D object detection, IoU thresholds need to account for factors like object distance, occlusion, and sensor type.
To address these limitations, the nuScenes dataset \cite{caesar2020nuscenes} 
introduces a new metric 
$
\mathrm{mTP}=\frac{1}{|\mathbb{C}|} \sum_{C \in \mathbb{C}} \mathrm{TP}_{C} \,,
$
where $\mathbb{C}$ denotes the set of object categories, and $\mathrm{TP}_{C}$ denotes the True Positive metric, namely ATE, AOE, ASE, AVE, or AAE,
representing mean average translation, orientation, scale, velocity, and attribute errors, respectively. 
For readability, the relevant definitions are listed as follows:
\begin{itemize}
  \item ATE: The Euclidean distance between center positions of prediction and ground truth in 2D BEV (measured in meters).
  \item AOE: The difference in yaw angle between the predicted bounding box and ground truth (measured in radians).
  \item ASE: The 3D IoU error after aligning translation and orientation (1- IoU).
  \item AVE: The L2 norm of the 2D velocity difference between the prediction and ground truth  (measured in m/s).
  \item AAE: The attribute classification error, defined as 1 minus attribute classification accuracy.
\end{itemize}

Besides, the calculation of AP in nuScenes 3D object detection challenge differs from the conventional counterpart. Instead of using IoU for threshold matching, the AP in nuScenes utilizes the Euclidean distance $D$ between the centers of predicted and ground truth bounding boxes. This unique design aims to decouple the impacts of object size as well as orientation on AP calculation results, and also considers the fact that an IOU of 0 is improperly assigned when small objects are detected with a small localization error \cite{caesar2020nuscenes}. The official mAP calculation formula provided by nuScenes is as follows:
\begin{equation}
\mathrm{mAP}=\frac{1}{|\mathbb{C}||\mathbb{D}|} \sum_{C \in \mathbb{C}} \sum_{D \in \mathbb{D}} \mathrm{AP}_{C, D} \,,
\end{equation}
where $\mathbb{D}= \left\{0.5, 1, 2, 4\right\}$ (in meters) represents the set of distance matching thresholds.

Finally, by integrating mAP with mTP, a comprehensive 3D object detection metric called nuScenes Detection Score (NDS) is deduced:
\begin{equation}
\mathrm{NDS}=\frac{1}{10}\left[5 \mathrm{mAP}+\sum_{\mathrm{mTP} \in \mathbb{TP}}(1-\min (1, \mathrm{mTP}))\right],
\end{equation}
where $\mathbb{TP}$ is the set of the five mean True Positive metrics. Since
mAVE, mAOE and mATE can be larger than 1, nuScenes \cite{caesar2020nuscenes} bounds each metric between 0 and 1.

\subsubsection{Evaluation Metrics for Object Tracking}
For model-based radar Multi-Object Tracking (MOT) algorithms, the most commonly used metric is Optimal Subpattern Assignment (OSPA) indicator \cite{schuhmacher2008consistent}, which combines the target number estimation error and position estimation error.
As for radar-camera fusion-based MOT, canonical evaluation metrics originate from the field of computer vision. 
To comprehensively evaluate the performance of MOT algorithms, multiple indicators need to be integrated.
Specifically, identity switches (IDSW) measure the ID changing times of all targets during the tracking process. FP (False Positive) represents the total number of falsely tracked objects, while FN (False Negative) denotes the total number of missed tracks.
Multi-Object Tracking Accuracy (MOTA) combines the above metrics to measure the matching errors between target measurements and tracking hypotheses in all frames:
\begin{equation}
MOTA = 1 - \frac{FP + FN + IDSW}{\mathrm{num_{gt}}}.
\end{equation}

Although the MOTA metric can intuitively describe the accuracy of object tracking algorithms in detecting objects and maintaining trajectory continuity, it lacks the ability to measure localization accuracy. To this end, another mainstream metric, Multi-Object Tracking Precision (MOTP) \cite{bernardin2008evaluating}, is applied to measure the localization accuracy of targets:
\begin{equation}
 MOTP = \frac{\sum_{i,t}d_t^i}{\sum_{t}TP_t},
\end{equation}
where $d_t^i$ represents the distance between the bounding box of tracking hypothesis $i$ and the ground truth at frame $t$, while $TP_t$ denotes the number of correct matches at frame $t$.
Additionally, nuScenes introduces two new MOT metrics, called Average Multi-Object Tracking Accuracy (AMOTA) and Average Multi-Object Tracking Precision (AMOTP). 
Their specific expressions are as follows:

\footnotesize
\begin{align}
    AMOTA & = \frac{1}{n-1}\sum_{r \in \left\{\frac{1}{n-1},\frac{2}{n-1},...,1 \right\}} MOTAR, \\
    MOTAR & = \mathrm{max} \left(\!0, 1\! - \!\frac{IDSW_r \!+ \!FP_r \!+\! FN_r \!-\! (1 \!- \!r) \times \mathrm{num_{gt}}}{r \! \times \! \mathrm{num_{gt}}} \right), \\
    AMOTP & = \frac{1}{n-1}\sum_{r \in \left\{\frac{1}{n-1},\frac{2}{n-1},...,1\right\}} MOTP,
\end{align}
\normalsize
where $n$ refers to the number of sampled evaluation points, while $r$ represents the recall rate.

\section{Key Techniques for Radar-Camera Fusion}
In this section, we endeavor to shed light on the key techniques for radar-camera fusion by emphatically discussing the following issues: sensor calibration, modal representation, data alignment, and fusion operation.

\subsection{Sensor Calibration} 
In order to represent sensor observations in a common reference frame, the rigid transformations (i.e. 3D rotation and translation) between radar and camera coordinates must be estimated accurately. To address this fundamental problem for further  fusion, several methods for radar-camera calibration have been proposed. 
The necessity of a specially designed calibration target in the calibration process is a crucial factor, resulting in two main groups of calibration strategies.

\subsubsection{Target-Based Calibration}
Given the inherent noise of radar sensors, target-based calibration approaches typically utilize specific calibration targets to simultaneously locate them with the radar and camera. The triangular corner reflector, which reflects RCS values and is detectable by both radar and camera, is popular for radar-camera calibration. 
Early radar extrinsic calibration algorithms \cite{sugimoto2004obstacle,wang2011integrating,kim2014data} are primarily developed to facilitate the fusion of radar and camera data, relying on computing the projective homography that aligns points from the radar sensing plane to the perpendicular camera image plane. Although most commercial radar sensors cannot accurately measure the elevation of distant targets, they are capable of detecting objects at a small elevation angle relative to the radar horizontal plane. Consequently, the reflectors need to be situated on the radar horizontal plane to acquire precise calibration. As such, the authors in \cite{sugimoto2004obstacle} address this by filtering radar-camera measurement pairs based on the returned radar signal intensity. The position of the trihedral reflector is at the peak on the radar horizontal plane.
Besides, more recent radar extrinsic calibration approaches focus on minimizing reprojection error, which measures the misalignment of objects visible to both sensors. The authors in \cite{kim2017comparative} utilize reprojection error to estimate the radar-to-camera transformation, assuming radar measurements are constrained to the zero-elevation plane. Besides, the authors in \cite{el2015radar} estimate the radar-to-camera transform by intersecting back-projected camera rays with the 3D arcs along the 2D radar plane. Subsequently, the authors in \cite{pervsic2019extrinsic} also utilize 3D point cloud alignment for calibration but improve the accuracy by modeling the relationship between target returned intensity and elevation angle. Furthermore, the authors in \cite{oh2018comparative} summarize and compare the homography and reprojection methods, concluding that both methods have similar performance. 

To obtain target positions from both radar and camera sensors, several uniquely designed calibration boards have been proposed. For instance, the authors in \cite{domhof2019extrinsic} employ a specialized calibration target that provides scale for the camera measurements, facilitating extrinsic calibration through point cloud alignment. Moreover, the authors in \cite{wang2021roadside} propose to combine a corner reflector with a Styrofoam board, using the latter for visual recognition by camera sensors without interfering with radar signals.
In addition, the authors in \cite{pervsic2021spatiotemporal} introduce a calibration board featuring a corner reflector and a checkerboard-patterned Styrofoam triangle, allowing both sensors to acquire precise target positions. By using the paired sets of image pixels and radar points from the same targets at various locations, the transformation matrix between radar and camera coordinates is computed. 

\subsubsection{Target-Free Calibration}
Apart from target-based calibration, several target-free calibration algorithms that do not rely on specialized reflective targets have emerged recently. For example, the authors in \cite{pervsic2021online} estimate the yaw angles between radar, camera, and LiDAR sensors by aligning the trajectories of tracked objects. The translation parameters are manually measured by calculating sensor overlapping fields of view. Some methods \cite{guo2018pedestrian,qiu2020real,wise2021continuous} utilize precise radar velocity measurements on objects and camera poses to implement radar-to-camera extrinsic calibration. The authors in \cite{wise2023spatiotemporal} introduce a method for extrinsic calibration in continuous time by utilizing instantaneous radar ego-velocity measurements and camera ego-motion measurements. 
With the development of deep learning, many approaches \cite{iyer2018calibnet,scholler2019targetless,zhao2021calibdnn} make efforts to design end-to-end neural networks to regress calibration matrices from raw camera images and radar data. 

Some methods obtain the radar extrinsic parameters by calibrating the radar with other types of sensors.
Heng \cite{heng2020automatic} presents the first reprojection error-based 3D radar-LiDAR extrinsic calibration algorithm that does not require specialized targets or overlapping fields of view. This method estimates the extrinsic calibration between multiple LiDAR units and constructs a 3D point cloud map using a known vehicle trajectory. The radar-LiDAR extrinsic calibration parameters are derived by minimizing the distance from radar point measurements to the closest plane in the LiDAR map and the radial velocity error.  Besides, the authors in \cite{kellner2015joint} propose a method for estimating the rotation between a car-mounted 2D radar and an IMU by minimizing the discrepancy in estimated lateral velocities. The authors in \cite{doer2020radar} extend this approach to estimate the full extrinsic calibration for a 3D radar-IMU pair, achieving remarkable calibration accuracy with simulated, low-noise radar measurements.

\subsection{Modal Fusion Representation}
Representing multimodal information in a format that is compatible with the model processing is essential to radar-camera fusion. 
Similar to \cite{bengio2013representation}, we employ the term feature and representation interchangeably, both denoting a vector or tensor depiction of the physical modal information, whether it is an image, radar spectrum/tensor, or radar point clouds.

Multimodal representation presents numerous issues, including integrating information from diverse data sources, managing varying levels of noise, and addressing missing data. The capability to meaningfully depict modal information is paramount in addressing multimodal tasks and serves as the foundation of the fusion framework \cite{zhao2024deep}. 
The last decade has witnessed the transition of data representation from hand-crafted solutions to methods driven by deep learning.
Currently, the majority of image and radar data is characterized by features acquired from neural networks like Convolutional Neural Networks (CNNs), Graph Neural Networks (GNNs), and Transformer models \cite{huang2023hdnet}. Specifically, for camera images and radar spectrum/tensor data, ResNet \cite{he2016deep} and ViT \cite{dosovitskiy2020image} are the two most commonly employed models that are applicable for image-format input.
As for sparse radar point clouds, point-wise methods and grid-wise approaches are two practicable representations. The former can be further categorized into PointNet \cite{qi2017pointnet} and Point-GNN \cite{shi2020point} series, while the latter consist of occupancy-based gridmaps \cite{wei2023deep} (representing obstacles and free space) and amplitude-based gridmaps \cite{werber2015automotive} (showing RCS values for each cell). 

\begin{figure}[htbp]
\centering
\includegraphics[width=0.83\linewidth]{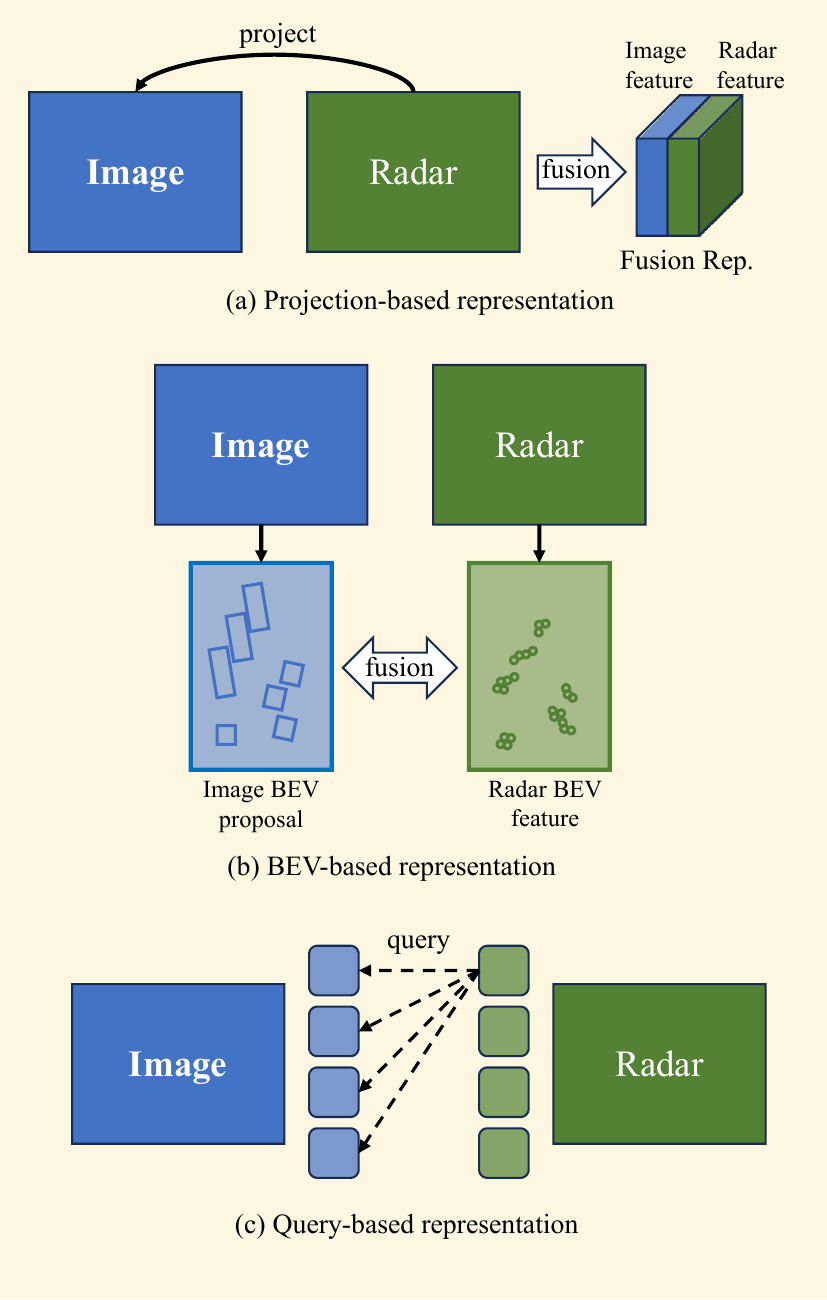}
\caption{An illustration of image-radar modal fusion representation}
\label{modalrep}
\end{figure}

Departing from the use of independent networks for processing image and radar modalities, utilizing a shared network to learn the fusion features of the modalities proves to be a more effective method.
As a result, there is a need for a compatible modality fusion representation that can effectively capture the features of both image and radar data.
Depending on the representation form, we categorize fusion representations into projection-based, BEV-based, and query-based representations, as shown in Fig. \ref{modalrep}.

\subsubsection{Projection-based Representation}
The workflow can be delineated as follows: First, radar point clouds are mapped onto the image plane via a radar-to-image projection matrix. Following this, fusion features are extracted using image processing networks (e.g., ResNet \cite{he2016deep}, UNet \cite{ronneberger2015u}, and ViT \cite{dosovitskiy2020image}).
For example, Chadwick \textit{et al.} \cite{chadwick2019distant} employ projection to derive range images and range-rate images from radar data, then fuse them with RGB images through concatenation and element-wise addition.
Similarly, CRF-Net \cite{nobis2019deep} maps radar point clouds onto the image plane and utilizes a multi-layer feature structure for layer fusion, dynamically adjusting fusion weights across various layers for optimal benefit.
RVNet \cite{john2019rvnet} proposes a one-stage object detection network based on the YOLOv3 with two input branches.
The network obtains sparse radar images through coordinate transformation, where the sparse radar images consist of three channels: depth, lateral velocity, and longitudinal velocity.
Moreover, the branch design provides enhanced flexibility in calculating fusion weights for the two modalities, which has been widely adopted in subsequent research \cite{john2021deep, john2021bvtnet}.
SONet \cite{john2020so} continues the fusion design of RVNet and extends it to a multi-task framework that incorporates semantic segmentation and object detection.
YOdar \cite{kowol2020yodar} entails lowering the threshold score and assigning radar point clouds to image slices, which are then combined into an aggregated output.
Unlike prior methods that extend radar points by copying its z-component along the vertical direction to create radar lines (which could lead to incorrect correspondences), the authors in \cite{singh2023depth} associate radar points with multiple potential surfaces in the image within a search image crop centered on the position of the point. This refined strategy enhances accuracy and avoids potential errors in correspondence assignment.
In addition, Fatseas \textit{et al.} \cite{fatseas2023association} associate 2D bounding boxes and radar detections based on the Euclidean distance between their projections. 
The authors transform feature vectors instead of single points to learn non-linear dynamics and utilize several features to provide a better description for each object.
MS-YOLO \cite{song2022ms} integrates a mapping transformation network to enable the fusion of radar and image information at the same scale, thereby enhancing the detection capability for small, distant targets.

\subsubsection{BEV-based Representation}
Recently, BEV representation has demonstrated outstanding performance in the fusion of image and radar data.
For image-based BEV representation, LSS \cite{philion2020lift} first proposes a lift-splat-shoot paradigm based on depth estimation.
Inspired by the remarkable effects of LSS, BEVDet \cite{huang2021bevdet} and BEVDepth \cite{li2023bevdepth} innovatively tackle the 3D object detection challenges by transforming multi-view 2D image features into a frustum with depth encoding. This process involves consolidating the features into a unified BEV representation by collapsing the height dimension. 
For radar-based BEV representation, it is often acquired through the compression encoding of spatial features along the $z$-axis of point cloud data.

In the context of radar-image BEV representation, there has been a considerable amount of research in this field.
Lim \textit{et al.} \cite{lim2019radar} utilize IPM to convert camera images into Cartesian coordinates, subsequently integrating them with 2D radar RA sprectrums.
Some methods \cite{bertozz1998stereo, oliveira2015multimodal} also adopt IPM to convert camera images from FV to BEV for fusion representation.
Cui \textit{et al.} \cite{cui20213d} project image and radar point cloud into BEV and design a CNN-based cross-fusion strategy to increase the quality of proposal generation.
Besides, Bansal \textit{et al.} \cite{bansal2022radsegnet} introduce extra point-based features, such as velocities and RCS values, into the BEV map.
Simple-BEV \cite{harley2023simple} produces a high-resolution BEV feature map through coordinate transformation from multiple radar sensors.
CRN \cite{kim2023crn} employs deformable attention to address the misalignment problem between radar and camera feature maps, resulting in accurate BEV representation.
Upon publication, CRN \cite{kim2023crn} exhibits outstanding performance and stands out among radar-image fusion methods.
Moreover, RCFusion \cite{zheng2023rcfusion} accomplishes multimodal feature fusion within a unified BEV framework, utilizing inputs from both 4D radar and camera.
Recently, EchoFusion \cite{liu2024echoes} enhances the BEV representation by integrating abundant and lossless distance and velocity cues from radar echoes with semantic cues from images, achieving performance comparable to LiDAR detection.

\subsubsection{Query-based Representation}
In addition to BEV-based methods, there exists a class of schemes that generate object queries to enable interaction between image and radar modalities.
Typically, this line of work utilizes transformer modules, particularly cross-attention mechanisms, to integrate features from various modalities.
Inspired by DETR \cite{carion2020end} (which supports end-to-end onject detection, simplifying complex post-processing steps), 
DETR3D \cite{wang2022detr3d} generates object inquiries and introduces a 3D reference point for each query. The reference points are utilized to consolidate multi-view image features as keys and values, while cross-attention is applied between object queries and image features. Ultimately, each query can decode a 3D bounding box for detection. Numerous subsequent studies have embraced the concept of object queries and reference points.
PETR \cite{liu2022petr} extends this concept by incorporating 3D positional embeddings into query generation. Besides, BEVFormer \cite{li2022bevformer} capitalizes on spatial and temporal information by engaging with predefined grid-shaped BEV queries, establishing spatial cross-attention where each BEV query extracts spatial features from regions of interest across multi-camera views.

As a representative radar-camera fusion work, CRAFT \cite{kim2023craft} associates image proposals with radar points in the polar coordinate system and then utilizes cross-attention based feature fusion layers to share spatio-contextual information between image and radar.
In addition, MVFusion \cite{wu2023mvfusion} leverages semantic consistency across multi-view images combined with radar features, facilitating robust feature interaction within the fusion transformer layer.
TransCAR \cite{pang2023transcar} introduces an approach that replaces the hard association based on sensor calibration between radar and vision with query soft correlation based on adaptive learning of cross-attention layers.
Recently, the authors in \cite{fent2024dpft} present a novel Dual Perspective Fusion Transformer (DPFT) module that conducts attention fusion between raw radar cube data and image data, thus overcoming the constraints posed by sparse radar point cloud fusion.
CR-DINO \cite{jin2024cr} devises
an innovative radar data representation strategy and introduces two camera-radar fusion frameworks inspired by Swin Transformer \cite{liu2021swin}.

\subsection{Data Alignment}
As a big part in radar-camera fusion, data alignment aims at finding relationships and correspondences between sub-components of instances from two modalities. The development trend of data alignment has evolved from projection-based alignment to DL-based alignment.
For succinctness, we categorize data alignment methods into  \textit{implicit} and \textit{explicit} alignment, also called \textit{hard} and \textit{soft association}. The rationale is that the former is rule-based, while the latter is learning-based.

\subsubsection{Explicit Alignment}
Literally, explicit alignment utilizes fixed extrinsic and intrinsic matrices for explicitly establishing correspondences between radar data and camera images.

\begin{figure}[htbp]
\centering
\includegraphics[width=1\linewidth]{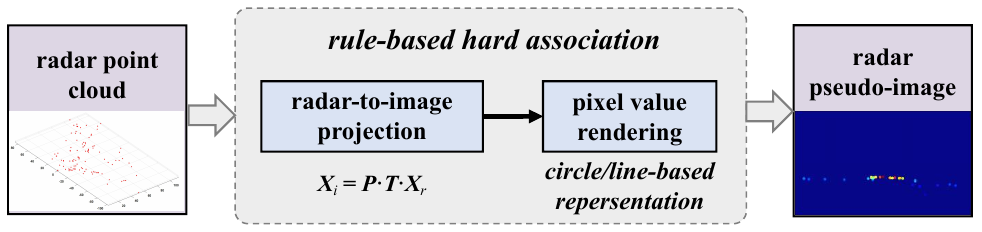}
\caption{An illustration of point-to-pixel projection.}\label{point-to-pixel}
\end{figure}

Owing to the fact that most public multi-sensor datasets (e.g., nuScenes \cite{caesar2020nuscenes}) have supplied the coordinate transformation matrices among all sensing modalities, early works resort to exploring projection-based data transformation for point-to-pixel alignment \cite{chadwick2019distant, nobis2019deep,john2020so,john2019rvnet,li2020feature,chang2020spatial}. In this vein, each radar point sweep is projected onto the pixel plane of the associated camera image, aiming to aggregate the channel feature information (e.g., camera's RGB channels, radar's RCS, position, and velocity channels). 
For easy presentation,  Fig.~\ref{point-to-pixel} depicts an illustration of point-to-pixel projection.  
Herein, $P$ denotes the intrinsic matrix and $T$ refers to the extrinsic matrix.

Unfortunately, due to the inherent discrepancy in angular resolution between radar and camera sensors, severe multi-path interference on radar echoes, and inevitable sensor calibration errors, the radar point cloud measurements projected onto the camera pixel plane tend to deviate significantly from the ideal positions, thus leading to misalignment and performance degradation. Moreover, depth ambiguity and occlusion problems exacerbate the difficulty of monocular perception on the perspective of the camera. As such, subsequent studies have circumvented these challenges by integrating camera and radar feature maps in BEV rather than FV, resulting in enhanced robustness.
Additionally, certain association tasks in rule-based approaches leverage kinematic model-based tracking matching \cite{wang2016road}. 
Nevertheless, this association procedure is manually crafted, relying on minimizing specific distance metrics alongside heuristic rules. This approach not only demands substantial engineering effort and tuning but also faces challenges in adapting to expanding datasets.

\subsubsection{Implicit Alignment}
Overall, multimodal alignment encounters several difficulties: 1) Limited availability of datasets with explicit annotations for alignments; 2) Complexity in devising similarity metrics between modalities; 3) There is a potential existence of multiple feasible alignments, and not all instances in one modality have correspondences in another.

Unlike explicit alignment methods that require predefined alignment instructions or supervised alignment pairs, implicit alignment methods rely on the model's ability to latently learn and discover alignment patterns during the training process.
One of the key advantages of implicit alignment is its adaptability to dynamic environments and changing sensor configurations. In dynamic scenarios where the relative positions and orientations of radar and camera sensors may vary, making explicit alignment information unreliable or even unavailable, implicit alignment enables the fusion model to adapt and align the data without depending on fixed calibration parameters.
Moreover, implicit alignment methods usually leverage advanced techniques such as attention mechanisms to capture fine-grained correlations between the modalities, as well as allow the model to dynamically focus on relevant parts of the input data, enhancing the alignment and fusion process.

Particularly, promising performance has been achieved by transformer-based multi-modal fusion \cite{kim2023craft,wu2023mvfusion,pang2023transcar,fent2024dpft}, which sheds light on the possibility of leveraging the transformer structure as a substitute for manually designed alignment operations.
By endowing the fusion model with the ability to adapt and learn the alignment patterns directly from the dataset, implicit alignment methods can effectively handle challenges such as occlusion, calibration errors, and noise interference commonly encountered in radar-camera fusion applications. This leads to more robust and accurate fusion results, making implicit alignment a valuable approach in advancing the capabilities of multi-modal sensor fusion systems.

\subsection{Fusion Operation}
Fusion operation is a crucial aspect in radar-camera fusion that enables the fusion model to effectively combine information from heterogeneous modalities.
Theoretically, multimodal integration can offer three primary benefits. Firstly, leveraging diverse modal information that refers to the same object could enhance sensing robustness. Secondly, exploiting multiple modalities can enable the capture of complementary information, which may not be discernible within individual modality. Thirdly, a multi-sensor fusion framework might maintain functionality even in the absence or corruption of one sensor.

The evolution of fusion operators has traversed through
simple concatenation \cite{chadwick2019distant,nobis2019deep,lim2019radar}, arithmetic calculation \cite{wang2021rodnet,shuai2021millieye,man2023bev}, Mixture of Experts (MoE) \cite{kim2020grif,kim2018robust,yuksel2012twenty}, and attention mechanisms \cite{chang2020spatial,bijelic2020seeing,cheng2021robust}. Among them, fusion methods by Transformer-based cross-attention demonstrate prominent performance across all benchmarks.
Concretely, fusion operation methods can be conceptually divided into two groups: \textit{conventional} and \textit{adaptive} fusion operation.

\subsubsection{Conventional Fusion Operation}
For concatenation, feature maps provided by heterogeneous modalities are flattened and concatenated along the channel dimension to enrich feature diversity. 
For arithmetic calculation like element-wise addition/multiplication/mean, corresponding elements from different modal feature maps are subjected to addition/multiplication/averaging, resulting in fused feature maps that encompass more comprehensive information. For proposal-wise ensemble operation, it is commonly used in object detection tasks, where distinct modal data are integrated through proposal regions.

\subsubsection{Adaptive Fusion Operation}
\begin{figure}[htbp]
\centering
\includegraphics[width=1\linewidth]{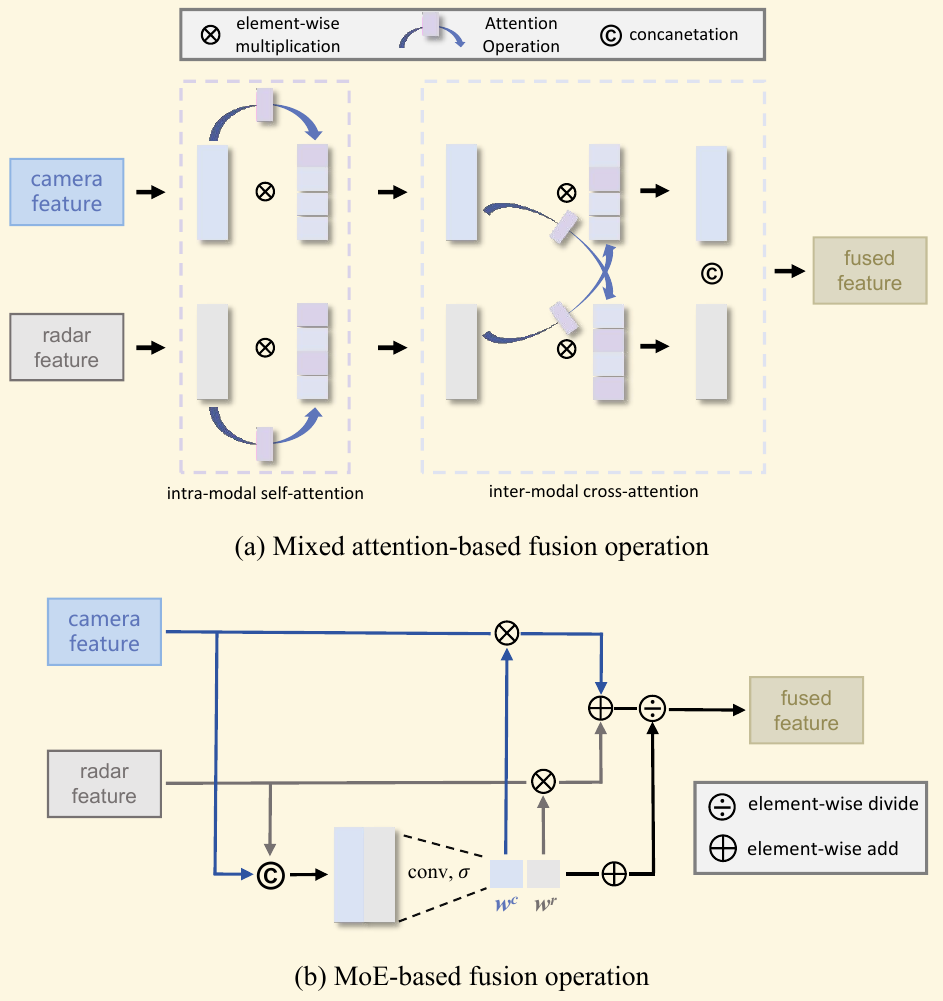}
\caption{A diagram of adaptive fusion operation.}\label{adaptive fusion}
\end{figure}

The aforementioned fusion operations fail to adequately consider the reliability and uncertainty of information from diverse modalities. For instance, the quality of camera data may degenerate in nighttime or rainy/snowy scenes, whereas the radar sensor remains functional properly. 
As such, a bunch of researches involves employing learning-based adaptive fusion methods to substitute rule-based fusion.
For example, MoE strategies \cite{kim2020grif,kim2018robust,yuksel2012twenty} adaptively calculate the weights of multi-modal feature maps via gating networks, and re-weight the confidence of modal information to optimize the fusion operation. Similarly, attention mechanisms empower the network with flexibility by allocating adaptive weights among input modalities to enhance the quality of feature representation. In this vein, representative methods include:
spatial attention \cite{chang2020spatial}, entropy attention \cite{bijelic2020seeing}, mixed attention \cite{cheng2021robust,lu2020milliego},   Squeeze-and-Excitation (SE) attention \cite{hu2018squeeze,gu2022radar,wolters2024unleashing}, Self-Supervised Model Adaptation (SSMA) attention \cite{cui20213d,valada2020self}, cross-attention \cite{pang2023transcar,kim2023craft,kim2023crn}, etc. 
For readability, Fig.~\ref{adaptive fusion} presents a diagram of the two most representative adaptive fusion operations, viz., mixed attention and MoE.

For succinctness, let $\boldsymbol{f}^{r}_{l}$ and $\boldsymbol{f}^{c}_{l}$ represent the feature maps of radar and camera modalities at layer $l$ of the network, and let $T_{l}(\cdot)$ denote the mathematical description of the layer transformation. Then, the typical fusion operations in radar-camera fusion can be summarized as shown in Table \ref{tab:Fusion Operation}.

\begin{table}[h] \scriptsize
\caption{\upshape Details for typical fusion operations} \label{tab:Fusion Operation}
\begin{center}
\renewcommand\tabcolsep{1.2pt}
\begin{tabular}{cc}
\toprule
    \begin{tabular}{c}
       \multicolumn{1}{c}{\textbf{Fusion Operations} }
    \end{tabular}   & 
       \multicolumn{1}{c}{\textbf{Mathematical Descriptions}}
    \\ 
    \midrule
    \begin{tabular}{c}
       \makecell[c]{Element-wise Addition} \\ 
    \end{tabular}   & 
    \begin{tabular}[c]{@{}l@{}}
        $\boldsymbol{f}_l = T_{l-1}\left(\boldsymbol{f}_{l-1}^r + \boldsymbol{f}_{l-1}^c\right)$
    \end{tabular}         \\ 
    \midrule
    \begin{tabular}{c}
        \makecell[c]{Element-wise Multiplication} \\  
    \end{tabular}   &  
    \begin{tabular}[c]{@{}l@{}}
       $\boldsymbol{f}_l = T_{l-1}\left(\boldsymbol{f}_{l-1}^r \otimes \boldsymbol{f}_{l-1}^c\right)$
    \end{tabular}\\
        \midrule
    \begin{tabular}{c}
        \makecell[c]{Channel-wise Concatenation} \\  
    \end{tabular}   &  
    \begin{tabular}[c]{@{}l@{}}
       $\boldsymbol{f}_l = T_{l-1}\left(\boldsymbol{f}_{l-1}^r \oplus \boldsymbol{f}_{l-1}^c\right)$
    \end{tabular}\\ 
        \midrule
    \begin{tabular}{c}
        \makecell[c]{Proposal-wise Ensemble} \\  
    \end{tabular}   &  
    \begin{tabular}[c]{@{}l@{}}
       $\boldsymbol{f}_l = T_{l-1}\left(\boldsymbol{f}_{l-1}^r\right) \cup T_{l-1}\left(\boldsymbol{f}_{l-1}^c\right)$
    \end{tabular}\\ 
        \midrule
    \begin{tabular}{c}
        \makecell[c]{Mixture of Experts} \\  
    \end{tabular}   &  
    \begin{tabular}[c]{@{}l@{}}
       $\boldsymbol{f}_l = T_{l-1}\left(w^r \! \cdot \! \boldsymbol{f}_{l-1}^r + w^c \! \cdot \!\boldsymbol{f}_{l-1}^c\right), w^r\! +\! w^c =1 $
    \end{tabular}\\ 
        \midrule
    \begin{tabular}{c}
        \makecell[c]{Attention Mechanism} \\  
    \end{tabular}   &  
    \begin{tabular}[c]{@{}l@{}}
       $\boldsymbol{f}_{l}=\left(\boldsymbol{att}_{r \rightarrow c} \otimes \boldsymbol{f}_{l-1}^c \right) \oplus \left(\boldsymbol{att}_{c \rightarrow r} \otimes \boldsymbol{f}_{l-1}^r\right)$
    \end{tabular}\\ 
    \bottomrule
\end{tabular}
\end{center}
\end{table}

\subsection{Summary and Lessons Learned}
In this section, we have summarized the key techniques for radar-camera fusion, including sensor calibration, modal representation, data alignment, and fusion operation. 
To sum up, we gather the following lessons learned:

\begin{itemize} 
\item Initially, researchers contemplate utilizing LiDAR-oriented frameworks for sparse radar point clouds. Nonetheless, this proves ineffective due to the distinct characteristics of radar data compared to LiDAR point clouds \cite{fan20244d}. Firstly, radar measurements exhibit notably lower point density than LiDAR point clouds and often lack crucial height information, making them unable to capture precise shape information. 
Secondly, radar point clouds are prone to notable azimuth errors and offer less precision in measuring object surfaces. 
Thirdly, each radar scan typically yields only one radar return at longer ranges, and occasionally no radar return for small objects. This sparse data distribution complicates 3D object detection and tracking by relying solely on radar points.

\item Compared to uni-modal data, multi-modal data necessitates careful consideration of the inherent characteristics of each modality. Camera data typically exhibit structured and regular information, with partial details contributing to the overall image. On the flip side, radar data often contain disordered spatial information, along with sparsity, imprecision, and noise. 
To tackle the sparsity, most work resorts to aggregating multiple radar sweeps to generate denser point clouds \cite{nobis2019deep,kim2020grif}. To address the imprecision, a mass of the literature has explored rectangle \cite{stacker2022fusion} or pillar expansion schemes \cite{nabati2021centerfusion} to enhance the vertical accuracy of radar point clouds. Besides, several works \cite{long2021radar,long2021full,li2024radarcam} make efforts in exploiting the depth and full-velocity information to densify the radar point cloud in a more accurate manner.

\item Generally, incorporating radar data in its BEV representation has demonstrated the highest effectiveness, as observed in prior works \cite{harley2023simple,kim2023crn}. An empirical study by RCBEV \cite{zhou2023bridging} indicates that employing specialized or resource-intensive point cloud processing feature backbones does not yield performance improvements. 
For the noise problem, 
traditional radar denoising approaches rely heavily on CFAR and peak detection algorithms, which lack robust generalization capabilities.
A promising solution is utilizing the DL-based signal processor to substitute CFAR. For instance, \cite{cheng2021new}  pioneers the adaptation of a cross-modal radar detector that leverages LiDAR data, thereby aiding in the removal of noisy points.

\item For data alignment, conventional approaches adopt homogeneous projection with calibration matrices. Nonetheless, these methods struggle to tackle the missing height and imprecise angular location of radar returns, as well as the observation that occluded parts of objects also reflect radar points \cite{li2024farfusion}. Recently, the cross-attention mechanism in the transformer decoder has elicited a significant amount of interest, since it automatically learns the flexible association between radar features and vision-updated queries, surpassing rigid sensor calibration-based associations \cite{pang2023transcar}. 
For fusion operation, traditional radar-camera fusion operators usually perform straightforward concatenation operations, ignoring the spatial connections and semantic alignment between radar and image features \cite{wu2023mvfusion}. 

\end{itemize}

\section{Object Detection with Radar and Camera}
\subsection{Camera-based Object Detection}
Before the widespread adoption of deep learning methods, pioneering object detection algorithms primarily relied on the divide-and-conquer strategy. This involved manually designing feature descriptors (e.g., HOG and Haar) and then combining them with machine learning classifiers (such as SVM and random forests) to search for objects in images that match the designed features.
Unlike traditional image object detection algorithms that require cumbersome manual feature engineering, DL-based methods can fit out complex function models end-to-end, breaking through the bottlenecks of traditional approaches in handling complex and variable scenes, thus making it possible for global optimal solution.

\begin{figure}[h]
	\centering	\includegraphics[width=0.5\textwidth]{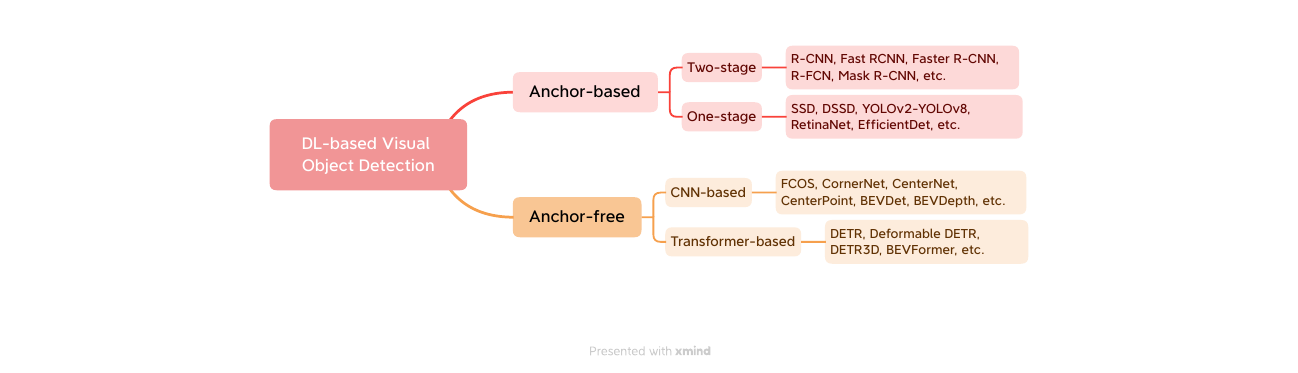}
	\caption{Deep learning-based visual object detection.}
	\label{Visual Object Detection}
\end{figure}

In general, DL-based visual object detection methods typically involve CNNs \cite{krizhevsky2017imagenet} and Transformer models \cite{han2022survey}. Depending on whether the algorithms demand the utilization of anchors, researchers commonly classify them into two main categories: \textit{anchor-based} and \textit{anchor-free}, as shown in Fig.~\ref{Visual Object Detection}.

\subsubsection{Anchor-based Visual Object Detection}
The core idea of anchor-based object detection is to predefine a large number of anchor boxes to cover the objects in the image as completely as possible. This method can be further divided into two patterns, namely \textit{two-stage} and \textit{one-stage}.

\textbf{\emph{a) Two-stage object detection algorithms}} split the algorithmic process into two phases. The first phase is aimed at efficiently selecting all the candidate regions (i.e., region proposals) covering the targets, while the second phase performs further refinement on the generated candidate boxes with more precise regression processing as well as predicts the category of each object.
R-CNN \cite{girshick2014rich}, SPPNet \cite{he2015spatial}, Fast R-CNN \cite{girshick2015fast}, and Faster R-CNN \cite{ren2015faster} serve as seminal networks for two-stage object detectors, followed by a large quantity of variants such as R-FCN \cite{dai2016r}, Mask R-CNN\cite{he2017mask}, Cascade R-CNN \cite{cai2018cascade} and so forth.

\textbf{\emph{b) One-stage object detectors}} predict both the regression parameters of bounding boxes as well as their class probability simultaneously. By omitting the step of candidate region generation, one-stage object detectors tend to run faster, albeit with lower accuracy.
The most common one-stage object detection algorithms include YOLO series 
\cite{redmon2017yolo9000,redmon2018yolov3,bochkovskiy2020yolov4,ge2021yolox,li2022yolov6,wang2023yolov7}, SSD series \cite{liu2016ssd,fu2017dssd,zhang2018single}, RetinaNet \cite{lin2017focal}, EfficientDet\cite{tan2020efficientdet}, etc.

In summary, both two-stage and one-stage object detectors essentially predict based on anchor boxes. In two-stage object detectors, the region proposal generation step acts as a preliminary screening of anchor boxes, alleviating the issue of imbalance between positive and negative samples. In contrast, one-stage object detectors uniformly distribute preset anchor boxes across the entire image, resulting in an imbalance between anchor boxes covering only the background and those involving the objects of interest. Generally, the advantage of two-stage object detectors lies in their high accuracy, while the superiority of one-stage object detectors lies in their fast speed.

\subsubsection{Anchor-free Visual Object Detection}
Anchor-based object detection algorithms involve manually selecting various anchor hyper-parameters (e.g., sizes, aspect ratios, and numbers) based on the distribution characteristics of all objects in the dataset. This process is cumbersome and limits the generalization ability of the object detection networks.
As such, anchor-free object detectors eliminate the need to preset the anchor boxes, naturally circumventing the imbalance between positive and negative samples caused by dense anchor box settings, as well as the computational and memory resources required for calculating the IoU between anchor boxes and the ground truth bounding boxes. Representative detectors in this type of method include FCOS \cite{tian2019fcos}, CornerNet \cite{law2018cornernet}, CenterNet \cite{zhou2019objects}, BEVDet \cite{huang2021bevdet}, BEVDepth \cite{li2023bevdepth}, etc.
It is worth noting that CenterNet obviates the conventional post-processing step of Non-Maximum Suppression (NMS), thus achieving end-to-end object detection strictly. Additionally, CenterNet can be easily extended to 3D object detection or tracking. Subsequent improved algorithms include CenterNetv2\cite{zhou2021probabilistic}, CenterPoint\cite{yin2021center}, CenterTrack\cite{zhou2020tracking}, etc.

Apart from the aforementioned CNN-based work, researchers have also studied end-to-end object detectors based on the Transformer architecture, with DETR \cite{carion2020end} as a groundbreaking work. 
Although DETR achieves comparable accuracy to two-stage object detectors like Faster RCNN, its model training converges slowly, and its detection capability for small objects is inferior. Therefore, Deformable DETR \cite{zhu2020deformable} adopts the core idea of Deformable Convolution Network (DCN) \cite{dai2017deformable} to modify the global attention module in DETR into a sparse local deformable attention mechanism to accelerate algorithm convergence. It also introduces a cross-attention mechanism for information interaction between multi-scale features to improve the detection performance of small objects. In addition, the DETR series of object detectors include PnP-DETR \cite{wang2021pnp}, Conditional DETR \cite{meng2021conditional}, Dynamic DETR \cite{dai2021dynamic}, and so forth.

\subsection{Radar-based Object Detection}
Radar object detection aims to extract valuable target features from noisy echo signals while suppressing clutter interference \cite{gini2008knowledge}. Early radar object detection methods are model-driven and typically involve filtering techniques, such as the well-known CFAR algorithm. The CFAR algorithm follows the Neyman-Pearson Criterion, maximizing detection probability by setting detection thresholds under a fixed false alarm rate premise. 
The traditional model-driven radar object detection process has been elaborated in extensive literature \cite{richards2014fundamentals}, with the core idea being the design of statistical models for radar signals and noise. This involves extracting representative statistical features from radar echo signals and making judgments based on these features to accomplish object detection. However, the complex and diverse clutter distributions in different environments, along with the mismatch between actual clutter distributions and ideal mathematical models, significantly limit the performance of conventional model-driven radar object detection algorithms in real-world scenarios.

\begin{figure}[h]
	\centering	\includegraphics[width=0.5\textwidth]{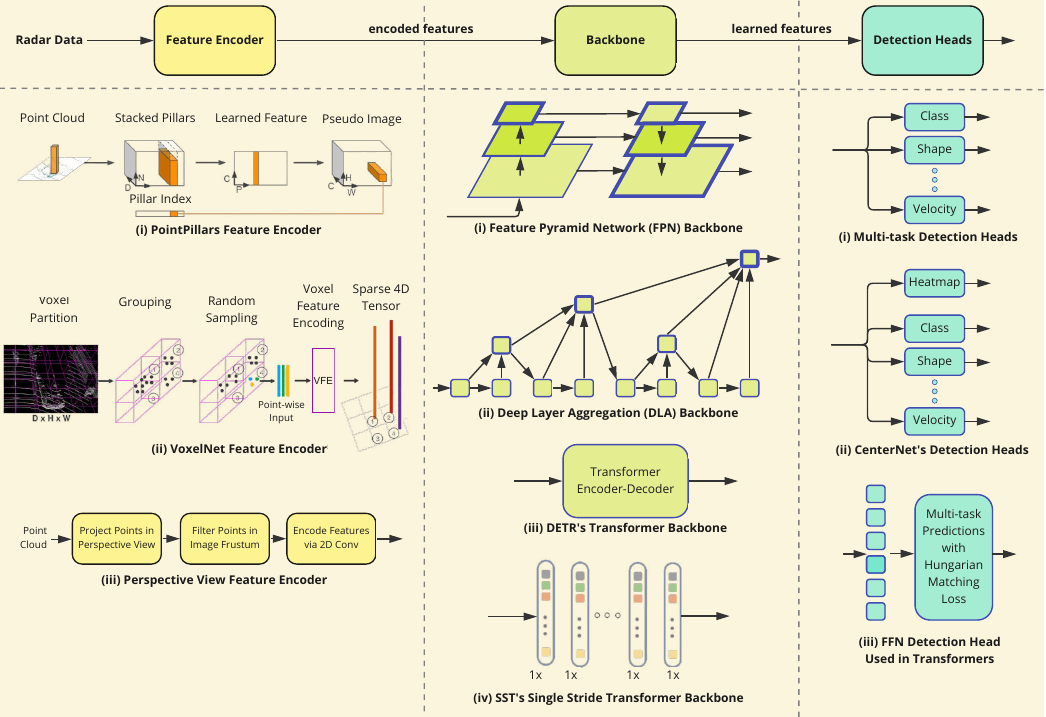}
	\caption{A conceptual block diagram of data-driven radar object detection (Figure from \cite{srivastav2023radars}).}
	\label{radar models}
\end{figure}

Recently, data-driven radar object detection approaches have emerged as crucial avenues supporting practical applications. The main challenges in employing deep learning models lie in representing radar signals in formats suitable for various DL algorithms and devising corresponding network architectures given the fixed format of radar input data (see Fig.~\ref{radar models}).
Currently, most commercial radar products for object detection and tracking are equipped with embedded algorithms for extracting object distance, azimuth, and Doppler velocity. Therefore, the most common data representation of these radars is sparse point cloud. 
Nevertheless, to filter out background and noise interference, radar point cloud is typically generated after processing with peak detection algorithms (e.g., CFAR), potentially causing the loss of certain targets of interest (i.e., objects with weak reflectivity) in the original data. 
Consequently, researchers have begun to explore DL algorithms under data representations that retain more raw information about radar signals, such as ADC data, spectrum/tensor data, and micro-Doppler data.
Among these, neural network algorithms based on radar ADC data generally apply only to classification tasks in closed environments \cite{zhao2023cubelearn}. Meanwhile, deep learning methods utilizing radar micro-Doppler features can only address classification problems such as determining the presence of targets, target categories, or distinguishing activity types \cite{kim2015human,seifert2017new,seifert2019toward}. Therefore, we highlight the radar object detection algorithms based on point cloud data and spectrum/tensor data.

\begin{figure*}[h]
	\centering	\includegraphics[width=1\textwidth]{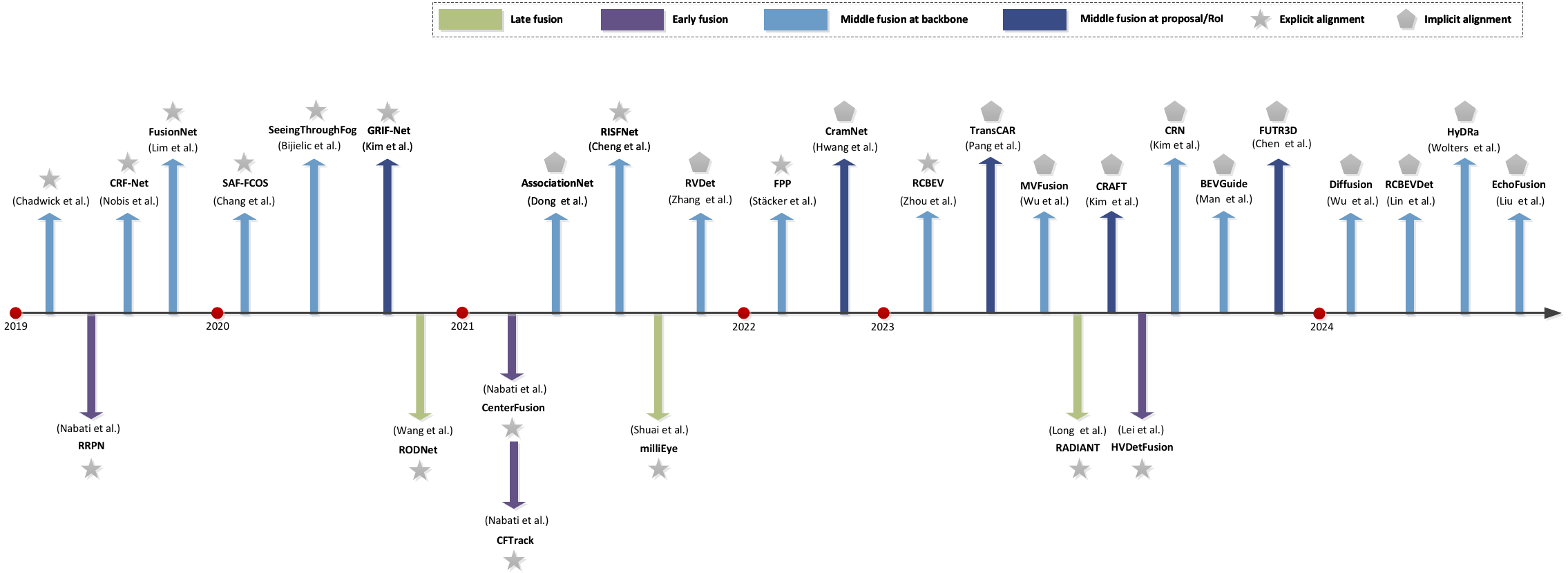}
	\caption{Chronological overview of DL-based radar-vision fusion methods for object detection.}
	\label{RV Fusion}
\end{figure*} 

\subsubsection{Object Detection based on Radar Point Cloud} DL-based object detection research solely relying on radar point cloud data started relatively late compared to its LiDAR counterpart, and initially borrowed from point cloud processing algorithms in the LiDAR domain. Object detection algorithms based on LiDAR point clouds broadly follow three approaches: 1) directly operating on point clouds (i.e., point-wise methods), such as PointNet series \cite{qi2017pointnet,qi2017pointnet++,qi2018frustum} and PointRCNN \cite{shi2019pointrcnn}; 2) projecting point clouds onto BEV or FV planes to generate pseudo images or grid maps, such as BirdNet \cite{beltran2018birdnet} and VeloFCN \cite{li2016vehicle}; 3) voxelizing point clouds in 3D space, like VoxelNet series \cite{zhou2018voxelnet,yan2018second,sindagi2019mvx} and PointPillars series \cite{lang2019pointpillars,paigwar2021frustum}. In addition, new methods are emerging, such as Transformer-based series \cite{mao2021voxel,he2022voxel,wang2022bridged,pan20213d}, graph-based methods \cite{shi2020point,wang2019dynamic}, hybrid methods \cite{shi2020pv,shi2023pv}, etc. 

With regard to radar point clouds, a variety of off-the-shelf point-based networks are adopted for object detection. Point-wise approaches \cite{danzer20192d,tilly2020detection,dubey2022haradnet} process original point clouds directly and capitalize on LiDAR-based algorithms, for example, PointNet \cite{qi2017pointnet}, PointNet++ \cite{qi2017pointnet++} and Frustum PointNets \cite{qi2018frustum}.
In contrast with the computationally intensive point-wise methods, grid-based and voxel-based approaches \cite{scheiner2020seeing,dreher2020radar,kohler2023improved,liu2023smurf} demonstrate superiority in terms of efficiency. This type of method first transforms the point clouds into 2D grid or 3D voxel structures, then corresponding object detectors (e.g., YOLOv3 \cite{redmon2018yolov3}, VoxelNet \cite{zhou2018voxelnet}) are employed. 
Moreover, graph-based radar point cloud object detection methods (e.g., Radar-PointGNN \cite{svenningsson2021radar} and RadarGNN \cite{fent2023radargnn}) leverage GNN to effectively capture spatial relationships and contextual information among the points. Nonetheless, graph construction and feature extraction tend to cause huge computational costs, especially when handling large-scale point clouds.

\subsubsection{Object Detection based on Radar Spectrum/Tensor}
The prevailing radar spectrum/tensor data includes Range-Azimuth (RA) matrix, Range-Doppler (RD) matrix, Range-Azimuth-Doppler (RAD) tensor, and Range-Azimuth-Doppler-Elevation tensor. 
Since radar spectrum resembles the image-like representation, researchers commonly employ image-oriented frameworks to conduct object detection on the RA matrix \cite{gao2019experiments,dong2020probabilistic,wang2021rodnet2}, RD matrix \cite{zhou2019yolo,ng2020range,decourt2022darod}, and RAD tensors \cite{major2019vehicle,palffy2020cnn,zhang2021raddet,gao2020ramp,rebut2022raw}. However, unlike camera image data, radar spectrum data lacks explicit physical interpretation, which poses challenges in transferring learned features from image-customized algorithms to radar data. More importantly, the high-dimensional intrinsic property of radar spectrum, along with the complicated interference, hinder its practical application in real-time scenarios.

\subsection{Radar-Camera Fusion based Object Detection}
The research on the fusion of radar and camera can roughly be divided into traditional methods and learning-based methods.
The fusion architecture of early work \cite{alessandretti2007vehicle,liu2011road,han2016frontal,hsu2019developing,jiang2019target} is relatively succinct and intuitive, typically utilizing radar to assist camera perception. 
For instance, the authors in \cite{wang2016road,han2016frontal} project radar data onto the image plane to generate RoIs containing potential targets. Then, they apply image-oriented active contour methods only within the RoIs for object detection, significantly reducing the complexity. If no valid object is detected within an RoI, the corresponding radar measurement is considered a false alarm and removed. The limitation of this method lies in its dependency on effective measurements from the radar, which affects the recall rate of target detection. If the radar branch fails to detect a valid target, the camera image branch is unable to compensate, potentially leading to missed detection.

With the profound progress of DNNs and the availability of large-scale radar datasets (e.g., nuScenes \cite{caesar2020nuscenes}), an increasing number of studies have been dedicated to problems associated with DL-based radar and vision fusion \cite{deng2022global}. 
For intuitiveness, Fig.~\ref{RV Fusion} summarizes a chronological overview of radar-camera fusion algorithms based on deep learning in recent years, highlighting the fusion stage and data alignment schemes.

Particularly, Chadwick \textit{et al.} \cite{chadwick2019distant} pioneered the adaptation of CNNs into the domain of radar-camera fusion for object detection. They initially project self-collected radar sparse point clouds onto the camera image plane and expand radar points into circles. The resulting radar pseudo-image shares the same dimensions as the camera image, with each pixel value of the pseudo-image obtained through mathematical formulas from the radar distance and velocity information. Subsequently, the preprocessed radar pseudo-image data and camera image data undergo feature extraction using CNNs in separate branches. The two branches are then fused either by channel-wise concatenation or element-wise addition operations. Finally, the fused feature map passes through feature extraction and detection heads to produce the ultimate object detection results. This fusion framework is intuitive and concise, and experimental results validate that the object detection mAP using the radar-camera fusion method surpasses that of methods relying solely on a single camera image.

\subsubsection{Early Fusion-based Object Detection}
Radar-camera early fusion methods aim at integrating the knowledge from radar data into camera data before feeding into an image-based detection pipeline.
These frameworks typically follow a sequential structure and can be categorized into two types: radar-guided and camera-guided knowledge fusion.

\textbf{\emph{a) Camera-guided knowledge fusion}}:
Literally, camera-guided knowledge fusion schemes first extract foreground knowledge from the camera image to refine object candidate regions within radar point clouds. 

\begin{figure}[h]
	\centering	\includegraphics[width=0.48\textwidth]{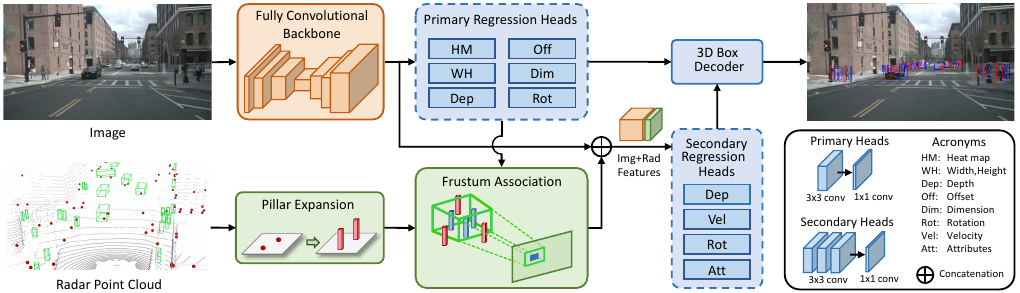}
	\caption{Overview of the CenterFusion model (from \cite{nabati2021centerfusion}).}
	\label{Centerfusion}
\end{figure}

Specifically, an image initially undergoes processing through a 2D object detector to produce 2D bounding boxes. Secondly, these 2D boxes are extended into 3D frustums, which are then regarded as the selected RoI on the radar point cloud to narrow down the search space.
One of the famous radar-camera fusion frameworks following this paradigm is CenterFusion \cite{nabati2021centerfusion}, which might be inspired by Frustum-PointNets \cite{qi2018frustum} and has something in common with Frustum-Pointpillars \cite{paigwar2021frustum}. CenterFusion \cite{nabati2021centerfusion} first employs CenterNet \cite{zhou2019objects} for 3D monocular object detection in the image. Subsequently, it devises a frustum-based association technique to correlate radar information with preliminary image-based objects, generating radar-based heatmaps (using distance and radial velocity measurements, which are two main imperfections of the image-only detector) to augment complementary features for the image features. As shown in Fig.~\ref{Centerfusion}, the combined features are fed into another regression head to enhance the initial detection. It is worth noting that the effectiveness of the fusion architecture depends on the object detection results from the camera.

Afterwards, CFTrack \cite{nabati2021cftrack}, FG-3DMOT \cite{poschmann2021optimization}, and HVDetFusion \cite{lei2023hvdetfusion} extend CenterFusion to the task of object tracking. 
Instead of following CenterFusion's strategy of adopting only the closest radar point, ClusterFusion \cite{kurniawan2023clusterfusion} alternatively  leverages all the radar points within an RoI frustum by both handcrafted and learning-based radar feature extraction strategy. This association mechanism not only matches radar points to preliminary image objects but also clusters radar points into local neighborhoods. Consequently, ClusterFusion is capable of extracting spatial features from the geometry of clusters, which would be unattainable through associating only one radar point per preliminary detection.

\textbf{\emph{b) Radar-guided knowledge fusion}}:
As another radar-camera early fusion pipeline, radar-guided knowledge fusion strategies exploit radar information to refine the object candidate regions within camera images. 
As a pioneering work in this field, RRPN \cite{nabati2019rrpn} generates radar projection-based object proposals to limit the detection area in camera images. 
To ensure the radar points are positioned on or close to the corresponding objects, RRPN employs a strategy where multiple anchors of various sizes and aspect ratios are generated and centered at the points of interest. After that, image-based detection processing is conducted merely within these anchors. Although RRPN can improve the computational efficiency, objects reflecting no radar points might be disregarded.
To this end, BIRANet \cite{yadav2020radar+} propose an improved approach where undetected objects from radar points can be detected.

\subsubsection{Middle Fusion-based Object Detection}
Middle fusion methods merge image and radar features during intermediate stages within fusion network, such as in the network backbone, during proposal generation, or in the RoI refinement phase.
 
The majority of middle-level fusion methods involve projecting the radar point clouds onto another coordinate, such as the BEV plane \cite{zhang2021rvdet,zhou2023bridging,kim2023crn} or the image plane \cite{chadwick2019distant,bijelic2020seeing,dong2021radar}. 
On the one hand, while converting the radar points onto the BEV plane preserves the spatial information within the point cloud and facilitates feature extraction, it necessitates non-trivial process to transform the image features to the BEV plane and fuse them with the radar features.
On the other hand, although projecting the radar points onto the image plane is trivial, it tends to flatten the crucial depth dimension, making it intractable to extract the spatial feature of the radar points.
As illustrated in Fig.~\ref{nuscenes data}, the size and orientation of objects are easily discernible on the BEV plane, whereas the collapsed depth dimension on the image plane presents challenges in achieving clarity.

\textbf{\emph{a) Fusion in backbone networks}}:
Numerous efforts have been devoted to progressively integrating camera and radar features within backbone networks. These methods typically start by establishing point-to-pixel correspondences through radar-to-camera transformations. Subsequently, leveraging these correspondences, features from radar and image backbones are fused via various fusion operations, such as concatenation \cite{stacker2022fusion}, convolutional attention \cite{bijelic2020seeing}, and Transformer \cite{kim2023crn}. 
Alternatively, it can be performed solely at the output feature maps of backbone networks, employing aggregation modules such as learnable alignment \cite{zhang2021rvdet,dong2021radar,wu2023mvfusion,liu2024echoes}. 
Specifically, EchoFusion \cite{liu2024echoes} employs distinct backbones to extract features from range-time radar maps and images. Then, the polar queries align the camera and radar features by cross-attention. 
MVFusion \cite{wu2023mvfusion} integrates semantic alignment into radar features using a semantic-aligned radar encoder to generate image-guided radar features. Next, a radar-guided fusion Transformer is introduced to fuse radar and image features, enhancing the correlation between the two modalities globally through the cross-attention mechanism.
Besides, RVDet \cite{zhang2021rvdet} employs an adaptive projection network to convert image features into the BEV plane of the radar occupancy grid map, aiming to align the two cross-modal features.
In addition, AssociationNet \cite{dong2021radar} explores the radar-camera association problem through deep representation learning, seeking to exploit feature-level interaction.

\begin{figure}[h]
	\centering	\includegraphics[width=0.48\textwidth]{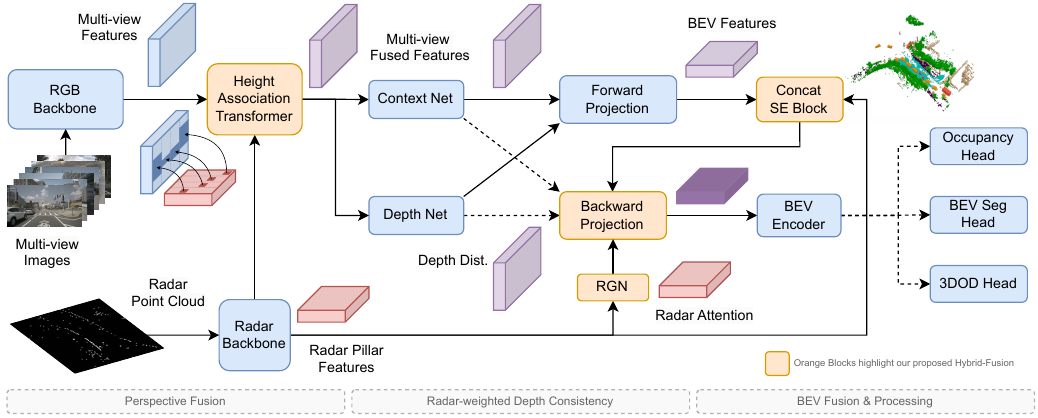}
	\caption{Overview of the HyDRa model (from \cite{wolters2024unleashing}).}
	\label{hyDRa}
\end{figure}

Recently, HyDRa \cite{wolters2024unleashing} significantly outperforms the previous state-of-the-art camera-radar fusion detectors on the nuScenes dataset, and naturally extends to become the first radar-enhanced occupancy prediction framework. An overview of the HyDRa architecture is provided in Fig.~\ref{hyDRa}.
In a nutshell, it presents a height association transformer module to overcome the limitations in previous BEV-generating depth networks, and establish a radar-weighted depth consistency strategy to improve sparse fusion features in the BEV. This addresses challenges like misaligned or unassociated features and occluded objects.
As such, HyDRa possesses the advantage of fully capitalizing on radar’s potential for unified robust depth sensing, thus contributes to improving the detection of low-visible objects, enhancing depth estimation robustness and velocity estimation accuracy. 

\begin{figure}[h]
	\centering	\includegraphics[width=0.48\textwidth]{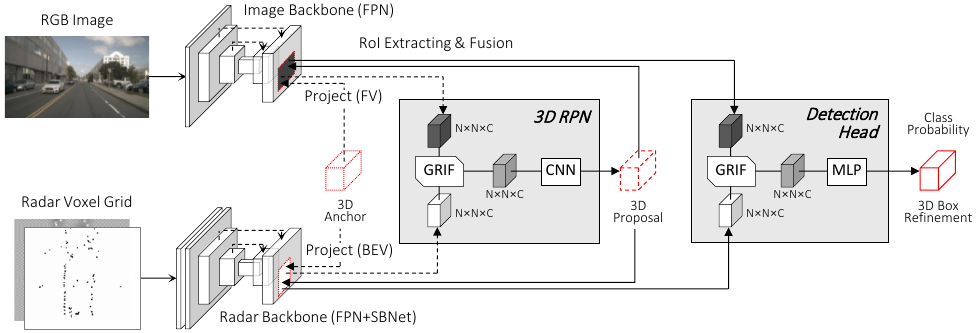}
	\caption{Overview of the GRIF-Net model (from \cite{kim2020grif}).}
	\label{GRIFNet}
\end{figure}

\textbf{\emph{b) Fusion in proposal generation and RoI head}}:
A variety of works combine the heterogeneous features during the proposal generation and RoI refinement phases. The basic idea of this approach derives from pioneering camera-LiDAR fusion frameworks dubbed MV3D \cite{chen2017multi} and AVOD \cite{ku2018joint}, where 3D object proposals are projected into the FV map and the BEV map. This allows for the extraction of features from both the image and LiDAR backbone. The cropped modal features are then combined in an RoI head to predict parameters for each 3D object.
Although this method is capable of producing high-recall proposals, however, the grid representation of radar data is inefficient and inapplicable when the radar point clouds are sparse.
Inspired by AVOD, GRIF-Net \cite{kim2020grif} proposes a gated RoI fusion 
mechanism, essentially utilizing convolutional MoE for adaptively weighting radar and camera features, as shown in Fig.~\ref{GRIFNet}. 
Additionally, Meyer \textit{et al.} \cite{meyer2019deep} also employ the AVOD framework for radar-vision fusion in vehicle detection.
While GRIF-Net adopts MoE for fusion, the authors in \cite{cui20213d} bring in SSMA attention for radar-camera fusion following the AVOD framework.
Moreover, RCBEV \cite{zhou2023bridging} integrates multi-modal features via a two-stage fusion pipeline, namely point-fusion and RoI-fusion, aiming at facilitating more sufficient information interaction during the fusion process.

\begin{figure}[h]
	\centering	\includegraphics[width=0.49\textwidth]{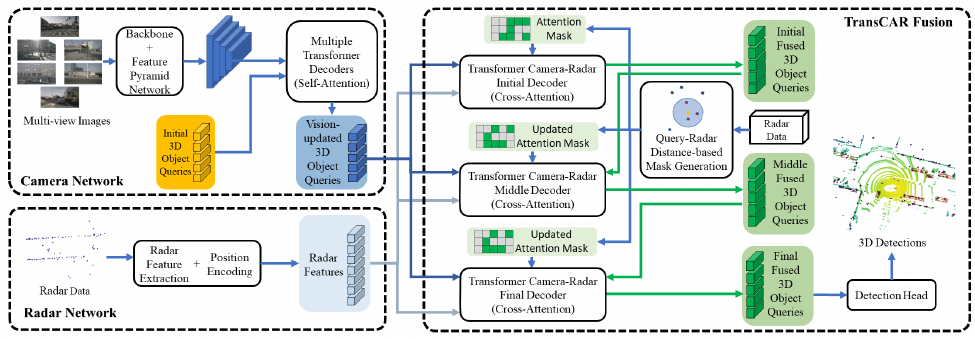}
	\caption{Overview of the TranCAR model (from \cite{pang2023transcar}).}
	\label{TransCAR}
\end{figure}

Unlike the aforementioned frameworks based on CNNs, a growing body of literature has begun to focus on Transformer-based proposal/RoI fusion methods. Through tokenization, Transformer models can easily integrate heterogeneous modalities, enabling adaptive radar-camera fusion in the BEV and reducing spatial misalignment during feature aggregation. For example, CramNet \cite{hwang2022cramnet} advances a ray-constrained cross-attention scheme to resolve the geometric correspondence ambiguities between camera and radar features. The insight is to modify the camera depth estimation by consulting radar features.
Besides, CRAFT \cite{kim2023craft} propose to associate image proposals with radar points in the polar coordinate system, aiming to efficiently manage discrepancies between coordinate systems and spatial properties.
Moreover, TransCAR \cite{pang2023transcar} employs Transformer structure to learn interactions between radar features and image object proposal queries (as shown in Fig.~\ref{TransCAR}), facilitating adaptive camera-radar association. Therein, a novel query-radar attention mask is devised to assist the cross-attention layer to avoid unnecessary interactions between distant image queries and radar features.
Inspired by DETR3D \cite{wang2022detr3d}, FUTR3D \cite{chen2023futr3d} develops a unified multi-sensor fusion framework for 3D object detection. Central to this framework is the query-based Modality-Agnostic Feature Sampler (MAFS), which allows the model to seamlessly operate with various sensor combinations and configurations.

\renewcommand\arraystretch{1}
\begin{table*}
\scriptsize
\centering
\caption{A taxonomy of representative papers for radar-camera fusion-based object detection}
\label{tab:An overview of selected RCF methods}
\setlength\tabcolsep{1pt}
\begin{tabular}{c|c|ccccc|c|c|c}
\hline 
 \multirow{2}{*}{\textbf{Paper}} & \multirow{2}{*}{\textbf{Modal Representation}} & \multicolumn{5}{c|}{\textbf{Fusion Stage}} & \multirow{2}{*}{\textbf{Data Alignment}} & \multirow{2}{*}{\textbf{Fusion Operation}} & \multirow{2}{*}{\begin{tabular}{l} \textbf{Dataset, Application,} \\ \textbf{Performance}\end{tabular}}  \\ \cline{3-7} 
 &   & \multicolumn{1}{c|}{Early} & \multicolumn{1}{c|}{Backbone} & \multicolumn{1}{c|}{Proposal} & \multicolumn{1}{c|}{RoI} & \multicolumn{1}{c|}{Late} & &  \\ \hline

 \specialrule{0.1em}{1pt}{1pt}
\makecell[c]{Oxford \cite{chadwick2019distant},\\ ICRA'19} & \begin{tabular}{l} 
R: PC-based pseudo-image \\ C: FV image features
\end{tabular}
&  \multicolumn{1}{c|}{}      & \multicolumn{1}{c|}{$\surd$}         & \multicolumn{1}{c|}{}         & \multicolumn{1}{c|}{}    &                               & $\mathrm{R}\rightarrow \mathrm{C}$ & \makecell[c]{concatenation /\\ element-wise addition} & \makecell[c]{self-collected, AD \\ AP: 0.279} \\ 

\hline \makecell[c]{CRF-Net \cite{nobis2019deep},\\SDF'19} & \begin{tabular}{l} R: PC-based pseudo-image \\ C: FV image features \end{tabular} & \multicolumn{1}{c|}{}      & \multicolumn{1}{c|}{$\surd$}         & \multicolumn{1}{c|}{}         & \multicolumn{1}{c|}{}    &                            & $\mathrm{R}\rightarrow \mathrm{C}$ & \makecell[c]{multi-level \\ concatenation} & \makecell[c]{modified nuScenes, AD \\ mAP: 0.439} \\

\hline \makecell[c]{RRPN \cite{nabati2019rrpn}, ICIP'19} & \begin{tabular}{l}R: PC-based RoI, C: FV image \end{tabular} & \multicolumn{1}{c|}{$\surd$}      & \multicolumn{1}{c|}{}         & \multicolumn{1}{c|}{}         & \multicolumn{1}{c|}{}    &                             & $\mathrm{R} \rightarrow \mathrm{C}$ & RoI selection & \makecell[c]{modified nuScenes, AD
\\ AP: 0.430, AR: 0.355} \\

 \hline \makecell[c]{FusionNet \cite{lim2019radar},\\ NeurIPS'19} & \begin{tabular}{l} R: RA spectrum features\\ C: BEV image features \end{tabular} &\multicolumn{1}{c|}{}      & \multicolumn{1}{c|}{$\surd$}         & \multicolumn{1}{c|}{}         & \multicolumn{1}{c|}{}    &                              & $\mathrm{C} \rightarrow \mathrm{R}$ & concatenation & \makecell[c]{self-collected, AD \\mAP: 0.735} \\
 
\hline \makecell[c]{SeeThroughFog \cite{bijelic2020seeing},\\ CVPR'20} & \begin{tabular}{l} R: PC-based pseudo-image \\ C: FV image features \end{tabular} & \multicolumn{1}{c|}{}      & \multicolumn{1}{c|}{$\surd$}         & \multicolumn{1}{c|}{}         & \multicolumn{1}{c|}{}    &                              & $\mathrm{R} \rightarrow \mathrm{C}$ & entropy attention & \makecell[c]{Dense, AD\\AP: 0.907}\\

\hline \makecell[c]{GRIF-Net \cite{kim2020grif},\\ IROS'20} &  \begin{tabular}{l} R: PC-based BEV gridmap \\ C: FV image features \end{tabular} & \multicolumn{1}{c|}{}      & \multicolumn{1}{c|}{}         & \multicolumn{1}{c|}{$\surd$}         & \multicolumn{1}{c|}{$\surd$}    &                              & \makecell[c]{feature map\\ encoding}  & \makecell[c]{convolutional MoE} & \makecell[c]{modified nuScenes, AD \\AP: 0.441} \\ 

\hline \makecell[c]{RODNet \cite{wang2021rodnet},\\ WACV'21} & \begin{tabular}{l} R: RA spectrum heatmaps \\ C: image detection heatmaps \end{tabular}  & \multicolumn{1}{c|}{}      & \multicolumn{1}{c|}{}         & \multicolumn{1}{c|}{}         & \multicolumn{1}{c|}{}    &                  $\surd$            & $\mathrm{C} \rightarrow \mathrm{R}$ & element-wise product & \makecell[c]{CRUW, AD\\AP: 0.837, AR: 0.856} \\
\hline \makecell[c]{CenterFusion \cite{nabati2021centerfusion},\\ WACV'21} & \begin{tabular}{l}R: PC-based heatmaps  \\ C: image heatmaps \end{tabular}  & \multicolumn{1}{c|}{$\surd$}      & \multicolumn{1}{c|}{$\surd$}         & \multicolumn{1}{c|}{}         & \multicolumn{1}{c|}{}    &                              & \makecell[c]{frustum-based\\ association} & \makecell[c]{RoI selection $\&$\\ concatenation} & \makecell[c]{nuScenes, AD \\ mAP: 0.326, NDS: 0.449} \\

\hline \makecell[c]{AssociationNet \cite{dong2021radar},\\CVPR'21} & \begin{tabular}{l} R: PC-based pseudo-image  \\ C: FV image features \end{tabular}  & \multicolumn{1}{c|}{}      & \multicolumn{1}{c|}{$\surd$}         & \multicolumn{1}{c|}{}         & \multicolumn{1}{c|}{}    &                              &  \makecell[c]{$\mathrm{R} \rightarrow \mathrm{C}$ $\&$\\  representation learning}  & concatenation & \makecell[c]{Self-collected, AD \\ Precision: 0.91, F1: 0.92} \\

\hline \makecell[c]{SSMA \cite{cui20213d},\\ ITSC'21} & \begin{tabular}{l} R: PC-based FV/BEV features\\ C: FV image features \end{tabular}  & \multicolumn{1}{c|}{}      & \multicolumn{1}{c|}{}         & \multicolumn{1}{c|}{$\surd$}         & \multicolumn{1}{c|}{$\surd$}    &                               & $\mathrm{R} \rightarrow  \mathrm{C}$ & \makecell[c]{self-supervised \\ model adaptation} & \makecell[c]{Astyx, AD \\ mAP: 0.690} \\ 

\hline \makecell[c]{RISFNet \cite{cheng2021robust},\\ ICCV'21} & \begin{tabular}{l} R: PC density map\\ C: FV image features \end{tabular}  & \multicolumn{1}{c|}{}      & \multicolumn{1}{c|}{$\surd$}         & \multicolumn{1}{c|}{}         & \multicolumn{1}{c|}{}    &                              & $\mathrm{R} \rightarrow \mathrm{C}$ & self + global attention & \makecell[c]{self-collected, USV \\
AP$^{35}$: 0.90, AP$^{50}$: 0.75}\\ 

\hline \begin{tabular}{l} milliEye \cite{shuai2021millieye},\\ loTDI'21 \end{tabular}  & \begin{tabular}{l} R: PC confidence score \\ C: image confidence
score \end{tabular} & \multicolumn{1}{c|}{}      & \multicolumn{1}{c|}{}         & \multicolumn{1}{c|}{}         & \multicolumn{1}{c|}{}    &               $\surd$               & $\mathrm{R} \rightarrow \mathrm{C}$& \makecell[c]{element-wise addition \\ $\&$ sigmoid layer} & \makecell[c]{self-collected, HS \\ mAP: 0.835} \\

\hline \makecell[c]{RVDet \cite{zhang2021rvdet},\\ ITSC'21} & \begin{tabular}{l} R: PC-based BEV gridmap \\ C:  BEV image features\end{tabular} & \multicolumn{1}{c|}{}      & \multicolumn{1}{c|}{$\surd$}         & \multicolumn{1}{c|}{}         & \multicolumn{1}{c|}{}    &                              & \makecell[c]{$\mathrm{C}\rightarrow \mathrm{R} $ through \\  MLP-based projection} & concatenation & \makecell[c]{self-collected, AD  \\ AP: 0.785} \\ 

\hline \makecell[c]{FPP \cite{stacker2022fusion},\\ WACV'22} & \begin{tabular}{l} R: PC-based pseudo-image \\ C: FV image features \end{tabular} & \multicolumn{1}{c|}{}      & \multicolumn{1}{c|}{$\surd$}         & \multicolumn{1}{c|}{}         & \multicolumn{1}{c|}{}    &                              & $\mathrm{R} \rightarrow \mathrm{C}$ & \makecell[c]{multi-level \\ concatenation} & \makecell[c]{modified nuScenes, AD  \\
mAP: 0.467}\\ \hline 

\makecell[c]{CramNet\cite{hwang2022cramnet},\\ECCV'22} & \begin{tabular}{l} R: Scanning RA images \\ C: BEV image features \end{tabular} & \multicolumn{1}{c|}{}      & \multicolumn{1}{c|}{}  & \multicolumn{1}{c|}{$\surd$}         & \multicolumn{1}{c|}{}    &          & \makecell[c]{cross-view \\ matching} & \makecell[c]{ray-constrained \\ cross-attention} & \makecell[c]{RADIATE, AD \\AP: 0.621} \\ \hline 

\makecell[c]{RADIANT \cite{long2023radiant},\\ AAAI'23} & \begin{tabular}{l} R: PC-based pseudo-image features  \\ C: FV image features \end{tabular} & \multicolumn{1}{c|}{}  & \multicolumn{1}{c|}{$\surd$}  & \multicolumn{1}{c|}{} & \multicolumn{1}{c|}{} & $\surd$ & \makecell[c]{$\mathrm{R} \rightarrow \mathrm{C}$ $\&$ \\ detection association} & \makecell[c]{depth fusion} & \makecell[c]{nuScenes, AD  \\ mAP: 0.380} \\ \hline 

\makecell[c]{FUTR3D \cite{chen2023futr3d},\\ CVPR'23} & \begin{tabular}{l} R: PC-based sampling features \\ C: MV image sampling features \end{tabular} & \multicolumn{1}{c|}{} & \multicolumn{1}{c|}{} & \multicolumn{1}{c|}{$\surd$} & \multicolumn{1}{c|}{} &  & \makecell[c]{modality-agnostic \\ feature sampler} & \makecell[c]{concatenation $\&$  \\ unified object queries} & \makecell[c]{nuScenes, AD  \\ mAP: 0.350, NDS: 0.459} \\ \hline 

\makecell[c]{RCBEV\cite{zhou2023bridging},\\ TIV'23} & \begin{tabular}{l} R: PC-based BEV features  \\ C: BEV image features \end{tabular} & \multicolumn{1}{c|}{}      & \multicolumn{1}{c|}{$\surd$}  & \multicolumn{1}{c|}{}         & \multicolumn{1}{c|}{$\surd$}    &          & \makecell[c]{$\mathrm{C} \rightarrow \mathrm{R}$ via \\  lift-splat} & \makecell[c]{concatenation $\&$ \\ element-wise product}& \makecell[c]{nuScenes, AD  \\
mAP: 0.406, NDS: 0.486} \\ \hline 

\makecell[c]{TransCAR \cite{pang2023transcar},\\ IROS'23} & \begin{tabular}{l} R: PC-based BEV features  \\ C: MV image-updated object queries\end{tabular} & \multicolumn{1}{c|}{} & \multicolumn{1}{c|}{}  & \multicolumn{1}{c|}{$\surd$}  & \multicolumn{1}{c|}{}  &  & \makecell[c]{soft-association} & \makecell[c]{Transformer \\ cross-attention decoder} & \makecell[c]{nuScenes, AD  \\ mAP: 0.422, NDS: 0.522} \\ \hline 

\makecell[c]{BEVGuide \cite{man2023bev},\\ CVPR'23} & \begin{tabular}{l} R: PC-based BEV features  \\ C: image BEV features \end{tabular} & \multicolumn{1}{c|}{}      & \multicolumn{1}{c|}{$\surd$}         & \multicolumn{1}{c|}{}         & \multicolumn{1}{c|}{}    &                              & \makecell[c]{sensor-agnostic attention} & \makecell[c]{element-wise operator} & \makecell[c]{nuScenes, AD  \\ mAP: 0.421, NDS: 0.537} \\  \hline

\makecell[c]{MVFusion\cite{wu2023mvfusion},\\ ICRA'23} & \begin{tabular}{l} R: PC-based pseudo-image features \\ C: MV image features \end{tabular} & \multicolumn{1}{c|}{}& \multicolumn{1}{c|}{$\surd$}& \multicolumn{1}{c|}{}  & \multicolumn{1}{c|}{} &   & \makecell[c]{semantic-aligned \\ radar encoder} & \makecell[c]{radar-guided \\fusion transformer} & \makecell[c]{nuScenes, AD \\ mAP: 0.453, NDS: 0.517} \\ \hline 

\makecell[c]{CRAFT \cite{kim2023craft},\\ AAAI'23} & \begin{tabular}{l} R: PC-based BEV features \\ C: MV image proposal features \end{tabular} & \multicolumn{1}{c|}{}      & \multicolumn{1}{c|}{}         & \multicolumn{1}{c|}{$\surd$}         & \multicolumn{1}{c|}{}    &                              & \makecell[c]{point query\\ soft polar association} & cross attention & \makecell[c]{nuScenes, AD  \\ mAP: 0.411, NDS: 0.523} \\

\hline \makecell[c]{CRN \cite{kim2023crn},\\ ICCV'23} & \begin{tabular}{l} R: PC-based BEV features \\ C: MV image BEV features \end{tabular} &\multicolumn{1}{c|}{}      & \multicolumn{1}{c|}{$\surd$}         & \multicolumn{1}{c|}{}         & \multicolumn{1}{c|}{}    &                              &\makecell[c]{point query deformable\\cross attention} & \makecell[c]{deformable \\ cross attention} & \makecell[c]{nuScenes, AD \\ mAP: 0.575, NDS: 0.624} \\ \hline

\makecell[c]{HVDetFusion \cite{lei2023hvdetfusion},\\ arXiv'23} & \begin{tabular}{l} R: PC-based BEV features \\ C: image heatmaps \end{tabular} & \multicolumn{1}{c|}{$\surd$}      & \multicolumn{1}{c|}{$\surd$}         & \multicolumn{1}{c|}{}         & \multicolumn{1}{c|}{}    &                              & \makecell[c]{PC selection by \\ imaged-based detection} & \makecell[c]{RoI selection $\&$\\ concatenation} & \makecell[c]{nuScenes, AD \\ mAP: 0.609, NDS: 0.674}  \\ \hline 

\makecell[c]{Diffusion \cite{wu2024robust},\\ WACV'24} & \begin{tabular}{l} R: PC-based pseudo-image features  \\ C: MV image features \end{tabular} & \multicolumn{1}{c|}{}  & \multicolumn{1}{c|}{$\surd$}  & \multicolumn{1}{c|}{} & \multicolumn{1}{c|}{}  &  & \makecell[c]{global-aware association\\ with denoising diffusion} & \makecell[c]{point-wise addition \\ $\&$ cross attention} & \makecell[c]{nuScenes, AD \\ mAP: 0.502, NDS: 0.594}\\ \hline 

\makecell[c]{RCBEVDet \cite{lin2024rcbevdet},\\ CVPR'24} & \begin{tabular}{l} R: PC-based BEV features  \\ C: MV image features \end{tabular} & \multicolumn{1}{c|}{}  & \multicolumn{1}{c|}{$\surd$}  & \multicolumn{1}{c|}{} & \multicolumn{1}{c|}{}  &  & \makecell[c]{deformable\\ cross-attention} & \makecell[c]{cross-attention} & \makecell[c]{nuScenes, AD \\ mAP: 0.550, NDS: 0.639} \\ \hline 

\makecell[c]{HyDRa \cite{wolters2024unleashing},\\ arXiv'24} & \begin{tabular}{l} R: PC-based BEV voxel features \\ C: MV image BEV features \end{tabular} & \multicolumn{1}{c|}{}      & \multicolumn{1}{c|}{$\surd$}  & \multicolumn{1}{c|}{}         & \multicolumn{1}{c|}{}    &          & \makecell[c]{height association \\$\&$ cross-attention} & \makecell[c]{concatenation  \\ $\&$ SE attention} & \makecell[c]{nuScenes, AD \\ mAP: 0.574, NDS: 0.642}\\ \hline 

\makecell[c]{EchoFusion \cite{liu2024echoes},\\ NeurIPS'24} & \begin{tabular}{l} R: range-time tensor in BEV features\\ C:  image BEV features \end{tabular} & \multicolumn{1}{c|}{}      & \multicolumn{1}{c|}{$\surd$}  & \multicolumn{1}{c|}{}         & \multicolumn{1}{c|}{}    &          & \makecell[c]{polar aligned attention} & \makecell[c]{cross-attention} & \makecell[c]{RADIal, AD, AP: 0.849 \\
K-Radar, AD, AP: 0.699}\\ \hline 

\multicolumn{10}{l}{\scriptsize  R: Radar, C: Camera, PC: Point Cloud, $\rightarrow$: Projection, MV: Multi-View, AD: Autonomous Driving, USV: Unmanned Surface Vehicle, HS: Human Sensing}\\
\end{tabular}
\end{table*}

\renewcommand\arraystretch{1.2}
\begin{table*}[h]
 \centering
  \caption{A numerical comparison of object detection methods on nuScenes test set} \label{tab:numerical comparison}
  \setlength\tabcolsep{11pt}
\begin{tabular}{c|c|c|c||c|c|c}
\hline \textbf{Methods} & \textbf{Modality Input} & \textbf{Backbone} &  \textbf{Image Size} & \textbf{NDS} $\uparrow$ & \textbf{mAP} $\uparrow$ & \textbf{FPS} \\
\hline PointPillars \cite{lang2019pointpillars} & LiDAR & Pillars & - & 0.453 & 0.305 & 61 \\
CenterPoint \cite{yin2021center} & LiDAR & Voxel & - & 0.673 & 0.603 & 30 \\
\hline KPConvPillars \cite{ulrich2022improved} & Radar & Pillars & - &  0.049 & 0.139 & - \\
\hline CenterNet \cite{zhou2019objects} & Camera & DLA34 \cite{yu2018deep} & 450 $\times$ 800 & 0.400 & 0.338 & - \\
CRAFT \cite{kim2023craft} & Camera$+$Radar & DLA34 \cite{yu2018deep} & 448 $\times$ 800 & $0.523 (+30.8 \%)$ & $0.411 (+21.6 \%)$ & 4.1 \\
\hline PETR \cite{liu2022petr} & Camera & V2-99 \cite{lee2019energy} & 900 $\times$ 1600& 0.504 & 0.441 & - \\
 MVFusion \cite{wu2023mvfusion} & Camera$+$Radar & V2-99 \cite{lee2019energy} & 900 $\times$ 1600 & $0.517 (+2.6 \%)$ & $0.453 (+2.7 \%)$ & - \\
\hline DETR3D \cite{wang2022detr3d} & Camera & V2-99 \cite{lee2019energy} & 900 $\times$ 1600 & 0.479 & 0.412 & - \\
 TransCAR \cite{pang2023transcar} & Camera$+$Radar & V2-99 \cite{lee2019energy} & 900 $\times$ 1600 & $0.522 (+9.0 \%)$ & $0.422 (+2.4 \%)$ & - \\
\hline BEVDepth \cite{li2023bevdepth} & Camera & V2-99 \cite{lee2019energy} & 900 $\times$ 1600 & 0.605 & 0.515 & - \\
 RCBEVDet \cite{lin2024rcbevdet} & Camera$+$Radar & V2-99 \cite{lee2019energy} & 900 $\times$ 1600 & $0.639 (+5.6 \%)$ & $0.550 (+6.8 \%)$ & - \\
\hline BEVDepth \cite{li2023bevdepth} & Camera & ConvNeXt-B \cite{liu2022convnet} & 640 $\times$ 1600 & 0.609 & 0.520 & 5.0 \\
 CRN \cite{kim2023crn} & Camera$+$Radar & ConvNeXt-B \cite{liu2022convnet} & 640 $\times$ 1600 & $0.624 (+2.5 \%)$ & $0.575 (+10.6 \%)$ & 7.2 \\
\hline BEVFormerV2 \cite{yang2023bevformer} & Camera & InternImage-B \cite{wang2023internimage} & 640 $\times$ 1600 & 0.620 & 0.540 & - \\
HVDetFusion \cite{lei2023hvdetfusion} & Camera$+$Radar & InternImage-B \cite{wang2023internimage} & 640 $\times$ 1600 & $0.674 (+8.7 \%)$ & $0.609 (+12.8 \%)$ & -- \\
\hline StreamPETR \cite{wang2023exploring} & Camera & V2-99 \cite{lee2019energy} & 900 $\times$ 1600 & 0.636 & 0.550 & - \\
 HyDRa \cite{wolters2024unleashing} & Camera$+$Radar & V2-99 \cite{lee2019energy} & 900 $\times$ 1600 & $0.642 (+0.9 \%)$ & $0.574 (+4.4 \%)$ & - \\
\hline
\end{tabular}
\end{table*}

\renewcommand\arraystretch{1.5}
\begin{table*}[h]
 \centering
  \caption{A summary of radar, camera, and LiDAR sensing methods} \label{tab:multi-modal}
\begin{tabular}{|m{2cm}<{\centering}|m{5cm}|m{3.9cm}|m{5.4cm}|}
    \hline \rowcolor{gray!40} 
    \textbf{Sensing Method} & \multicolumn{1}{c|}{\textbf{Strengths}} & \multicolumn{1}{c|}{\textbf{Weaknesses}} & \multicolumn{1}{c|}{\textbf{Accuracy}} \\
    \hline
    Radar-only & High robustness at long distances and under poor visibility conditions. & Low resolution in identifying the shape and size of objects, struggles with non-metallic objects. & Suitable for detecting object presence and velocity, lower accuracy in localization and classification (mAP$\approx$0.20, nuScenes dataset). \\
    \hline
    LiDAR-only & Provides high-resolution 3D point clouds, ideal for mapping and detection. & Performance degrades in adverse weather conditions, struggles with expensive costs. & High accuracy in object detection and classification (mAP$\approx$0.70, nuScenes dataset), accurate depth estimation. \\
    \hline
    Camera-only & High resolution, capable of detecting colors and textures, useful for object classification and environmental understanding. & Sensitive to lighting and  weather conditions, limited depth perception. & High classification accuracy under good conditions (mAP$\approx$0.60, nuScenes dataset), less accurate depth estimation. \\
    \hline
    Radar-Camera Fusion & Combines radar's robustness and long-range detection capabilities with the camera's high resolution and classification abilities. & Fusion complexity, sensor calibration issues. & Improved accuracy in object detection and classification compared to radar-only or camera-only systems (mAP$\approx$0.67, nuScenes dataset). \\
    \hline
    LiDAR-Camera Fusion & Combines LiDAR's depth information with camera's color and texture data, providing high-resolution 3D understanding. & Fusion complexity, sensor synchronization issues. & Provides the highest accuracy in object detection, classification, and localization (mAP$\approx$0.78, nuScenes dataset). \\
    \hline
\end{tabular}
\end{table*}

\subsubsection{Late Fusion based Object Detection}
In late fusion methods, multi-modal fusion operation is performed on the outputs, which are the detection results obtained from separate radar-only and image-only object detectors.
For example, RODNet \cite{wang2021rodnet2} fuses the likelihood probability heatmaps of radar and camera detections by element-wise production.
A substantial amount of improvement work has been proposed successively based on RODNet \cite{sun2021squeeze,zheng2021scene,wang2021rethinking,yu2021radar,jiang2022t,li2023improving}.
Besides, milliEye \cite{shuai2021millieye} combines the object confidence scores of
radar and camera prediction by addition and a sigmoid layer.
Moreover, RADIANT \cite{long2023radiant} conducts radar-camera depth fusion at both the detection level, where a camera detection is associated with the radar detection if the predicted classes of the two modalities match, and the projected centers as well as the depths are in close proximity. RADIANT emphasizes  utilizing radar data exclusively for improving depth estimation, given that radar's superior depth accuracy is a key advantage over cameras, while other aspects of detection such as object dimensions and image position are unlikely to benefit from radar.

\subsection{Summary and Lessons Learned}
In this section, we have conducted discussions on object detection based on radar and camera. A taxonomy of radar-camera fusion object detection methods is in Table \ref{tab:An overview of selected RCF methods}. The lessons learned can be summarized as follows:

\begin{itemize}
\item  The primary obstacle to monocular vision 3D object detection involves resolving the 2D–3D projection ambiguity due to the absence of precise 3D measurements. Recent strategies, using single-view image input, have employed geometric constraint regularization \cite{liu2022learning} and depth estimation interaction \cite{reading2021categorical} to aid in 3D object detection. Conversely, multi-view 3D object detectors aim to forecast the 3D bounding boxes and categories of objects by utilizing 2D-3D lift \cite{philion2020lift} or querying 2D from 3D \cite{li2022bevformer}.

\item The accuracy of object detection based on sparse point clouds from radar remains insufficient (e.g., in nuScenes detection leaderboard, radar-only methods achieve a maximum mAP of 0.205, while camera-only and LiDAR-only methods can reach 0.668 and 0.705, respectively), and it is challenging to compensate for missed detection caused by filtering. Algorithms based on radar spectrum data can uncover richer underlying information but require larger memory and computational resources \cite{liu2024echoes}. Additionally, they face difficulties in ground truth annotation, often necessitating additional cross-modal supervision \cite{ding2023hidden}.

\item Generally, camera-only object detection approaches tend to outperform their radar-only counterparts significantly. Therefore, state-of-the-art approaches primarily rely on image-based 3D object detectors and aim to integrate radar information across various stages of a camera detection pipeline \cite{wang2024end}. 
Table~\ref{tab:numerical comparison} clearly presents the numerical results on the radar-camera fusion's advantages over the camera-only methods. Besides, in view of the complexity of radar-based and camera-based detection systems, combining the two modalities together generally brings additional computational overhead and inference time latency. Therefore, how to efficiently fuse multi-modal information remains an open challenge. 

\item Radar-camera fusion-based object detection faces challenges due to the inherent characteristics of radar data, such as clutter interference and sparsity, leading to uncertainty issues. As a contrast, LiDAR-camera fusion perception is generally more accurate under normal weather, while radar data may reduce the confidence of fused data if included indiscriminately in the fusion processes. Consequently, prevalent publications like TransFusion \cite{bai2022transfusion}, BEVFusion \cite{liang2022bevfusion}, and DeepFusion \cite{li2022deepfusion} rely solely on camera images and LiDAR point clouds for object detection. 
For readability, Table~\ref{tab:multi-modal} provides a summary of the comparison between radar-only, LiDAR-only, camera-only, radar-camera fusion, LiDAR-camera fusion based sensing methods in terms of strengths, weaknesses, and accuracy.

\begin{table*}[h!]
\centering
\caption{Comparison of post-processing data types and fusion techniques} \label{tab:post-processing}
\begin{tabular}{|m{2.5cm}|m{4.5cm}|m{4.5cm}|m{4.5cm}|}
\hline \rowcolor{gray!40} 
\multicolumn{1}{|c|}{\textbf{Fusion Technique}}       & \multicolumn{1}{c|}{\textbf{Performance}}         & \multicolumn{1}{c|}{\textbf{Advantages} }                                             & \multicolumn{1}{c|}{\textbf{Disadvantages} }                                          \\ \hline
\begin{tabular} c \\ Point cloud fusion \end{tabular}    & 
\begin{tabular}[c]{@{}p{4.5cm}@{}}- High precision: Provides detailed 3D spatial information.\\- High computational cost: requires significant processing power for high-resolution data.\end{tabular} & 
\begin{tabular}[c]{@{}p{4.5cm}@{}}- Rich spatial information: excellent for 3D detection and scene reconstruction.\\- Facilitating multi-modal fusion with LiDAR data.\end{tabular} & 
\begin{tabular}[c]{@{}p{4.5cm}@{}}- Large data volume: can be demanding in terms of storage and processing.\\- Sparse or noisy data: less effective in modality-missing or severe clutter conditions.\end{tabular} \\ \hline

\begin{tabular} c \\ ROI fusion \end{tabular}   & 
\begin{tabular}[c]{@{}p{4.5cm}@{}}- Improved accuracy: focuses on relevant regions, potentially enhancing target detection.\\- Increased efficiency: reduces the amount of data processed, speeding up computations.\end{tabular} & 
\begin{tabular}[c]{@{}p{4.5cm}@{}}- High processing efficiency: speeds up the fusion process by focusing only on relevant data.\\- Flexibility: can dynamically adjust ROIs based on specific tasks or conditions.\end{tabular} & 
\begin{tabular}[c]{@{}p{4.5cm}@{}}- Information loss: potentially misses important context outside the ROI.\\- Dependence on preliminary detection: accuracy is dependent on the effectiveness of initial detection steps.\end{tabular} \\ \hline

\begin{tabular} c \\ Backbone Fusion \end{tabular} & 
\begin{tabular}[c]{@{}p{4.5cm}@{}}- Enhanced integration: fusion within the backbone network improves feature extraction and representation.\\- Potential for real-time processing: can improve efficiency by integrating features during network training.\end{tabular} & 
\begin{tabular}[c]{@{}p{4.5cm}@{}}- Improved feature representation: combines features at an intermediate stage, enhancing object detection and classification accuracy. \\ - Effective in real-time applications: improves processing speed by reducing the need for additional fusion layers.\end{tabular} & 
\begin{tabular}[c]{@{}p{4.5cm}@{}}- Increased network complexity: may complicate network design and increase training time. \\ - Potential for overfitting: more complex models may require regularization to prevent overfitting.\end{tabular} \\ \hline
\end{tabular}
\end{table*}

\item The early fusion methods combining radar and camera data typically involve augmenting radar point clouds with visual information before they are processed through an image-based object detection pipeline. These methods are often compatible with various image-based detectors and are capable of enhancing the detection performance effectively. However, early-fusion techniques usually handle multi-modal fusion and 3D object detection in a sequential paradigm, which may increase inference latency. 
On the other hand, middle-fusion based radar-camera fusion approaches promote a deeper integration of multi-modal representations, resulting in higher-quality detection bounding boxes. Despite this, there still remain challenges in fusing camera and radar features due to their intrinsic heterogeneity and different viewpoints. For convenience, we provide a comparison of the post-processing data types for fusion techniques in Table~\ref{tab:post-processing}.

\end{itemize}

\section{Object Tracking with Radar and Camera}
\subsection{Camera-based Object Tracking}
Depending on the number of targets to be tracked, object tracking can be divided into two categories: Single-Object Tracking (SOT) and Multi-Object Tracking (MOT).

\subsubsection{Visual Single-Object Tracking}
SOT first establishes the initial feature model of the target, typically relying on manually annotating the target's position in the first frame of the image sequence. Then, feature vector of the target region is extracted and used as the initial feature representation of the target. In subsequent frames, SOT algorithms utilize the target's feature model to match the position of the target being searched. The search process can be based on diverse similarity measurement methods, such as Kernelized Correlation Filter (KCF) \cite{henriques2014high}, deep learning feature matching, etc. Nevertheless, since the appearance of target may change due to variations in pose or environment, the target's feature model needs to be updated accordingly to adapt to these variations. As such, SOT algorithms typically adopt an online updating mechanism to update the target's feature model. This can be achieved by re-extracting the feature of the target region in each new frame and updating it with the previous feature model. The most representative visual single object tracker is SiamFC \cite{bertinetto2016fully}.

\subsubsection{Visual Multi-Object Tracking}
The target number and categories in MOT are unknown and diverse, leading to new challenges such as object occlusion, appearance similarity, data association between detections and trajectories, and management of target trajectories. Early visual MOT research mainly focused on optimization strategies for data association problems, such as Probabilistic Data Association (PDA) \cite{lee2003efficient}, Multiple Hypothesis Tracking (MHT) \cite{Blackman2009Multiple}, Joint Probabilistic Data Association (JPDA) \cite{Roecker1993Suboptimal}, etc. 
Currently, mainstream visual MOT research is gradually shifting from the aforementioned paradigms to the Tracking-by-Detection (TBD) paradigm \cite{bewley2016simple,zhou2020tracking,zhang2022bytetrack,zhang2023bytetrackv2}.
The TBD paradigm typically encompasses four major modules, namely the object detection module, motion estimation module, data association module, and trajectory management module \cite{pang2023simpletrack}. 
The object detection module identifies and locates  potential targets in each frame of the image. The motion estimation module predicts and updates the motion state of each target. The data association module matches the detections in the current frame with the track predictions from the previous frame. 
The trajectory management module is responsible for managing the lifetime of targets, including target birth, death, and survival.

Among the visual MOT algorithms, ByteTrack \cite{zhang2022bytetrack,zhang2023bytetrackv2} has achieved outstanding performance in both the accuracy and speed, owing to its concise, versatile, and efficient data association strategies. ByteTrack exploits the similarity between object detection boxes and tracking trajectories to retain high-confidence detection boxes while eliminating background from low-confidence detection boxes, thereby significantly reducing missed detections and improving tracking continuity. Additionally, given the powerful contextual data association capabilities of the Transformer architecture, researchers have attempted to apply Transformer models to enhance tracking robustness. Pioneering work in this field includes TransTrack \cite{sun2020transtrack}, TransMOT \cite{chu2023transmot}, TransCenter \cite{xu2022transcenter}, etc.

\subsection{Radar-based Object Tracking}
Traditional radar target tracking primarily focuses on point targets, fundamentally an estimation problem \cite{mahler2007statistical}, where target numbers and motion parameters are estimated based on measurements from each frame. 
Early radar target tracking relied heavily on filtering techniques, with landmark works such as Wiener filter \cite{wiener1949extrapolation} and Kalman filter \cite{kalman1960new}. Subsequently, nonlinear filtering algorithms like extended Kalman filter (EKF) \cite{einicke1999robust} and unscented Kalman filter \cite{julier1997new} emerged, propelling the development of radar target tracking theory. Additionally, the introduction of the Singer model \cite{singer1970estimating} spurred researchers' interest in motion models, such as the constant velocity (CV) model and constant acceleration (CA) model. 
Afterwards, maneuvering target multiple model methods \cite{dunik2020state} began to emerge for motion state uncertainty, especially the Interacting Multiple Model (IMM) method \cite{mazor1998interacting} and the Variable Structure IMM (VS-IMM) method \cite{li1996multiple}.
The IMM algorithm utilizes Markov-Chain to mitigate the mismatch between the actual motion model and the preset model, while the VS-IMM algorithm further reduces the computational burden of the IMM algorithm and improves tracking accuracy.

Firstly, radar MOT methods roughly fall into two modes: Detection-Based-Tracking (DBT) and Detection-Free-Tracking (DFT). 
The DBT mode first performs object detection on measurement data from each frame to generate traces and then integrates multiple observations to construct target motion trajectories. Considering that object tracking involves removing false alarms generated during the object detection phase, reducing the detection threshold to obtain effective measurements of objects with low SNR can significantly increase the computational burden. Therefore, the DBT architecture is not suitable for tracking low SNR targets. 
In contrast, the DFT paradigm directly operates on multi-frame radar measurements to estimate the number and state of targets, enhancing the detection capability of low SNR targets while reducing false alarm rates \cite{tonissen1996peformance}. The challenge faced by the DFT paradigm lies in the highly nonlinear mapping from radar measurements to target states and the non-Gaussian nature of noise distribution \cite{davey2007comparison}. Accordingly, researchers have developed two approaches based on analytic and sampling methods. Analytical DFT methods mainly utilize nonlinear models such as EKF for linearization approximation, but only applies in high SNR target scenarios. Sampling-based TBD approaches discretize the target state space and introduce dynamic programming \cite{barniv1985dynamic} for processing, but involve substantial computation. Therefore, Particle Filter (PF) algorithms \cite{salmond2001particle} characterizing target states with dynamic grids, are widely adopted.

Secondly, radar MOT researches can also be classified as two branches: model-based and data-driven.
Model-based radar MOT is conventional and concerns mature algorithms, for instance, PDA \cite{streit2002multitarget}, MHT \cite{tang2017multiple}, and Random Finite Set (RFS) \cite{mahler2003multitarget}. In particular, the RFS theory can significantly reduce the computational complexity of MOT problems. The RFS-based MOT methods include Probability Hypothesis Density (PHD) filters \cite{mahler2003multitarget,shi2013pose,yang2023distributed}, Cardinalized PHD (CPHD) filters \cite{Vo2007Analytic,Mahler2008PHD,Ristic2012Adaptive}, Multi-target Multi-Bernoulli (MeMBer) filters \cite{Ristic2013A,Ba-Ngu2017an,yang2018optimization} and their variants such as Cardinality Balanced MeMBer (CBMeMBer) filters \cite{vo2008cardinality}, Labeled Multi-Bernoulli (LMB) filters \cite{yang2023road}, Generalized LMB (GLMB) filters \cite{vo2019multi}, and Poisson Multi-Bernoulli Mixture (PMBM) filters \cite{garcia2018poisson}. Overall, RFS-based trackers demonstrate two main strengths: 1) providing explicit statistical models for integrating multi-target dynamics (e.g., target birth and death) and sensor observations (e.g., omissions and false alarms); 2) circumventing the intractable data association procedure.
Unfortunately, model-based methods are susceptible to prior assumptions and face a severe challenge of mismatch between the assumed mathematical model and the actual physical property.
Another line of research, data-driven radar MOT, has gradually revealed its potential practicality in complex scenes \cite{pearce2023multi}. 
To date, research on deep learning-based radar MOT is relatively limited \cite{ebert2020deep,akita2019object,liu2019deepda}, especially regarding the exploration of target motion uncertainty.

\begin{figure}[h]
	\centering	\includegraphics[width=0.48\textwidth]{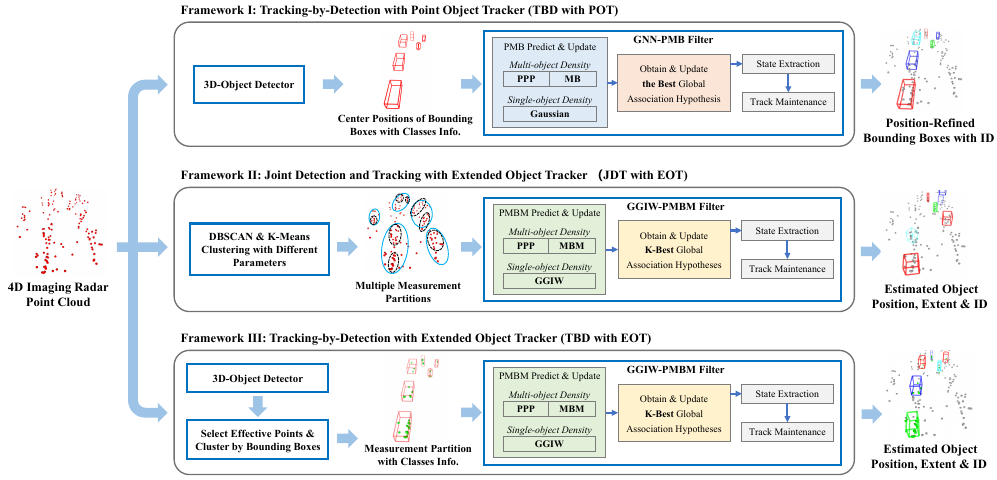}
	\caption{Three MOT strategies for 4D radar  (from \cite{liu2024framework}).}
	\label{Object Tracking2}
\end{figure} 

Thirdly, radar MOT problems can further be conceptually divided into Point-object Tracking (POT) and Extended-object Tracking (EOT). 
In POT, an object only produces one measurement per time step, while in EOT, an object can generate multiple measurements per time step.
The authors in \cite{liu2024framework} systematically conduct a seminal study on comparing POT and EOT frameworks for 3D MOT with 4D imaging radar point clouds. 
They evaluate and contrast the well-established TBD-POT framework, the recently explored joint-detection-and-tracking (JDT) framework, and their devised TBD-EOT framework (as depicted in fig.~\ref{Object Tracking2}). Experimental results indicate the traditional TBD-POT framework remains the preferred choice for 3D MOT, given its superior tracking performance and lower computational complexity.

\subsection{Radar-Camera Fusion based Object Tracking}
Researches on radar-camera fusion for object tracking can be categorized into tracking-before-fusion and tracking-after-fusion.
Generally, tracking-before-fusion schemes involve conducting individual target tracking first and then matching and fusing the output trajectories \cite{chang1997optimal,tian2010algorithms,lu2015exact,botha2016data}. Common trajectory matching methods include trajectory error calculation \cite{wang2014bionic} and neural networks \cite{ji2010incremental}. 
Tracking-after-fusion methods match the centers of image targets with radar targets before executing the tracking module \cite{richter2008radar,sengupta2019dnn,wang2018vehicle}. Common methods for cross-modal target matching include Euclidean distance, Mahalanobis distance, and nearest neighbor algorithms \cite{wu2009collision}.

Specifically, the authors in \cite{dimitrievski2019people} devise a radar-vision fusion method for human tracking. They project the object detection results from the camera image onto the range-azimuth spectrum collected by TI AWR1443 radar, and compute the joint detection likelihood. Peaks in the RA spectrum represent the presence of targets, and these peak positions are fed into a PF tracker. Based on the target-trajectory associations, particles are updated using either associated detections (tracking-after-fusion) or the raw likelihood data itself (tracking-before-fusion). The advantage of utilizing raw likelihood data lies in the continuous tracking of losing targets even if camera or radar signals fall below detection thresholds.
Besides, the authors in \cite{liu2021robust} propose an object detection and tracking framework where Delphi ESR radar is the primary sensor with the camera as a supplementary one, using a decision-level fusion approach. The authors employ Mahalanobis distance for matching observations with targets and utilize the JPDA method \cite{Roecker1993Suboptimal} for target tracking and fusion. They also conduct practical tests in complex scenarios such as foggy, rainy, and congested traffic conditions.

Moreover, the authors in \cite{bai2021robust} develop a multi-object detection and tracking method based on radar-vision fusion. Initially, radar and camera branches independently perform target detection, and the detection results are correlated in the image plane to generate a random finite set. Subsequently, based on the GM-PHD algorithm, improvements are made to the tracking process using ellipse discrimination thresholds, attenuation functions, and simplified pruning methods. Experimental results demonstrate that this algorithm can accurately estimate the number and state of targets in occluded scenarios.
In addition, Cui \textit{et al.} \cite{cui20213d} advance a CNN-based cross-fusion strategy for 3D vehicle detection and tracking. They employ two low-cost 4D mmWave radars and a monocular camera for testing in real-world scenarios. 

\begin{figure}[h]
	\centering	\includegraphics[width=0.48\textwidth]{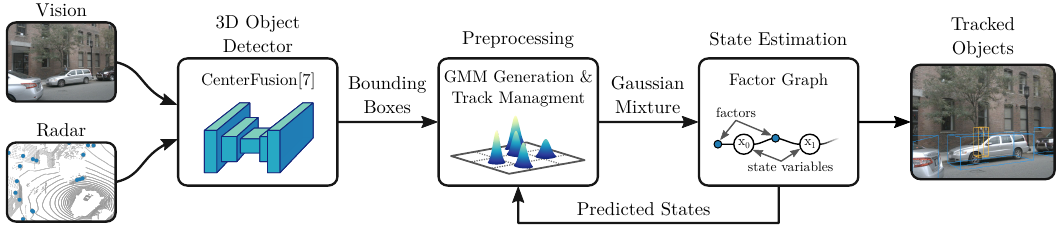}
	\caption{Overview of the FG-3DMOT model (from \cite{poschmann2021optimization}).}
	\label{GMM}
\end{figure}

Given the outstanding performance and scalability of the CenterFusion \cite{nabati2021centerfusion} network in radar-vision fusion for object detection, some researchers have extended it to the field of radar-vision fusion-based tracking. For instance, FG-3DMOT \cite{poschmann2021optimization} first leverages the CenterFusion network for frame-by-frame object detection (as shown in Fig.~\ref{GMM}). It then represents all detections from each frame as a Gaussian Mixture Model (GMM) and assigns the GMM to each tracked target. Target trajectories are estimated through optimization based on factor graphs \cite{poschmann2020factor}, implicitly addressing data association issues. Although FG-3DMOT has the advantage of achieving online 3D multi-object tracking without explicit data association, the performance of the factor graph is highly dependent on the quality and accuracy of the input data. If there is significant noise in the sensor data, the accuracy of the tracking results may be adversely affected.
Additionally, Nabati \textit{et al.} \cite{nabati2021cftrack} combine the strengths of CenterFusion and CenterTrack \cite{zhou2020tracking} to create an end-to-end radar-vision fusion tracker dubbed CFTrack. It employs CenterFusion for object detection and improves the greedy algorithm in CenterTrack for data association. The authors additionally incorporate depth and velocity information of detected targets to enhance the cost function in data association, making the tracking algorithm more adaptable to complex scenes with target occlusion.

\begin{figure*}[h]
	\centering	\includegraphics[width=1\textwidth]{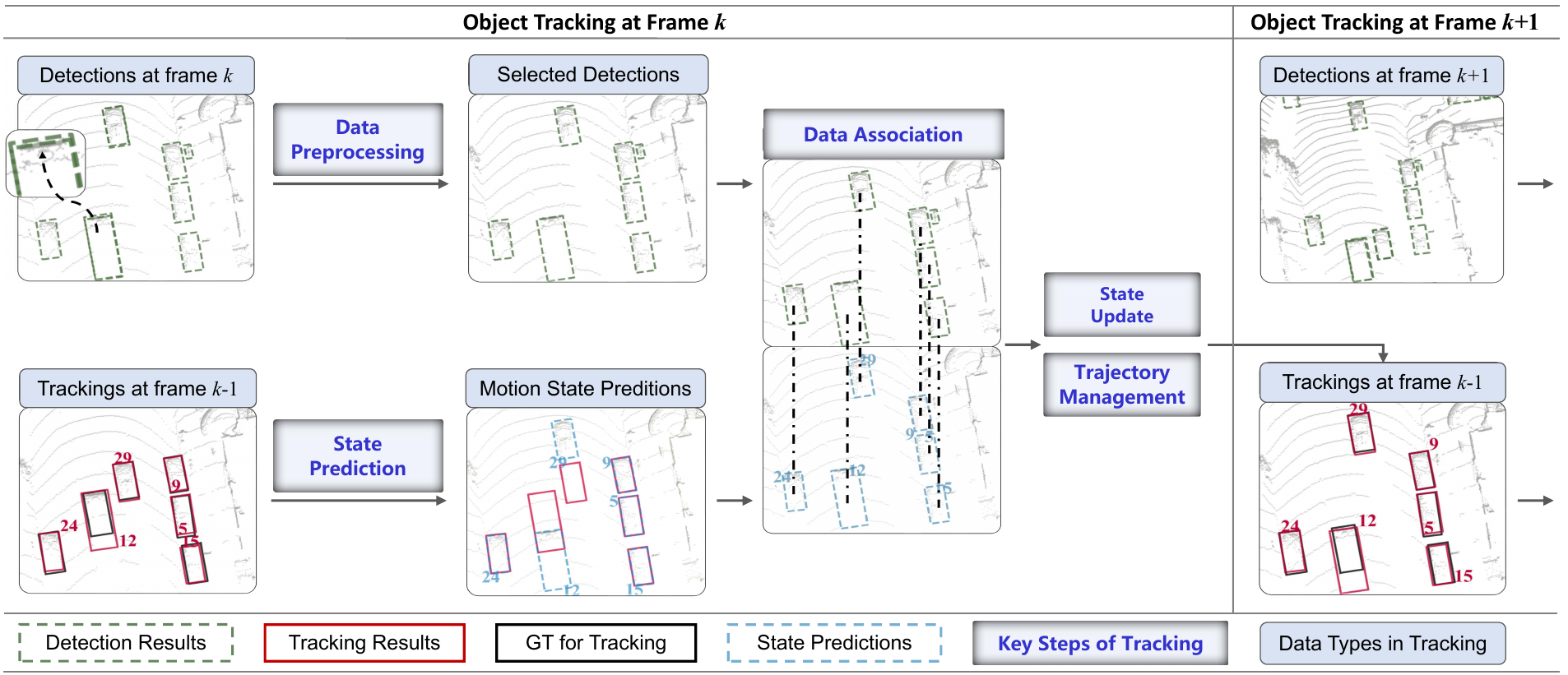}
	\caption{An illustration of the mainstream tracking-by-detection paradigm.}
	\label{Object Tracking}
\end{figure*} 

Recently, the authors in \cite{zhang2023mixedfusion} propose an effective multi-modal perception framework called MixedFusion, aiming to achieve hierarchical fusion of camera images, LiDAR point clouds, and radar point clouds, alongside optimizing the combination of detection and tracking algorithms. 
In fusion-based tracking module, the authors design a cascaded multipriority matching strategy between LiDAR-camera fusion detections and radar observations to conduct object tracking and state updating within a unified 3D world coordinate system. 
The experimental outcomes illustrate a significant enhancement in the performance of 3D object tracking, particularly in challenging weather conditions.
Additionally, the authors in \cite{wolters2024unleashing} propose HyDRa, a creative camera-radar fusion paradigm for diverse 3D perception tasks including object detection, tracking, and occupancy prediction.
It realizes an innovative state-of-the-art performance in camera-radar fusion with 58.4$\%$ AMOTA on the authoritative nuScenes dataset, showing a promising path for future research in the field of radar-camera tracking.

\subsection{Summary and Lessons Learned}
In this section, we have expatiated on various camera-based, radar-based, as well as radar-camera fusion-based strategies for object tracking. 
From this review, we conclude the following lessons learned:

\begin{itemize}
\item Currently, the most advanced visual object tracking algorithms mostly follow the tracking-by-detection paradigm \cite{pang2022simpletrack} (as shown in Fig.~\ref{Object Tracking}), which have shown commendable results. For the fusion of radar and vision in multi-object tracking, the most promising methods also adhere to this pipeline. 
A significant concern is the heavy reliance of these methods on object detectors for their performance. Another nuisance is the complexity of MOT solutions, which introduces numerous parameters and makes algorithm development challenging \cite{tang2023strong}. This complexity also hinders method tuning, while the difficulty in implementation and result reproducibility adds further barriers for other researchers.

\item  Apart from the tracking-by-detection pipeline, rich features and precise temporal data acquired from both cameras and radars can be integrated into an end-to-end framework, preventing the accumulation of errors from detection to tracking \cite{cheng2024deep}. Extensive training data encompassing various scenarios is crucial for training a one-shot network. As such, addressing overfitting, dataset annotations, and the design of one-shot networks are crucial for overcoming the primary constraints of the end-to-end tracking framework.

\item Research on DL-based radar-camera fusion for multi-object tracking is still in its nascent stages. Existing literature in this field primarily focuses on improving accuracy and inference speed, with limited attention given to uncertainty and robustness \cite{kuan2024boosting}. Additionally, critical issues such as data association and target trajectory management require more in-depth theoretical researches and engineering practices.

\item Intricate external environments necessitate MOT technology with a heightened level of precision and broad applicability. Nonetheless, the timeliness of fusion-based tracking is often unattainable on account of numerous parameters and significant computational expenses \cite{zhang2024online}. 
Consequently, there is a need for further investigation into achieving real-time performance and enhancing inference speed on typical devices.

\end{itemize}

\section{Open Issues and Future Perspectives}
\label{sec:open:issue}
In this section, we discuss the emerging perspectives in the direction of radar and camera fusion.

\subsection{Model Robustness}
The current fusion systems integrating multiple modalities heavily rely on deep learning, which has shown vulnerabilities to forgeries. Consequently, this introduces inherent security risks (e.g., sabotage, adverse conditions, blind spots) to automated industries \cite{qian20223d}. 
Moreover, the advancement of 6G/V2X communication technology has facilitated extensive exploration and implementation of multi-sensor fusion in data-shared collaborative perception. Nevertheless, collaborative data transmission is susceptible to security breaches and data privacy vulnerabilities.

Adversarial attacks in the context of 3D object detection are still relatively nascent in development.
The authors in \cite{cao2021invisible} conduct seminal research on the security problems of multi-modal fusion-based perception in autonomous driving, aiming to explore the feasibility of launching simultaneous attacks on all fusion sources. Besides, they effectively devise a real-world adversarial attack strategy with the goal of creating deceptive adversarial 3D objects. These objects are intended to deceive a victim autonomous vehicle into failing to detect them, leading to a collision.
Besides, the authors in \cite{tu2021exploring} advance adversarial attacks on multi-modal fusion networks by introducing an adversarial textured mesh that can be affixed to a vehicle, rendering it undetectable to the multi-modal fusion networks.

On the other hand, given the vulnerability of RGB images and the defects in radar data, ensuring model robustness in degraded sensor conditions or adverse scenarios is a key challenge for radar-camera fusion. While most existing methods focus on breaking through the public dataset accuracy, only a few works try to address sensory malfunction with corrupted modality input in adverse environments. 
For instance, RadSegNet \cite{bansal2022radsegnet} employs Semantic-Point-Grid (SPG) encoding to independently extract information from the camera and radar, enabling reliable performance with radar data alone when the RGB camera deteriorates. 
Recently, attention mechanisms have also achieved promising performance in incorporating multi-modal information and handling feature corruption. TokenFusion \cite{wang2022multimodal} dynamically predicts the importance of feature tokens and substitutes inferior tokens with corresponding ones from the other modality among single-modal Transformer layers to preserve better information. ImmFusion \cite{chen2023immfusion} utilizes the self-attention Transformer to select informative token features from the two input modalities to accommodate corrupted modalities.

Furthermore, generative models present another avenue for enhancing network robustness by detecting sensor defects or generating novel scenarios. For example, Lekic \textit{et al.} \cite{lekic2019automotive} introduce a Conditional Multi-Generator Generative Adversarial Network (CMGGAN) to generate pseudo-images with surrounding objects detected by radar sensors. 
Developing specialized deep generative models like diffusion models \cite{wu2024robust} is promising for enhancing radar-camera fusion robustness. Diffusion models have the benefit of alleviating the ambiguous nature of radar signals by creating a denoising diffusion model with the embedding of semantic features, even without additional supervision from LiDAR.

\subsection{Modal Uncertainty}
In practice, almost all sensor measurements carry some degree of uncertainty, which can be classified as \textit{epistemic uncertainty} (a.k.a. model uncertainty) or \textit{aleatoric uncertainty} (a.k.a. data uncertainty). Epistemic uncertainty can be decreased by enhancing system knowledge, while aleatoric uncertainty is unavoidable since it stems from inherent observation randomness \cite{feng2021review}.
In particular, radar data frequently exhibits substantial aleatoric uncertainty on account of severe clutter and low resolution. 

Exploiting uncertainty is a feasible scheme to cope with unfamiliar perception scenarios.
More precisely, a multi-modal network is expected to  exhibit higher uncertainty when encountering unseen objects.
For instance, YOdar \cite{kowol2020yodar} utilizes uncertainty to gauge the importance of radar and camera network outputs. By leveraging Bayesian neural networks for estimating posterior distributions, YOdar notably enhances the efficacy of radar-camera fusion, especially in situations with dim illumination. 
Moreover, researchers in \cite{lou2023uncertainty} investigate the integration of predictive uncertainties from individual sensors in multi-modal fusion. They evaluate uncertainty scores and introduce a fusion module that utilizes the Mixture-of-Experts (MoE) framework to incorporate multi-modal uncertainties into fusion pipelines at the proposal level.
Specifically, the uncertainty-incorporated MoE first leverages the Monte Carlo dropout technique \cite{gal2016dropout} and direct modeling method \cite{kendall2017uncertainties} to gauge the uncertainty of each sensor. Subsequently, a dedicated expert network processes the detection outcomes of each sensor while considering the encoded uncertainty. Finally, a gating network scrutinizes the output characteristics of each expert network to ascertain the fusion weights.
In \cite{cai2023robust}, authors introduce innovative approaches utilizing portable thermal cameras and single-chip mmWave radars for reliable human detection, incorporating a unique Bayesian feature extractor and an uncertainty-driven fusion technique to tackle challenges arising from thermal images and noisy radar point clouds.

\subsection{Data Augmentation}
Multi-modal data augmentation faces more complex issues compared to unimodal data augmentation.
For radar data, various data augmentations, including random rotation, scaling, flipping, and shifting, are employed to increase the diversity of samples within the radar point cloud data \cite{nabati2021centerfusion, zheng2022tj4dradset, zhou2023bridging}.
Alternatively, image-based augmentation techniques such as horizontal flipping, range translation, interpolation, and mixing can be employed to augment the radar spectrum data \cite{gao2020ramp}.
For image data, aside from the conventional model-free data augmentation methods \cite{shorten2019survey, khalifa2022comprehensive}, model-based generative augmentation techniques are also prevalent in image processing field, which means generative models are utilized to generate augmented images \cite{hong2021stylemix, zheng2021generative, xu2022style}.
Besides, model-based radar data augmentation is considered a promising area of research.

Due to the differences in information density and data format between radar and image modalities, unimodal data augmentation methods often struggle to be directly applied.
For instance, when implementing CutMix \cite{yun2019cutmix} on the aligned radar and image feature maps, the radar target features are more susceptible to degradation compared to the image target features.
This vulnerability can be attributed to the higher information density present in image data. The dense semantic features of image data contribute to its robustness, whereas radar data, being sparse, tends to be more fragile \cite{zhang2023improving}.
As such, recklessly adopting the same data augmentation for multi-modal fusion data not only fails to boost the model performance but also leads to incorrect model inference \cite{park2023investigating}.
Therefore, radar-camera fusion data augmentation necessitates the consideration of modality individuality and representation commonality.

\subsection{Training Strategy}
Multi-modal fusion networks are prone to facing challenges related to overfitting, leading them to potentially disregard one branch if the training hyper-parameters favor the other branch. In \cite{wang2020makes}, the authors contend that the rates of overfitting and generalization differ among various modalities. Therefore, employing a uni-modal training approach for training a multi-modal network might not yield optimal results for the entire network.
An effective strategy to balance performance involves introducing separate loss functions for each modality. In this vein, even after one modality reaches convergence, the remaining modality can still generalize effectively. 
Additionally, employing weighting factors on loss functions is advantageous in adjusting to the learning rates of individual modalities. 

Additionally, utilizing dropout operations can aid in overcoming overfitting. For example, the authors in \cite{nobis2019deep} introduce BlackIn, a method that deactivates camera image data. By reducing camera input, the network is compelled to rely more on sparse radar data. 
In the same vein, CramNet \cite{hwang2022cramnet} integrates sensor dropout operations during the point cloud fusion phase in training. Unlike the approach in \cite{nobis2019deep}, CramNet opts to mask 3D point features instead of directly manipulating input data. This ensures that the cross-attention mechanism can still be trained using appropriate 2D features in the usual manner.

Moreover, fine-tuning a multi-modal network using pre-trained uni-modal encoders can yield superior results compared to starting fusion training from scratch. The authors in \cite{lim2019radar} employ a weight freezing technique to optimize training hyper-parameters for a single-branch network. These optimized weights are then transferred to the respective branches for training the fusion network. Experimental findings indicate the most effective approach involves pre-training the camera branch and subsequently training the holistic network while preventing gradient propagation through the camera branch.

\subsection{Model Deployment}
Radar-camera fusion presents considerable promise for real-world applications, especially when the fusion models can be implemented on edge equipment \cite{qian2024edge}. 
Compared to powerful servers, edge devices typically come with restricted computational resources, encompassing limitations in memory, bandwidth, and CPU/GPU capabilities. Despite these constraints, they are expected to fulfill demanding criteria for low latency and optimal performance.
Besides, the foundational inquiries on how to implement fusion methods in electric vehicles and automated driving remain obscure. Furthermore, the challenge of mitigating bias in the decision-making process of individual modality component has not been adequately addressed.
Currently, only a couple of studies have reported radar-camera fusion runtime on edge devices like NVIDIA Jetson \cite{shuai2021millieye,yao2024waterscenes}.
Specifically, milliEye \cite{shuai2021millieye} notably surpasses the image-only YOLO baseline in terms of mAP by a discernible margin, achieving up to a $24.3\%$ improvement. However, this performance gain comes at the cost of a $30\%$ increase in runtime, rising from 57.6ms to 74.4ms on the NVIDIA Jetson TX2. Besides, the radar-camera fusion-based YOLOv8 framework in \cite{yao2024waterscenes} achieves an increase in mAP50 from  $84.4\%$ to $88.8\%$ compared to the camera-based YOLOv8 model, with a slight reduction in inference speed from 58.8FPS to 54.2FPS on a NVIDIA RTX 3090.
As such, how to enhance fusion efficiency remains an interesting yet largely unexplored area of research. Implementing network acceleration techniques such as pruning and quantization \cite{stacker2022fusion} presents viable options for improving radar-camera fusion model efficiency. 
For instance, the pruning-based radar-camera fusion method in \cite{stacker2022fusion} enhances the image-only RetinaNet baseline, increasing the mAP from $34.75\%$ to $36.78\%$ on a NVIDIA GeForce RTX 2080, with a marginal increase in latency from 36.4ms to 36.7ms.
Additionally, training multiple models in a distributed manner (e.g., federated learning) can further reduce computational cost \cite{zheng2023autofed}.

Recently, Mamba \cite{gu2023mamba,dao2024transformers} introduces an input-activate mechanism to enhance the state space model, achieving effective sequence modeling with linear computation complexity. Motivated by its potential for efficient training and inference, researchers are committed to employing Mamba model for numerous multi-modal tasks, including large language model \cite{qiao2024vl,zhao2024cobra}, object detection \cite{li2024cfmw,dong2024fusion}, and semantic segmentation \cite{xie2024fusionmamba,wan2024sigma}, and object tracking \cite{huang2024mamba,huang2024exploring}. 
The selective scanning mechanism is a key part of the Mamba model, which was initially tailored for 1D causal sequential data. To manage the inherently non-causal data, existing methods mostly utilize bi-directional scanning \cite{zhu2024vision}. Furthermore, to adequately capture the spatial information in 2D or higher-dimensional data, current strategies often expand the scanning directions. Despite these modifications, there remains a significant demand for more innovative scanning methods to exploit the potential of higher-dimensional non-causal data efficiently.

\subsection{Multi-task Learning}
Integrating multiple tasks into a single model has significant value for practical applications. For multi-task learning, knowledge gained during training for one task can be shared and leveraged to enhance performance across other tasks \cite{liang2019multi}. Moreover, by sharing model features among multiple tasks, the total number of parameters and computational requirements can be reduced, enhancing efficiency in real-time perception scenarios \cite{hu2023planning}. 
Although the research on multi-task learning-based radar-camera fusion is still in its infancy, it is absolutely a critical area of inquiry warranting in-depth research. 
The architecture of multi-task learning-based fusion typically encompasses a shared encoder network, followed by task-specific decoders. This structural design facilitates the network in learning a joint representation from the shared feature extraction layers. The subsequent task-specific layers are then employed to address diverse tasks such as object detection, tracking, and depth estimation. The network's training involves a combined loss function that integrates multiple task-specific losses. One of the paramount advantages of utilizing multi-task learning-based fusion is the enhancement in learning efficiency. By concurrently learning multiple tasks, the model achieves superior generalization compared to single-task learning models. Additionally, the shared representations contribute to enhanced feature extraction capabilities. Nonetheless, the deployment of multi-task learning-based fusion presents several challenges. Firstly, the complexity in network design is a foremost consideration. Crafting an effective multi-task learning network architecture that optimally balances the learning of multiple tasks is non-trivial. Secondly, addressing the computational overhead is crucial. Multi-task learning models, particularly those involving complex tasks and multimodal data, can be computationally intensive, necessitating substantial processing power and memory. Lastly, combining multiple tasks into a unified optimization objective presents a complex optimization challenge, particularly when tasks are interrelated but have distinct performance metrics.

Recently, WaterScenes \cite{yao2024waterscenes} introduces a groundbreaking multi-task dataset that explores the previously uncharted territory of 4D radar-camera fusion for water surface panoptic perception. These tasks include object detection, semantic segmentation, drivable-area segmentation, waterline segmentation, and radar point cloud semantic segmentation. Achelous++ \cite{guanaachelous++} is the first versatile radar-camera fusion framework capable of concurrently performing these five perception tasks with both impressive accuracy and power efficiency. 
It prioritizes the development of energy-efficient and high-speed neural networks specifically suited for edge devices. 
Besides, MCAF-Net \cite{sun2023multi} integrates radar and camera data in a multi-task model that simultaneously detects objects and segments free space, allowing the network to focus on occupied areas of the scene. It aids in learning better feature representations and enhances the radar latent space with segmentation mask information, resulting in an improvement in mAP over performing the detection task alone.  Additionally, 
\cite{sun2024enhanced} advances a learning-based approach to infer the height of radar points and 
devises a multi-task training method. This approach combines height estimation with free space segmentation, leveraging the complementary nature of these tasks to boost model performance. Specifically, by training on both tasks, the model learns to leverage shared features and representations, which can improve overall performance and generalization. The multi-task approach encourages the model to produce more accurate and detailed height maps by integrating contextual information from free space segmentation. 
Technically, most multi-task learning frameworks exhibit high complexity, and striking a balance between model accuracy, speed, and power consumption poses a challenge.

\subsection{Knowledge Distillation}
Knowledge Distillation (KD), initially introduced in \cite{hinton2015distilling}, has emerged as a potent technique for enhancing the performance of radar-camera fusion networks by transferring valuable knowledge from a proficient teacher model to facilitate the learning process of the student model. 
In traditional KD, knowledge is propagated from a large teacher model to a smaller student model within the same modality. Cross-modal KD extends this concept to transfer knowledge across heterogeneous modalities, confronting challenges like aligning modal representations, managing modality-specific nuances, and ensuring efficient transfer without information loss \cite{bang2024radardistill}.

In radar-camera fusion scenarios, data labeling requires significant manual effort and is time-consuming, since the target contours are not readily identifiable from radar data alone. Auto-annotation in radar datasets represents a promising avenue for tackling the labor-intensive nature. In fact, annotations for radar data can be derived by leveraging the pseudo-label from corresponding camera images \cite{wang2021rodnet}. The deficiency of this labeling method lies in its imperfection, as radar targets may not consistently align with the pseudo label from images.
Recently, a growing body of work has pivoted to exploring
fusion-related cross-modal KD, such as UniDistill \cite{zhou2023unidistill} and X$^3$KD \cite{klingner2023x3kd}. 
LiDAR-Camera (LC) fusion methods have consistently shown superior performance for 3D perception compared to Radar-Camera (RC) fusion. As such, 
the authors in \cite{zhao2024crkd} pioneer the application of cross-modal KD architecture to enable LC-to-RC distillation for 3D object detection in the BEV space. Besides, they devise four distillation loss functions to tackle the substantial domain discrepancy and streamline the distillation procedure. Overall, KD methods have the advantage of enhancing the perception performance of the student radar-camera fusion model by learning crucial features from the teacher LiDAR-camera model. The disadvantage is that it requires designing intricate distillation losses and the training is dependent on data quality.

\subsection{Large-scale Pre-training}
Training of DL methods heavily depend on large-scale, human-labeled datasets for achieving outstanding performance. To date, the scale of multi-modal datasets, combining radar and camera data, are typically smaller compared to uni-modal image or text datasets \cite{fei2023self}.
Besides, due to the difficulties involved in labeling radar point cloud datasets, the concept of large-scale pre-training for point cloud representation has inspired cutting-edge researches. 
This strategy focuses on acquiring versatile and beneficial representations from unlabeled data, eliminating the necessity for laborious manual annotations \cite{zhou2023anomalyclip}.

In the context of image and point cloud fusion, there are two mainstream approaches for image-to-point representation learning based on large-scale pre-trained models, namely CLIP-based \cite{radford2021learning} and  MAE-based \cite{he2022masked}.  
For the former, PointCLIP \cite{zhang2022pointclip} is the pioneer to transfer CLIP’s 2D pre-trained knowledge into 3D point cloud understanding. Subsequent work includes PointCLIP V2 \cite{zhu2023pointclip}, CLIP$^2$ \cite{zeng2023clip2}, ULIP \cite{xue2023ulip}, to name a few.
For the latter, PiMAE \cite{chen2023pimae} is the first to align RGB images with point clouds with MAE pre-training. This approach offers several benefits compared to CLIP-like contrastive learning, notably reducing the need for data augmentation. Following research encompasses Point-MAE \cite{pang2022masked}, I2P-MAE \cite{zhang2023learning}, Point-M2AE \cite{zhang2022point}, and so forth.
Overall, reconciling the incompatibility of multi-modal data within a unified system using large-scale pre-training is promising, whilst encountering unique challenges in real-world scenarios, particularly in dynamic and unstructured environments. It is worth noting the unstructured environments \cite{guastella2020learning} refer to settings or surroundings that lack a predefined or orderly organization, with key characteristics including irregular layouts e.g., natural landscapes, urban streets, or crowded public spaces, where objects and obstacles are not arranged in a predictable manner), dynamic elements (e.g., the movement of objects introduces additional complexity and unpredictability), and variability (e.g., weather conditions, lighting changes, or seasonal variations can significantly alter the appearance of an environment). Moreover, most pre-training MAE-based and CLIP-based image-to-point aligning methods are limited for structured point cloud datasets such as 
ModelNet40 \cite{wu20153d} and  ScanObjectNN \cite{uy2019revisiting}.  So far, few works \cite{yao2022detclip,vidit2023clip,chen2023clip2scene} have exploited the large-scale pre-training for point cloud representation in outdoor real-world scenarios. 

\subsection{Missing Modality}
Existing multi-sensor fusion typically presumes that all modalities are accessible and complete. Nonetheless, in real-world applications, multimodal data are prone to be incomplete with some modalities missing due to unexpected factors such as equipment damage, data corruption, and storage loss. Technically speaking, incomplete multi-modal fusion can be achieved through multimodal contrastive learning (e.g., CLIP \cite{radford2021learning}), which commonly maps features of diverse modalities into an aligned embedding. Nevertheless, this method may not be decent since heterogeneous modalities contain both shared information (which is consistent across modalities) and unique information (which is specific to each modality) \cite{dong2024simmmdg}, making it insurmountable to align them completely. Taking cameras and radars as an example, the camera typically provides intuitive information like visual texture, which is unattainable in radar data. Conversely, radar offers instantaneous velocity and frequency-domain information that is absent in image data. These unique pieces of information are specific to each modality. As such, directly projecting the diverse modal features into a common embedding space preserves merely modal-shared information while neglecting modal-specific details, tending to bring about performance degradation on downstream tasks \cite{jiang2023understanding}. Alternatively, some approaches \cite{wang2023multi2,dong2024simmmdg,yao2024drfuse} disentangle shared and complementary information across heterogeneous modalities, leveraging the shared information for reconstruction or downstream tasks. Nevertheless, most of these works focus on modalities that share a significant amount of common information. Managing missing modalities in highly heterogeneous settings, such the fusion of radar and camera data, remains an open challenge.

\subsection{Dataset Expanding}
The scaling law indicates that DL methods require huge training datasets to achieve higher accuarcy. Nonetheless, multi-modal datasets that combine radar and camera data are significantly smaller compared to single-modal image datasets. 
Additionally, category distribution in radar-camera datasets is often skewed, with a predominance of vehicle labels and fewer instances of pedestrians and bicycles. This imbalance in category distribution can lead to overfitting in DL models.
As such, how to expand radar-camera datasets for detection and tracking tasks is a critical area of inquiry warranting further investigation. 
One promising approach is synthetic data generation, i.e., developing simulation tools (such as CARLA \cite{dosovitskiy2017carla}) to create synthetic datasets. These tools can simulate radar signals in various scenarios, providing labeled data for training and testing. Besides, AIGC \cite{du2024age} techniques such as Diffusion \cite{chi2024rf} can be exploited to generate radar data. Synthetic data can be particularly valuable for simulating rare events or hazardous situations that are hard to capture in real-world settings. 
Another inspiring method is automatic annotation, i.e., integrating auto-labeling schemes (such as OpenAnnotate2\cite{zhou2024openannotate2}) to efficiently acquire high-quality annotated realistic data.

\section{Conclusion}
\label{sec:conclusion}
Camera and radar sensors are synergistic in nature, as they harness complementary data and operate effectively across diverse environmental scenarios. 
Despite recent strides made in LiDAR-camera fusion, the fusion of radar and camera perception remains an evolving area, fraught with notable challenges to be addressed.

This paper presents a comprehensive and up-to-date survey of multi-modal object detection and tracking focusing on radar-camera fusion.
To promote methodological progress and facilitate a thorough comparison, we review key techniques for radar-camera fusion, including sensor calibration, modal representation, data alignment, and fusion operation. 
Finally, we provide several future perspectives in radar-camera fusion.  We expect this paper to provide insights for both theoretical researchers and system designers. 

\section*{Acknowledgement}
This study was supported under the RIE2020 Industry Alignment Fund – Industry Collaboration Projects (IAF-ICP) Funding Initiative, as well as cash and in-kind contribution from the industry partner(s).

\bibliographystyle{ieeetr} 
\bibliography{COMST}

\end{document}